\pdfoutput=1

\documentclass[11pt]{article}

\usepackage[]{acl}

\usepackage{times}
\usepackage{latexsym}

\usepackage[T1]{fontenc}

\usepackage[utf8]{inputenc}

\usepackage{microtype}
\usepackage{inconsolata}
\usepackage{amsmath,amsfonts,bm}

\def\eqref#1{equation~\ref{#1}}
\def\1{\bm{1}}

\DeclareMathAlphabet{\mathsfit}{\encodingdefault}{\sfdefault}{m}{sl}
\SetMathAlphabet{\mathsfit}{bold}{\encodingdefault}{\sfdefault}{bx}{n}

\usepackage{booktabs}
\usepackage{graphicx}

\usepackage{tabularx}
\usepackage{ragged2e}
\newcolumntype{Y}{>{\RaggedRight\arraybackslash}X}
\usepackage{adjustbox}
\usepackage{rotating}
\usepackage{transparent}
\usepackage{multirow}
\usepackage[linesnumbered,ruled,vlined]{algorithm2e}
\usepackage{subcaption}
\usepackage{xcolor, colortbl}

\makeatletter
\DeclareRobustCommand\onedot{\futurelet\@let@token\@onedot}
\def\@onedot{\ifx\@let@token.\else.\null\fi\xspace}

\def\eg{\emph{e.g}\onedot} 
\def\ie{\emph{i.e}\onedot}

\makeatother

\newcommand{\sta}[2]{\scriptsize{#1 $\pm$ #2}\normalsize}
\definecolor{niceorange}{HTML}{F86624}
\definecolor{niceblue}{HTML}{072AC8}
\definecolor{nicepurple}{HTML}{5D3A9B}
\newcommand{\blue}[1]{\textcolor{niceblue}{#1}}
\newcommand\orange[1]{\textcolor{niceorange}{#1}}
\newcommand\purple[1]{\textcolor{nicepurple}{#1}}

\title{Sequential Compositional Generalization in Multimodal Models}

\newcommand{\cop}{4}
\newcommand{\koc}{2}
\newcommand{\kuis}{3}
\newcommand{\hac}{1}
\newcommand{\pio}{5}

\author{Semih Yagcioglu$^{\hac}$ ~ Osman Batur İnce$^{\koc,\kuis}$ ~ Aykut Erdem$^{\koc,\kuis}$\\ \textbf{Erkut Erdem}$^{\hac}$ ~ \textbf{Desmond Elliott}$^{\cop, \pio}$ ~ \textbf{Deniz Yuret}$^{\koc,\kuis}$\\
  $^{\hac}$Hacettepe University\quad
  $^{\koc}$Koç University \quad 
  $^{\kuis}$KUIS AI Center \\
  $^{\cop}$University of Copenhagen\quad
  $^{\pio}$Pioneer Centre for AI}

\begin{document}
\maketitle
\begin{abstract}

The rise of large-scale multimodal models has paved the pathway for groundbreaking advances in generative modeling and reasoning, unlocking transformative applications in a variety of complex tasks. However, a pressing question that remains is their genuine capability for stronger forms of generalization, which has been largely underexplored in the multimodal setting. Our study aims to address this by examining sequential compositional generalization using \textsc{CompAct} (\underline{Comp}ositional \underline{Act}ivities)\footnote{Project Page: \url{http://cyberiada.github.io/CompAct}}, a carefully constructed, perceptually grounded dataset set within a rich backdrop of egocentric kitchen activity videos. Each instance in our dataset is represented with a combination of raw video footage, naturally occurring sound, and crowd-sourced step-by-step descriptions. More importantly, our setup ensures that the individual concepts are consistently distributed across training and evaluation sets, while their compositions are novel in the evaluation set. We conduct a comprehensive assessment of several unimodal and multimodal models. Our findings reveal that bi-modal and tri-modal models exhibit a clear edge over their text-only counterparts. This highlights the importance of multimodality while charting a trajectory for future research in this domain.

\end{abstract}

\section{Introduction}\label{sec:intro}

\begin{figure*}[t]
    \centering
    \scalebox{0.76}{
        \begin{tabular}{cc@{$\;$}c@{$\;$}cc}
            \toprule
            \multicolumn{4}{c}{\textbf{Inputs (keyframes and utterances)}} & \textbf{Targets (next utterance)} \\ \midrule
            \multirow{12}{*}{\begin{rotate}{90}{Training}\end{rotate}} & 
            \adjincludegraphics[valign=M,width=0.275\linewidth]{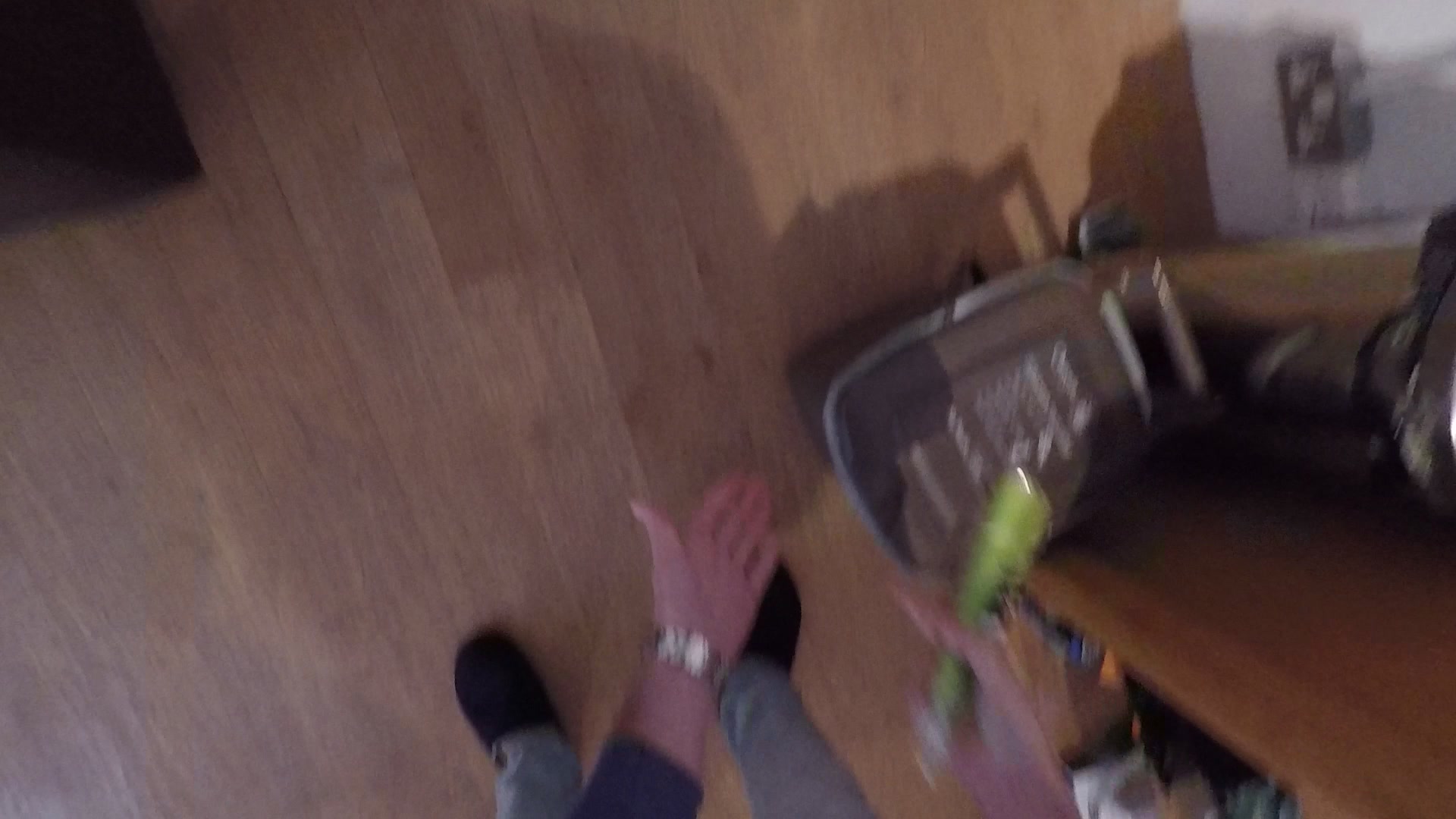} & \adjincludegraphics[valign=M,width=0.275\linewidth]{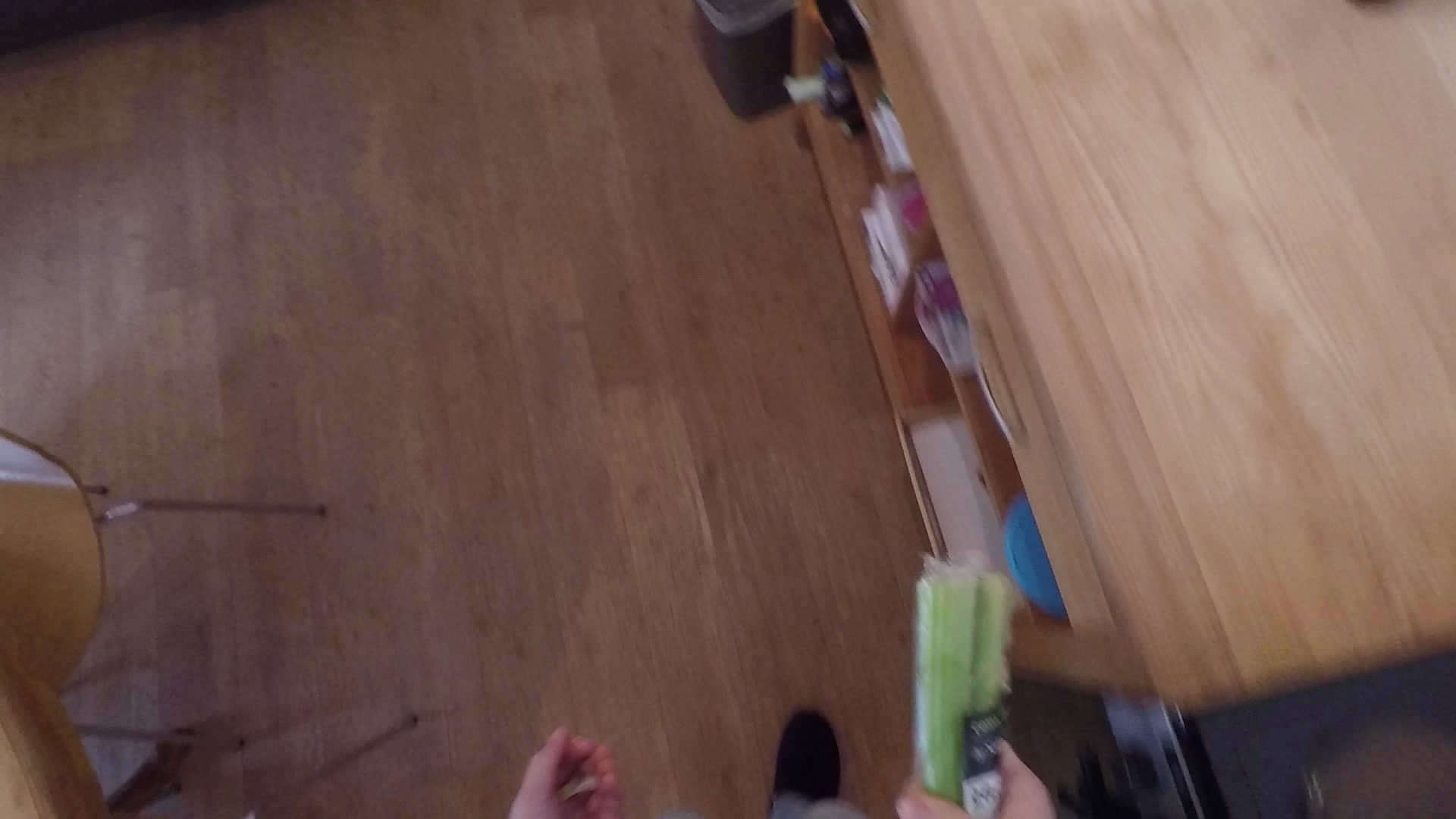} & \adjincludegraphics[valign=M,width=0.275\linewidth]{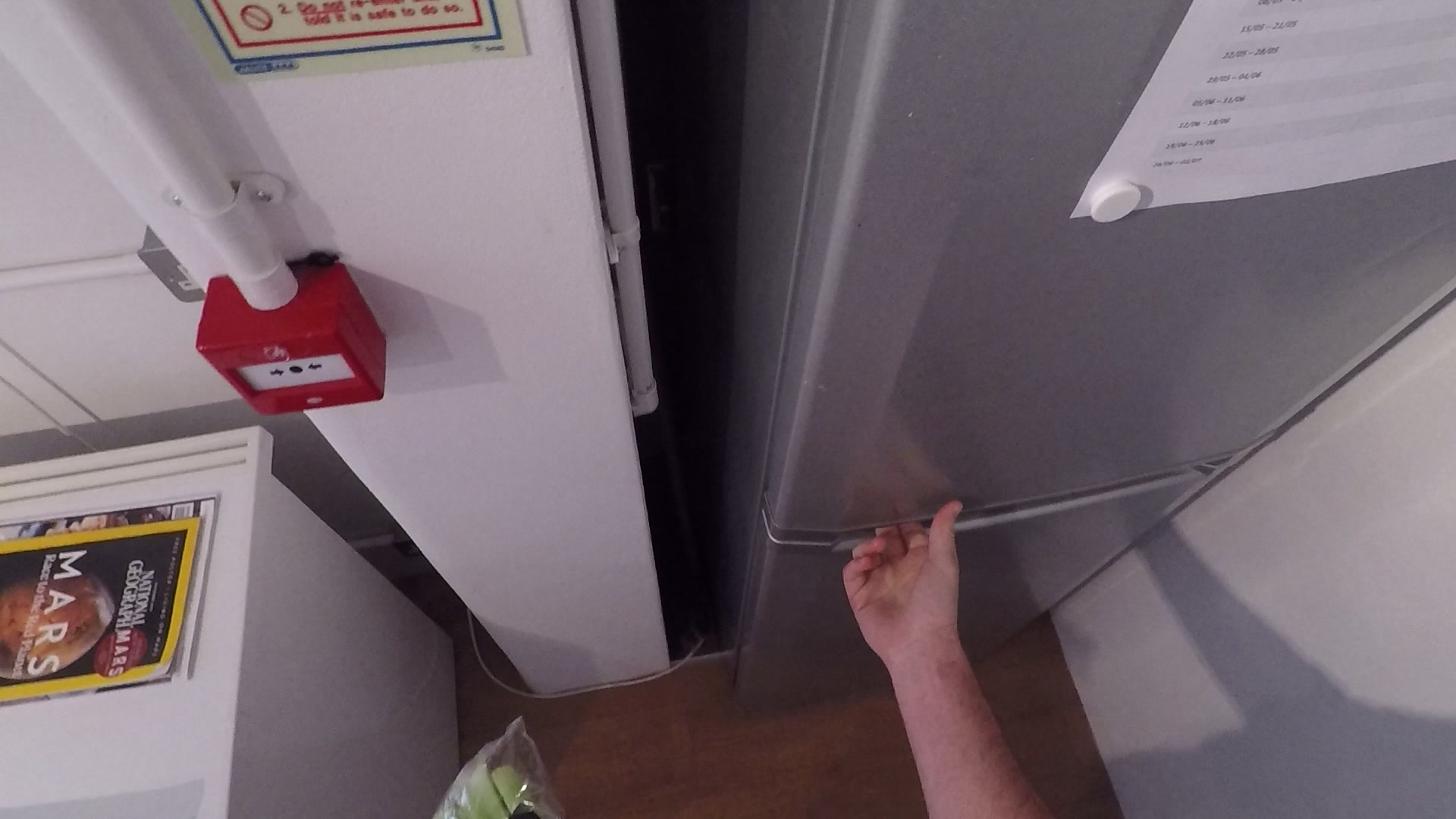}  & 
            {\transparent{0.4}\adjincludegraphics[valign=M,width=0.275\linewidth]{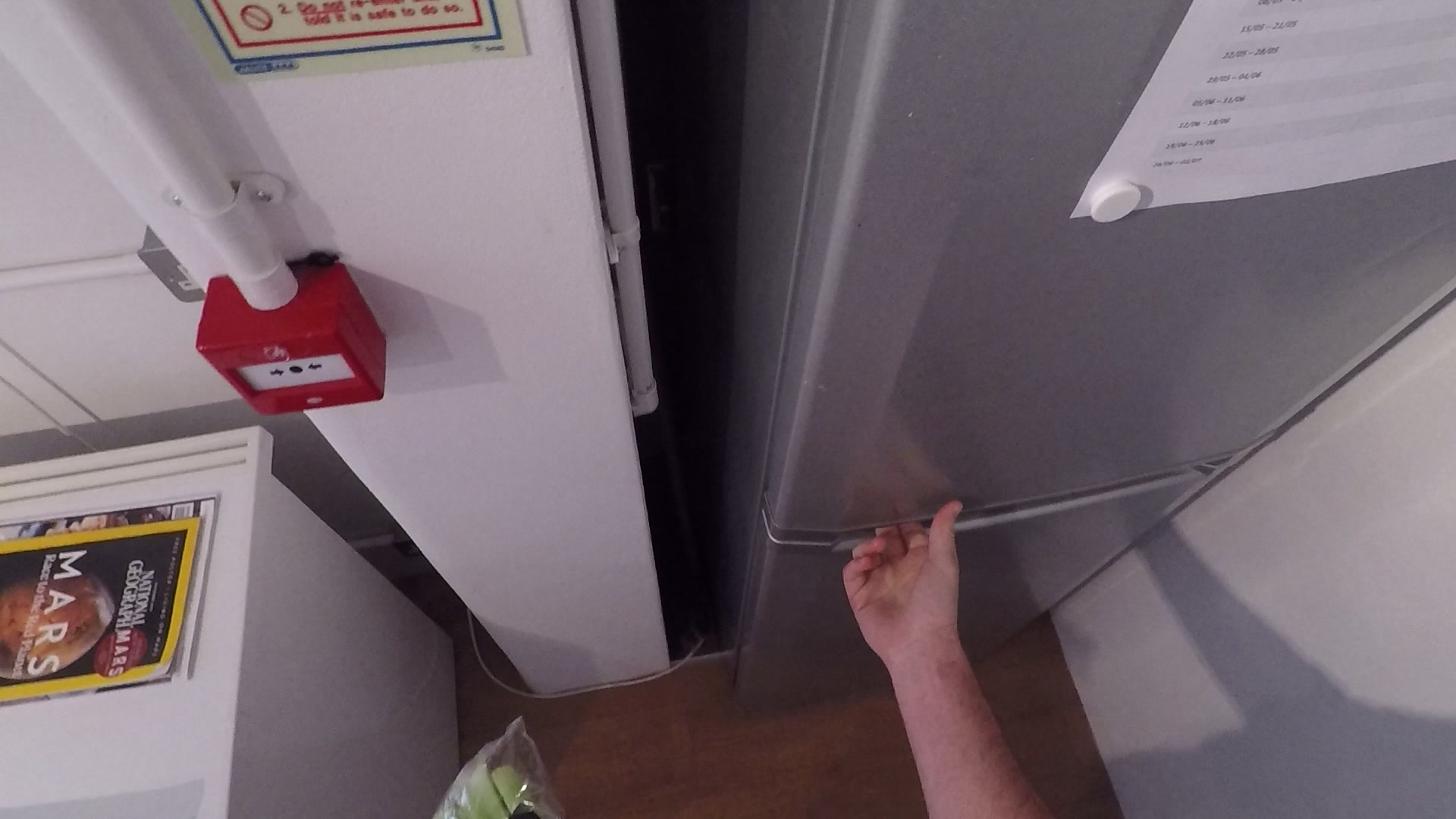}} \\
                 & take celery & {\footnotesize throw things into garbage bin} & open fridge &  put celery back into fridge \vspace{0.3cm}\\
            \vspace{0.1cm}
            & \adjincludegraphics[valign=M,width=0.275\linewidth]{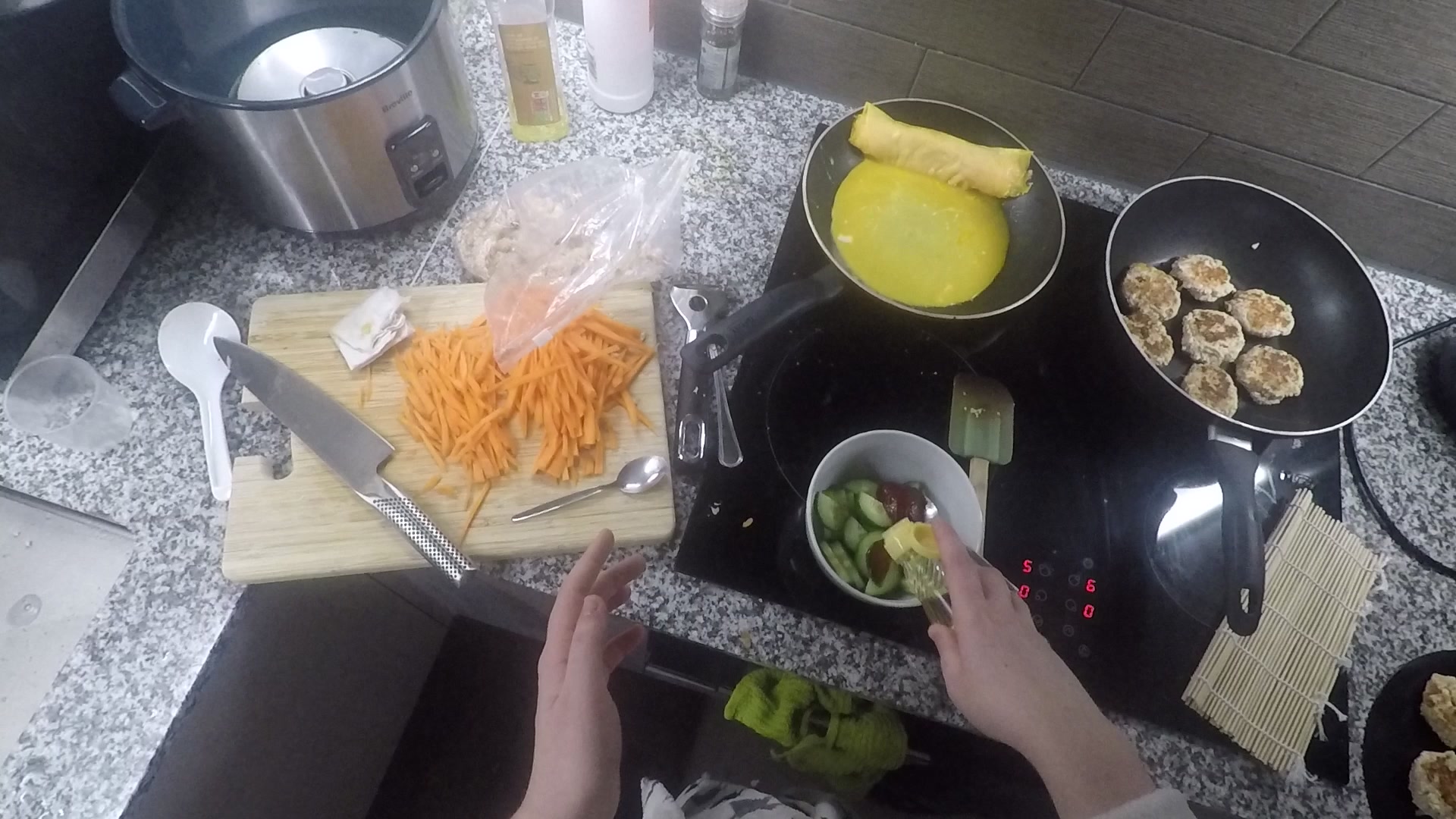} & \adjincludegraphics[valign=M,width=0.275\linewidth]{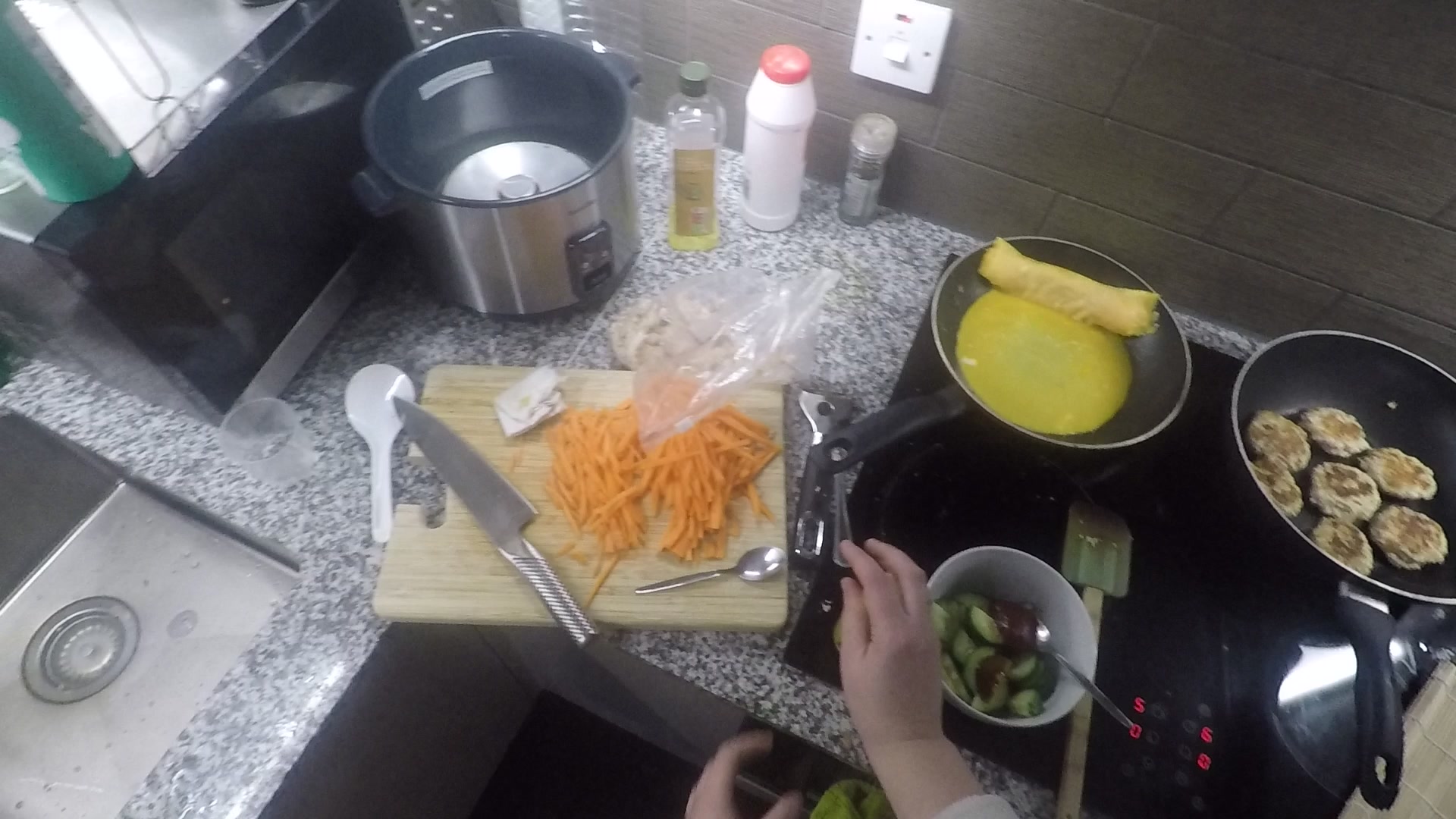} & \adjincludegraphics[valign=M,width=0.275\linewidth]{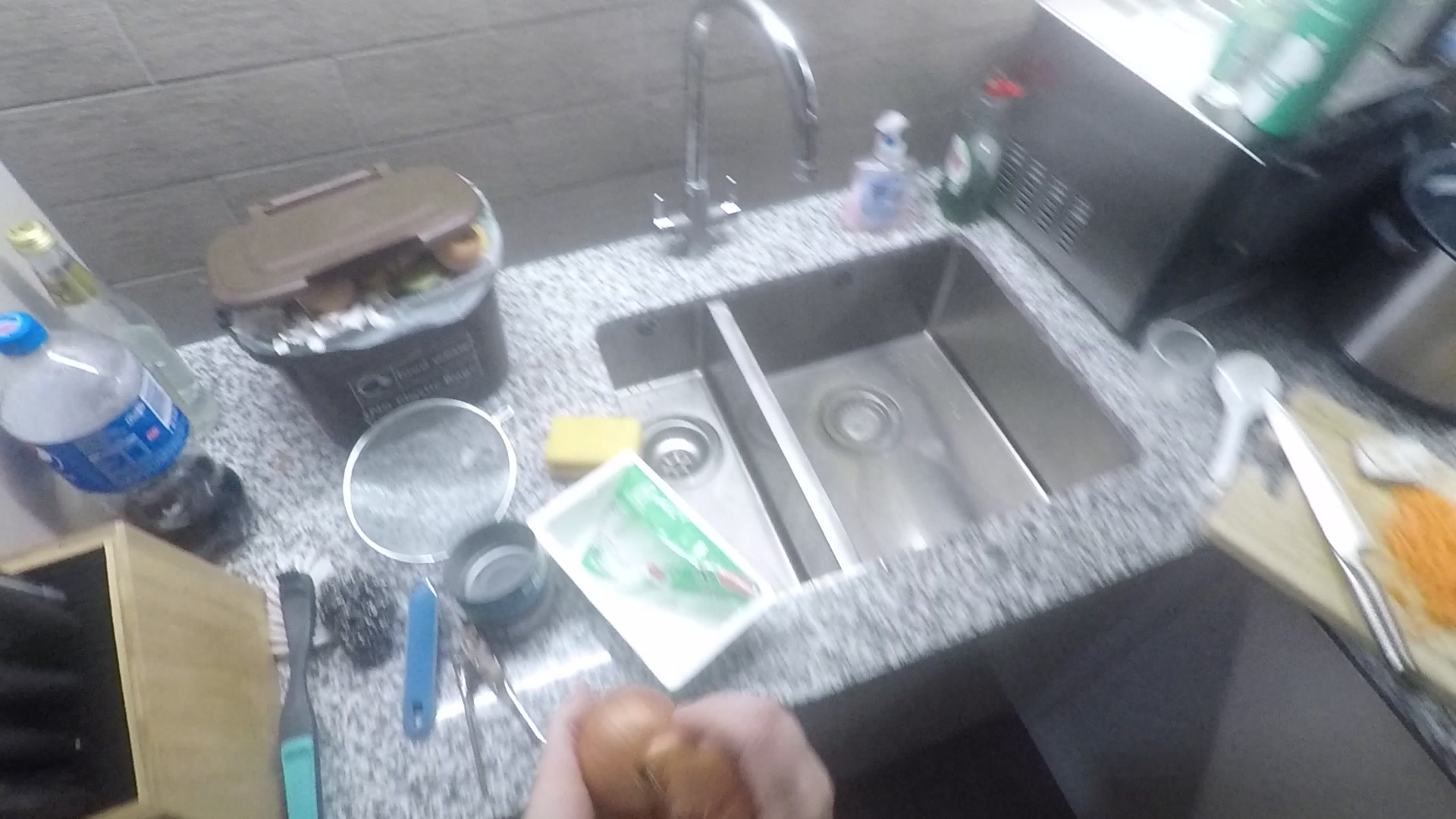} &
            {\transparent{0.4}\adjincludegraphics[valign=M,width=0.275\linewidth]{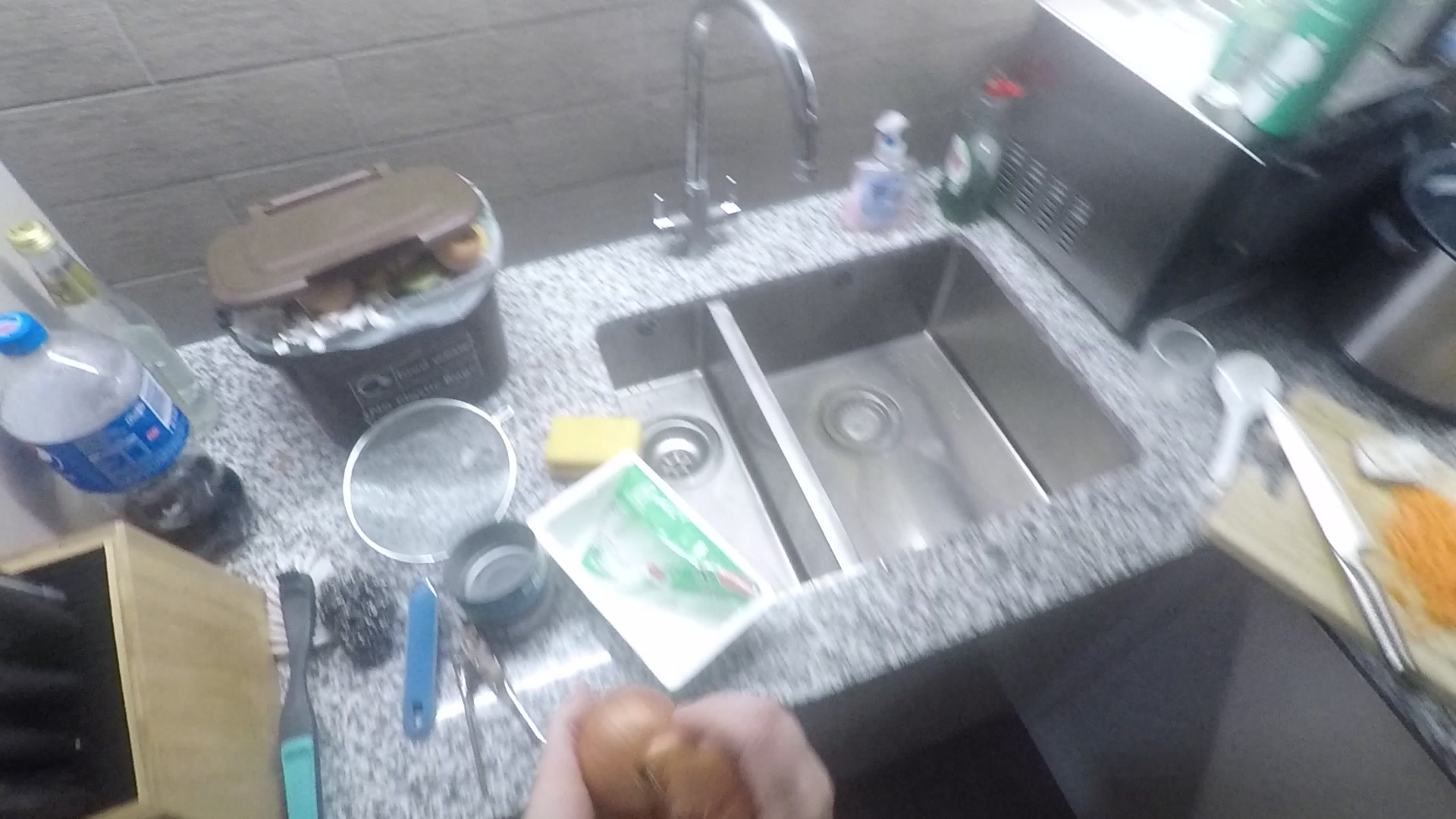}}
                 \\
                 & pour sesame oil & close sesame oil & pick up onion & cut onion in half \vspace{0.3cm} \\
           
            \midrule
            &
            \adjincludegraphics[valign=M,width=0.275\linewidth]{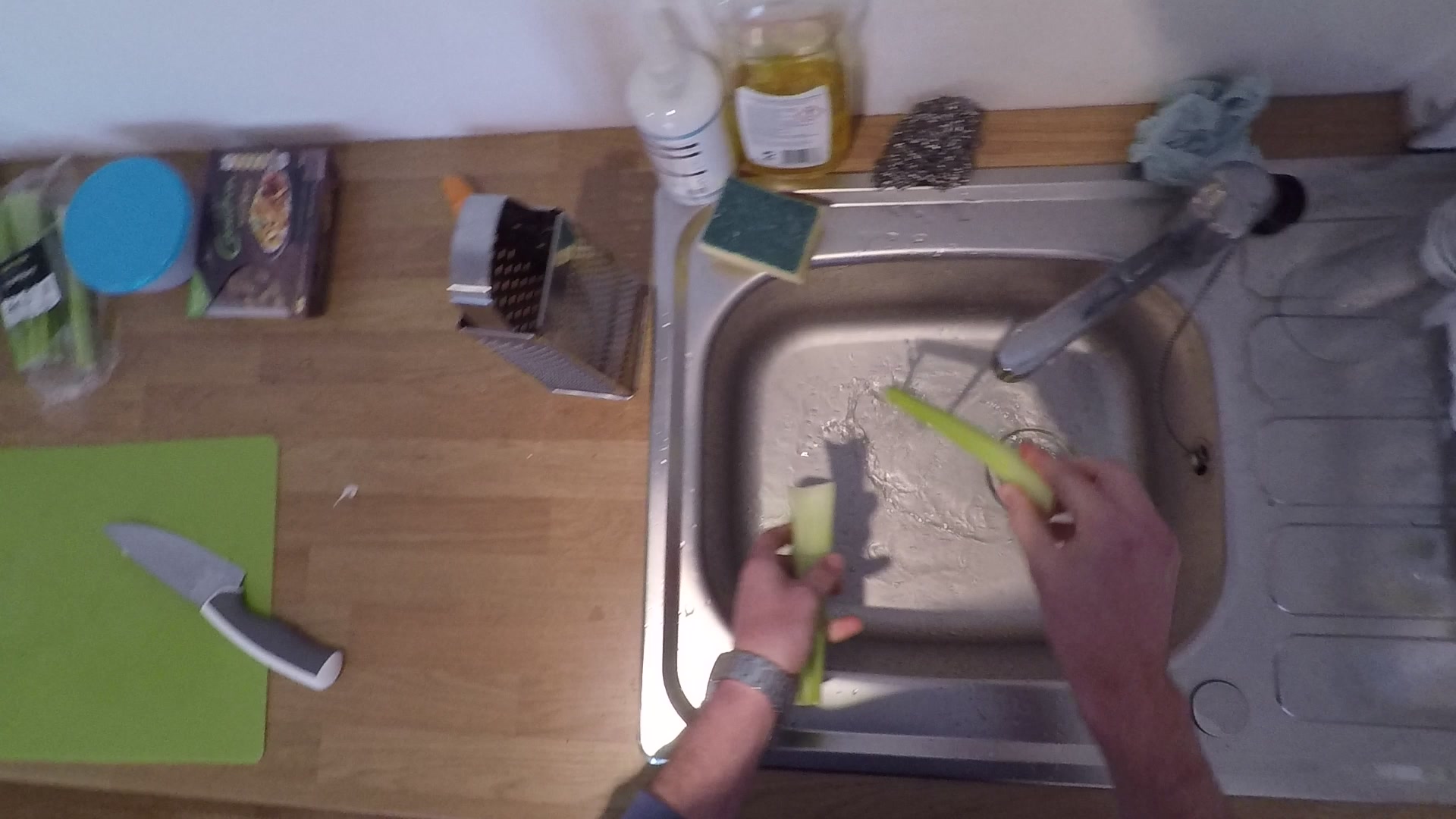} & \adjincludegraphics[valign=M,width=0.275\linewidth]{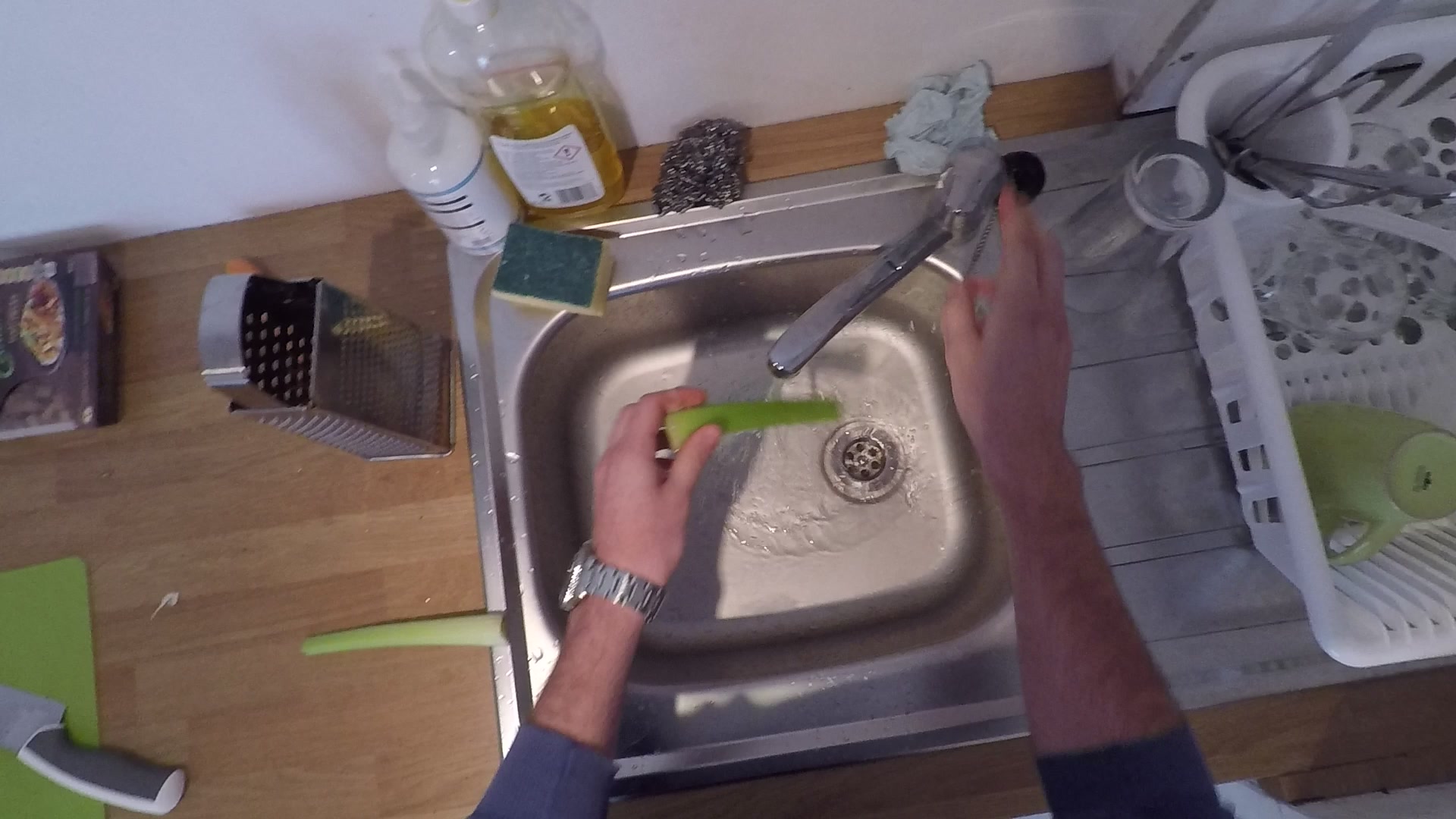} & \adjincludegraphics[valign=M,width=0.275\linewidth]{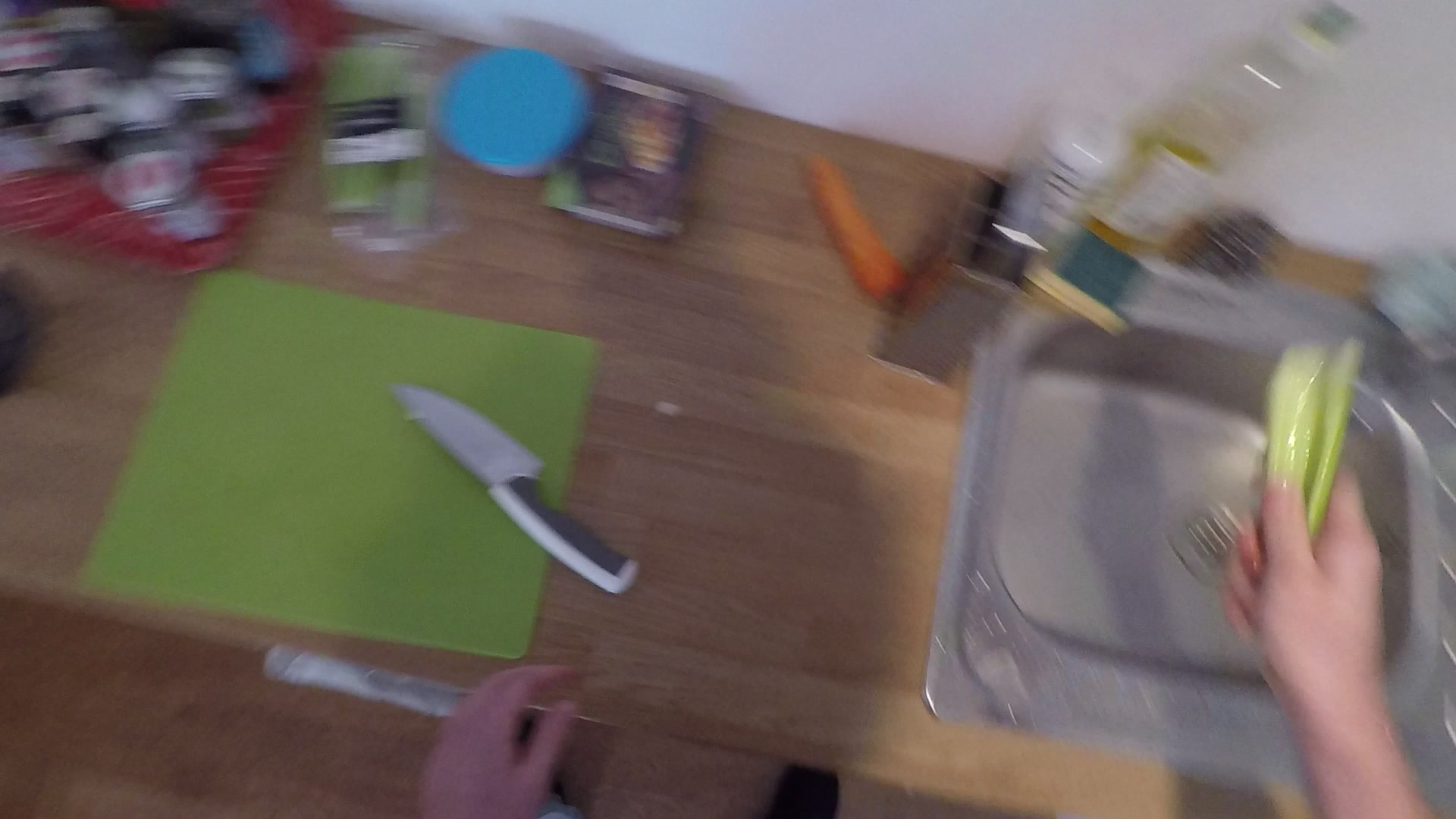}  & 
            {\transparent{0.4}\adjincludegraphics[valign=M,width=0.275\linewidth]{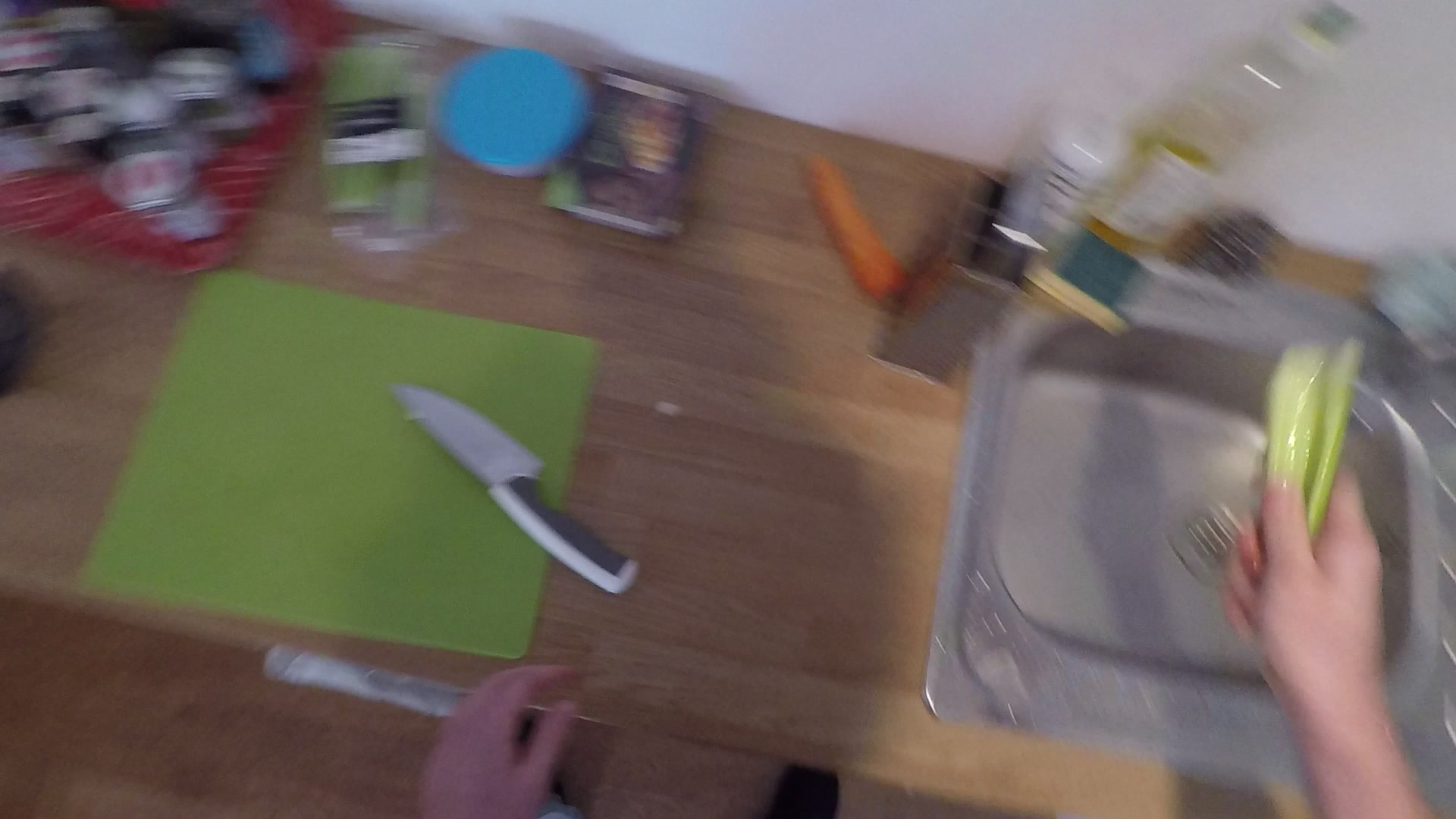}} 
                 \\
                \multirow{2}{*}{\begin{rotate}{90}{$\qquad$ Evaluation}\end{rotate}} & wash celery & close tap & put down celery & {\cellcolor[rgb]{0.7725,0.8784,0.7059}
                 cut celery} \\
          
            \bottomrule
        
        \end{tabular}
    }
    \caption{Overview of the compositional generalization setup in our \textsc{CompAct} dataset. During training, the model has seen the verbs \textsl{wash}, \textsl{close}, \textsl{put down}, \textsl{throw}, \textsl{open}, \textsl{pour}, \textsl{cut} and \textsl{pick up} with the objects \textsc{garbage bin}, \textsc{fridge}, \textsc{sesame oil}, \textsc{onion}, and \textsc{celery}. It has never seen the composition of \textsl{cut} and \textsc{celery}, and thus needs to generalize to this novel composition at test time.}\label{fig:overview}
\end{figure*}

Humans possess a remarkable ability to rapidly understand new concepts by leveraging and combining prior knowledge. This compositional generalization allows for an understanding of complex inputs as a function of their constituent parts. For instance, having grasped the meanings of ``dax'' and ``walk twice'' humans can effortlessly understand ``dax twice''~\citep{lake2017generalization}. However, even as neural networks trained on increasingly larger datasets achieve impressive results across a wide range of tasks, their ability to compositionally generalize remains limited.
Recently, the research community has demonstrated growing interest in evaluating models under different distributions, such as temporal shifts~\citep{lazaridou2021mind,liska2022streamingqa}, or unseen compositions~\citep{lake2017generalization,ettinger2018assessing,bahdanau2018systematic,suris2019learning}. 
Within the domain of multimodal learning, prior investigations into compositionality have primarily delved into visual grounding~\citep{thrush2022winoground}, downstream multimodal tasks like image captioning~\citep{nikolaus2019compositional,DBLP:journals/corr/abs-2005-00785} and visual question answering~\citep{bahdanau2018systematic}, or vocabulary acquisition from videos~\citep{suris2019learning} or with interactive agents~\citep{hill2019emergent}.

Addressing the challenge of compositional generalization in the context of multimodal models is increasingly important with the recent advances in large multimodal foundation models, such as \mbox{GPT-4}~\cite{openai2023gpt4}, Flamingo~\citep{alayrac2022flamingo}, and IDEFICS~\citep{Laurenccon2023OBELISCAO}. Experimenting with closed-source or proprietary models introduces challenges, including reproducibility issues, associated costs, and limited transparency regarding their development and training methodologies \citep{nityasya2023scientific}. 
This inspires us to investigate the potential of open-source large multimodal foundation models -- IDEFICS in particular, for multimodal sequential compositional generalization, which we define as the model's capability to understand and generate predictions about novel compositions of primitive elements derived from sequential multimodal inputs
-- for instance, video data wherein actions unfold in a discernible order. Consider the process of cooking onions: one typically needs to \textsl{peel} and \textsl{slice} an \textsc{onion} before \textsl{frying} it in a \textsc{pan}. Our central inquiry revolves around the proficiency of models in comprehending such sequential and compositional activities.\footnote{Note that this differs from in-context learning, where large-scale pretrained models are prompted for a task in a zero-shot setting, given a support set of task demonstrations.}

In this study, we introduce \textsc{CompAct} (\underline{Comp}ositional \underline{Act}ivities) to investigate multimodal sequential compositional generalization, a uniquely constructed compositional dataset curated from the EPIC KITCHENS-100 dataset~\citep[EK-100]{damen2020ek100}. The EK-100 dataset encompasses 100 hours of egocentric video footage from 45 distinct kitchens, documenting people performing routine household tasks. Each video contains three streams of information: \textit{visual data} in the videos; \textit{audio data} involving non-narrative audio elements --such as the sounds associated with chopping an onion; and \textit{textual data} in the form of short, crowd-sourced descriptions of the depicted activities, like ``slice the carrot'', ``pick up the milk'', or ``wash the plate''. From these descriptions, individual verb and object concepts such as  \textsl{slice}, \textsl{pick up}, \textsl{wash}, and \textsc{carrot}, \textsc{milk}, \textsc{plate} are extracted. The compositional splits are devised based on the verb and object concepts gleaned from the video descriptions, resulting in training and evaluation sets showcasing similar distributions of atomic concepts but featuring varied combinations therein. Consequently, models should compositionally generalize from the training data. Aligning with the ``dax twice'' principle from~\citet{lake2017generalization}, if a model has been trained with videos illustrating how to \textsl{slice} various food items, excluding \textsc{root vegetables}, then it should be capable of compositionally generalizing to understand what it means to \textsl{slice} the \textsc{root vegetables} from previously unseen instances.

In our study, we conduct a comprehensive evaluation of publicly available models, encompassing encoder-only pretrained models such as ImageBind~\citep{girdhar2023imagebind} and MERLOT Reserve~\citep{zellers2022merlot} in addition to (multimodal) large language models (LLMs) like LLaMA2~\citep{Touvron2023Llama2O} and IDEFICS~\citep{Laurenccon2023OBELISCAO}. These models exhibit versatility in processing various combinations of input streams, ranging from language-only to combinations like video + language, video + audio, and even video + language + audio. Our key experimental finding indicates the formidable challenge that all of these models face in mastering compositional generalization. Yet, it becomes abundantly clear that the utilization of multimodal input sources yields discernible advantages, suggesting a promising direction for refining future models.

\section{The \textsc{CompAct} Dataset}

In our pursuit to systematically examine multimodal sequential compositional generalization, we devised the \textsc{CompAct} dataset, leveraging sequences from the EK-100 dataset \citep{damen2020ek100}. As previously noted, each video in the EK-100 features first-person perspectives of unscripted kitchen activities occurring within natural household environments. A video is composed of a sequence of shorter clips, represented as $\mathbf{V} = (\mathbf{v}_1,\dots,\mathbf{v}_k)$, each of which is accompanied by manually annotated English narrations, denoted by $\mathbf{x}_1,\dots,\mathbf{x}_k$, describing the activities within. Additionally, these clips are integrated with audio tracks, $\mathbf{a}_1,\dots,\mathbf{a}_k$, which contain the sounds of occurring actions. We define an \emph{instance} in the dataset as a combination of video -- audio -- narration: $(\mathbf{V},\mathbf{A},\mathbf{X})$
Each instance consists of a window of 4 clips, with the initial 3 clips
serving as context and the last one designated for prediction.

Given this dataset, our primary focus is to facilitate researchers in exploring how multimodal models compositionally generalize to unseen combinations of concepts. We meticulously curate the \textsc{CompAct} dataset to ensure a specific property: the individual concepts are consistently distributed across training and evaluation sets, while their compositions are novel in the evaluation set. This design mandates that a model should exhibit systematic generalization when interpreting the evaluation set. To illustrate, refer to the example shown in Fig.~\ref{fig:overview}. During training, the model comes across nouns such as \textsc{celery}, \textsc{garbage bin}, \textsc{fridge}, \textsc{onion}, and verbs including \textsl{take}, \textsl{wash}, \textsl{close}, \textsl{put down}. In our evaluation set, we seek instances where an object-verb composition has not been previously encountered during training; for example, the unique pairing of the \textsl{cut} with the \textsc{celery}.

\subsection{Forming the Compositional Splits}

We use the \emph{Maximum Compound Divergence} heuristic~\citep{keysers2019measuring} to curate a dataset that requires compositional generalization. The EK-100 dataset is annotated with 97 verb classes and 300 noun classes; these become the noun and verb \emph{atoms}. Each instance in the dataset is assigned to the training / validation / test split based on the atomic and compound divergence (similarity) based on weighted distributions using Chernoff coefficient $C_\alpha(P \Vert Q) = \sum_{k} p_k^\alpha \, q_k^{1-\alpha} \in [0, 1]$ ~\citep{chung1989measures}. To make atom distributions similar in train and test, we use $\alpha=0.5$ for atom divergence. Here, we set $\alpha=0.1$ to reflect that it is more important for a compound to be found in $P$ (train) rather than the probabilities in $P$ (train) and $Q$ (test) match exactly. Following this logic, we define compound divergence, and atom divergence for a train set $U$ and test set $W$ as follows:
\begin{align*}\label{eq:divergence}
    \mathcal{D}_C(U \Vert W) &= 1\,-\, C_{0.1}(\mathcal{F}_C(U) \, \Vert \, \mathcal{F}_C(W)) \\
    \mathcal{D}_A(U \Vert W) &= 1\,-\, C_{0.5}(\mathcal{F}_A(U) \, \Vert \, \mathcal{F}_A(W))
\end{align*}
\noindent where $\mathcal{F}_A(T)$ denotes frequency distribution of atoms, and $\mathcal{F}_C(T)$ denotes the distribution of compounds for a given set $T$ and $D_A$ and $D_C$ denote atom and compound divergences, respectively.
We calculated divergence scores for each instance until the atomic divergence of train and test set $D_A<0.02$ and compound divergence of train and test set $D_C$ $>$ $0.6$, which represents a sweet spot in terms of target distributions of atoms and compounds in the train and test sets (see Fig.~\ref{fig:atomcompound} in the Sec.~\ref{appendix:curating}). Finally, we randomly divide this test set into a validation and test set. The resulting dataset has 8,766 instances, which are split into 4,407 training, 2,184 validation, and 2,175 test instances (see Sec.~\ref{sec:app_data} for the implementation details and Sec.~\ref{sec:app_exploratory} for a more detailed analysis of the \textsc{CompAct} dataset).

\section{Multimodal Sequential Compositional Generalization}

Anticipating what comes next is a fundamental aspect of human cognition \citep{bar2007proactive, clark2015surfing}. From a cognitive perspective, it also serves as an engaging training paradigm~\citep{baroni2019linguistic}. In Multimodal Sequential Compositional Generalization, we seek to understand the extent to which multimodal foundation models are capable of understanding what comes next in activity sequences. We propose two tasks to measure multimodal sequential compositional generalization in the \textsc{CompAct} dataset: (i) next utterance prediction, and (ii) atom classification.

\subsection{Next Utterance Prediction Task}\label{sec:problem_formulation}

The next utterance prediction task is a language generation problem, in which models need to predict the text narration that describes the final input in a sequence. Let $\mathcal{S}=(\mathbf{X},\mathbf{V},\mathbf{A})$ denote a triplet representing a short video clip with $\mathbf{X}=\{\mathbf{x}_i\}_{i=1}^{K}$ being a sequence of $K$ utterances, which describe a household activity
and grounded with visual and audio signals, denoted by $\mathbf{V}=\{\mathbf{v}_i\}_{i=1}^{K}$ and $\mathbf{A}=\{\mathbf{a}_i\}_{i=1}^{K}$, respectively. This task involves generating the $(K+1)^{th}$ utterance, $\mathbf{y}=\mathbf{x}_{K+1}$,  following the preceding $K$ utterances and multimodal cues. 
The training data for this task consists of a set of sequences of microsegments,  $\{(\mathcal{S},\mathbf{y})\}$.
\subsection{Atom Classification Task}

The atom classification is a simplified form of the next utterance prediction task. Here, a model only needs to predict the verb and noun in the final input, in isolation from generating grammatically correct sentences. As such, it can be approached as a multi-class classification problem. 
Diverging from conventional action anticipation tasks~\citep{damen2020ek100, gammulle2019predicting,ke2019time}, our unique setup allows us to approach atom classification through a compositional lens, enabling the prediction of verbs and nouns separately. 
More formally, let $\mathcal{S}=(\mathbf{X},\mathbf{V},\mathbf{A})$ denote a triplet representing a video clip with $\mathbf{X}=\{\mathbf{x}_i\}_{i=1}^{K}$ representing a sequence of $K$ utterances, which describe a household activity
grounded with visual and audio signals, denoted by $\mathbf{V}=\{\mathbf{v}_i\}_{i=1}^{K}$ and $\mathbf{A}=\{\mathbf{a}_i\}_{i=1}^{K}$, respectively. More specifically, our atom classification task involves predicting the verb or noun in the $(K+1)^{th}$ utterance, $\mathbf{y}=\mathbf{x^{C}}_{K+1}$, following the preceding $K$ utterances and multimodal cues where $C$ denotes the verb or noun class.

\section{Models}\label{sec:fup_models}

In our experiments, we benchmark a variety of neural network models on the proposed next utterance prediction and atom classification tasks, including several text-only (unimodal) and multimodal models for better understanding the importance of different modalities in compositional generalization.

\subsection{Text-only Unimodal Baseline (L)} 

Our first baseline is a text-only model to account for unexpected biases in \textsc{CompAct}~\citep{thomason-etal-2019-shifting}. This is an encoder-decoder Transformer~\citep{vaswani2017attention} with a hidden size of 256 units, where each microsegment is encoded within its context. The model is trained using only the textual utterances $\mathbf{x}_{1:K}$ from the microsegment as the input, and the next utterance $\mathbf{x}_{K+1}$ as the target, \ie to predict $p\left(\mathbf{x}_{K+1}|\mathbf{x}_{1:K}\right)$. We use the same backbone in all of our multimodal baselines.

\begin{figure*}[!t]
\centering
\includegraphics[width=1\textwidth]{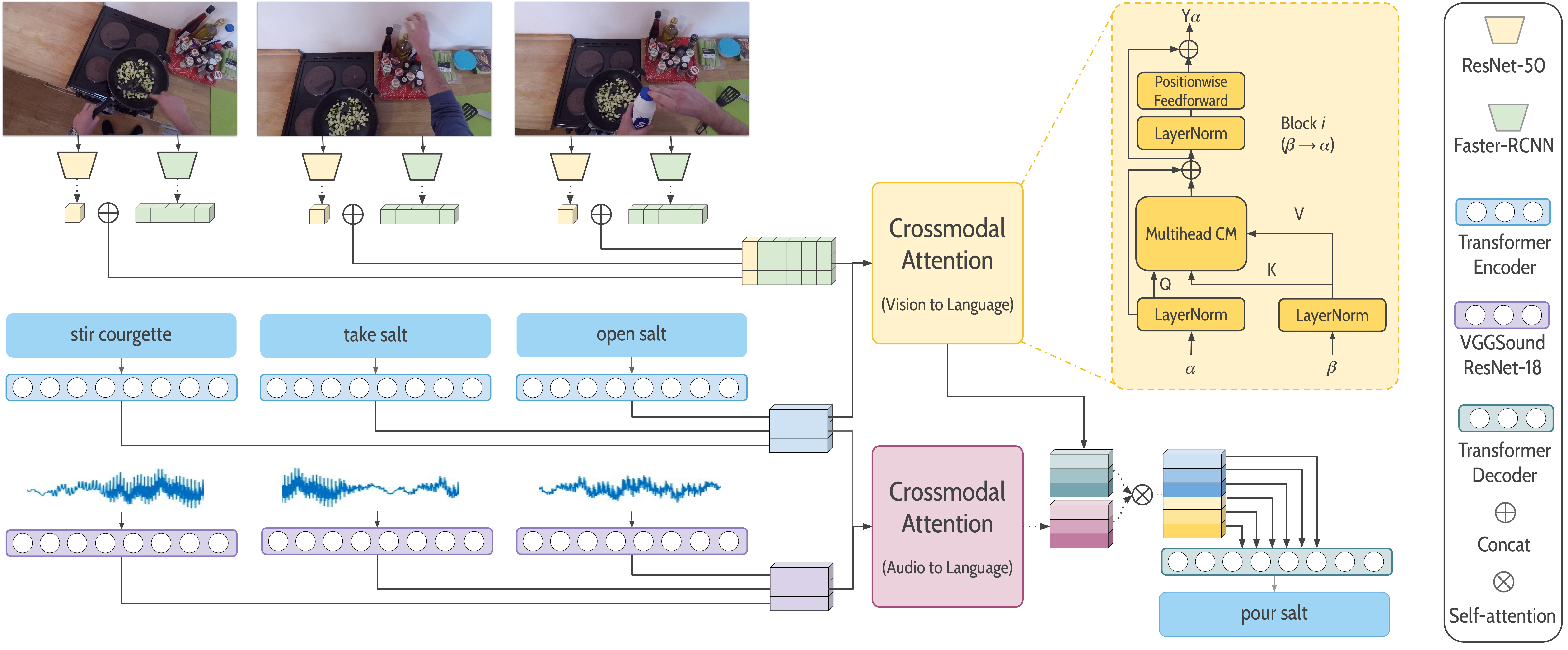}
\caption{Overview of the AVL baseline which integrates image, object-level, audio, and textual features utilizing two crossmodal attention blocks incorporated within an encoder-decoder Transformer to predict the next utterance.}
\label{fig:multimodal-baseline}
\end{figure*}

\subsection{Multimodal Baselines}

\paragraph{Vision and Language (VL):}\hspace{-1em} Our Vision and Language baseline encodes both textual and visual context for the next utterance prediction task. This model encodes the textual utterances $\mathbf{x}_{1:K}$ of each action from microsegments and the keyframe images $\mathbf{v}_{1:K}$ to predict the next utterance $\mathbf{x}_{K+1}$, \ie $p\left(\mathbf{y}=\mathbf{x}_{K+1}|\mathbf{x}_{1:K},\mathbf{v}_{1:K}\right)$. This model is adapted from a model that parses a visual scene and learns cross-modal self-attention \citep{tsai2019MULT} over textual inputs and visual data.
The visual inputs are encoded using pretrained CNN, and the textual inputs are encoded using a Transformer. More specifically, for the visual modality, we extract two types of features: one type represents global visual features, and the other represents object-level features. For the global features, we use a pretrained ResNet50 model~\citep{he2016deep} with ImageNet weights~\citep{russakovsky2015imagenet}. Object-level features are extracted using a Faster-RCNN object detector~\citep{ren2015faster} with a ResNet-101 backbone~\citep{he2016deep} which is pretrained on MSCOCO~\citep{lin2014microsoft} and finetuned on EK-100. We extract visual features from 5 objects for each keyframe. The resulting representation of a visual keyframe is the concatenation of the global and the object-level features. This concatenated vector is projected into a lower-dimensional space with a linear layer. The textual inputs are encoded using a Transformer with a 256D hidden layer. %
The visual and textual modalities are then encoded by a cross-modal (\text{CM}) self-attention mechanism. In this model, we consider two modalities $\alpha$ and $\beta$, sequences of each modalities are denoted as $X_\alpha \in \mathbb{R}^{T_\alpha \times d_\alpha}$ and $X_\beta \in \mathbb{R}^{T_\beta \times d_\beta}$, respectively and $T_{(\cdot)}$ denotes sequence length and $d_{(\cdot)}$ denotes feature dimension. In this model, $\alpha$ is the language modality, and $\beta$ is the visual modality. In the cross-modal attention, the textual features are the \emph{keys}, and the visual features are the \emph{queries} and \emph{values}, for aligning visual features to textual features.
Let the $\mathrm{Query}$ be defined as $Q_\alpha = X_\alpha W_{Q_\alpha}$, the $\mathrm{Key}$s as $K_\beta = X_\beta W_{K_\beta}$, and the $\mathrm{Value}$s as $V_\beta = X_\beta W_{V_\beta}$, where $W_{Q_\alpha} \in \mathbb{R}^{d_\alpha \times d_k}, W_{K_\beta} \in \mathbb{R}^{d_\beta \times d_k}$ and $W_{V_\beta} \in \mathbb{R}^{d_\beta \times d_v}$ are learnable weights. The cross-modal self-attention from $\beta$ to $\alpha$ is formulated as a latent adaptation $Y_\alpha \in \mathbb{R}^{T_\alpha \times d_v}$:
\begin{equation}
\label{eq:crossmodal_selfattention}
\resizebox{0.89\hsize}{!}{
$
\begin{aligned}
    Y_\alpha &= \text{CM}_{\beta \rightarrow \alpha}(X_\alpha, X_\beta) = \text{softmax}\left(\frac{Q_\alpha K_\beta^\top}{\sqrt{d_k}}\right) V_\beta
\end{aligned}
$}
\end{equation}
The output $Y_\alpha$ has the same length as $Q_\alpha$, but it is represented in the feature space of $V_\beta$. This enables the model to fuse different modalities, learning an alignment between the visual and textual features (see Eq.\ref{eq:crossmodal_selfattention}). There are different strategies proposed in the literature for modeling cross-modal interactions and fusing different modalities \citep{xu2023multimodal}. In our vision and language baseline, we fuse different modalities via a self-attention layer over the aligned vision and language features, which are then fed to a 3-layer Transformer decoder with 4 attention heads that generate the next utterances.

\paragraph{Audio and Language (AL):}\hspace{-1em} The Audio and Language baseline has the same structure as the VL baseline. The key difference is that we represent the additional context using audio features instead of visual features. The model encodes both the textual utterances $\mathbf{x}_{1:K}$ and the accompanying audio data $\mathbf{a}_{1:K}$ to predict the next utterance $\mathbf{x}_{K+1}$, \ie $p\left(\mathbf{x}_{K+1}|\mathbf{x}_{1:K},\mathbf{a}_{1:K}\right)$. The audio features are 512D vectors extracted using VGGSound~\citep{chen2020vggsound}, which is pretrained on 200K videos from YouTube, totaling 550 hours of audio data. Here, the proposed AL model learns cross-modal attention over audio and textual features, analogously to the VL model, as inputs to a Transformer decoder.

\paragraph{Object and Language (OL):}\hspace{-1em} The Object and Language baseline once again uses the same architecture as VL baseline, but the visual context is represented using the labels of detected objects instead of continuous visual features. In this model, we embed object tags as a secondary set of textual features to our model along with the input utterances. Here, the object tags are represented as 292D one-hot encoded vectors (based on the number of unique tags) and projected to 256D with a linear layer. In this case, the cross-modal attention aligns the object tag features with the language features.

\paragraph{Audio, Vision, and Language (AVL):}\hspace{-1em} In the AVL baseline, we leverage the audio, visual, and textual data using two cross-modal self-attention blocks. We use textual utterances $\mathbf{x}_{1:K}$ of each action along with the visual features $\mathbf{v}_{1:K}$ from the keyframes, and the VGGSound audio features $\mathbf{a}_{1:K}$ to predict the next utterance $\mathbf{x}_{K+1}$, \ie $p\left(\mathbf{y}=\mathbf{x}_{K+1}|\mathbf{x}_{1:K},\mathbf{v}_{1:K},\mathbf{a}_{1:K}\right)$. In this model, the input to the Transformer decoder is the concatenation of the audio-aligned textual features from the audio-textual cross-modal block with the visual-aligned textual features from the visual-textual cross-modal block (see Fig.~\ref{fig:multimodal-baseline} for an overview).

\paragraph{Object, Audio and Language (OAL):}\hspace{-1em}
The OAL baseline model adds an extra modality to the OL baseline model to determine whether audio features affect the performance of a model that uses object tags. Here, we incorporate the extracted audio features from each microsegment into the OL model.

\subsection{Pretrained Models}
To comprehend the significance of large-scale pretraining, we conduct an extensive evaluation involving several publicly available models, namely LLaMA2~\citep{Touvron2023Llama2O}, IDEFICS~\citep{Laurenccon2023OBELISCAO}, MERLOT Reserve~\citep[MerlotR]{zellers2022merlot}, and ImageBind~\citep{girdhar2023imagebind}. In our setup, we investigate the performance of encoder-only models across both tasks, whereas auto-regressive models are evaluated exclusively through prompting within the context of the next utterance prediction task. It is worth noting that interpreting the performance of the pretrained models can be complicated as they may violate the distributional consistency between the train and test splits during their pretraining \citep{kim2022uncontrolled}.

\paragraph{Unimodal Models:}\hspace{-1em} 
LLaMA2 is a text-only pretrained large language model trained on 500B tokens. We evaluate the LLaMA2-Chat 6.7B variant, as this version results in more coherent and relevant predictions due to instruction tuning and RLHF.

\paragraph{Multimodal Models:}\hspace{-1em} 
\underline{MerlotR} learns to extract representations over video frames, text, and audio. The model is composed of an image encoder, an audio encoder, and a joint encoder that fuses textual, visual, and audio representations. This model employs contrastive span training, where an aligned span of audio and text is masked. In its training setup, the objective is to maximize representation similarity to an independent encoding of the masked audio and text spans. We extract multimodal audio and vision features through its pretrained encoder utilizing a similar backbone as in the VL model.
\underline{ImageBind} is a multimodal model that learns joint embeddings for 6 different modalities, including language, vision, and audio. It is trained only on image-paired data to bind the modalities together. 
We train a decoder using features extracted from the vision, language, and audio modalities.
\underline{IDEFICS} is a large-scale multimodal large language model based on Flamingo~\citep{alayrac2022flamingo} architecture. It is composed of a frozen language model and a frozen vision encoder with learnable cross-attention blocks connecting language and vision modalities. Considering that Flamingo is not publicly available and IDEFICS performs better than other open-source Flamingo implementations such as OpenFlamingo~\citep{Awadalla2023OpenFlamingoAO}, we experiment with IDEFICS 9B version as the vision LLM.
We prompt these models without any finetuning and report 5-shot results for LLaMA2 and IDEFICS (see Sec. \ref{sec:app_prompting_prep} for prompting formats and Sec. \ref{sec:app_prompting_abl} for prompting ablations).

\subsection{Task-Specific Changes}
For atom classification, we modify the previously described models by altering their architectures. Specifically, we replace the decoder Transformer with two fully connected layers. We train these models with a classification objective conditioned on predicting verbs or nouns in atom classification.

\begin{figure*}[ht]
    \centering
    \resizebox{1\linewidth}{!}{%
        \begin{tabular}{cl}
            \toprule
            \textbf{Inputs (utterances and auxiliary modalities)} & \textbf{Prediction (next utterance)} \\ \midrule    
                        \begin{tabular}{@{}c@{$\;$}c@{$\;$}c@{$\;$}c@{}}
                \centering   
            & \adjincludegraphics[valign=M,width=0.27\linewidth]{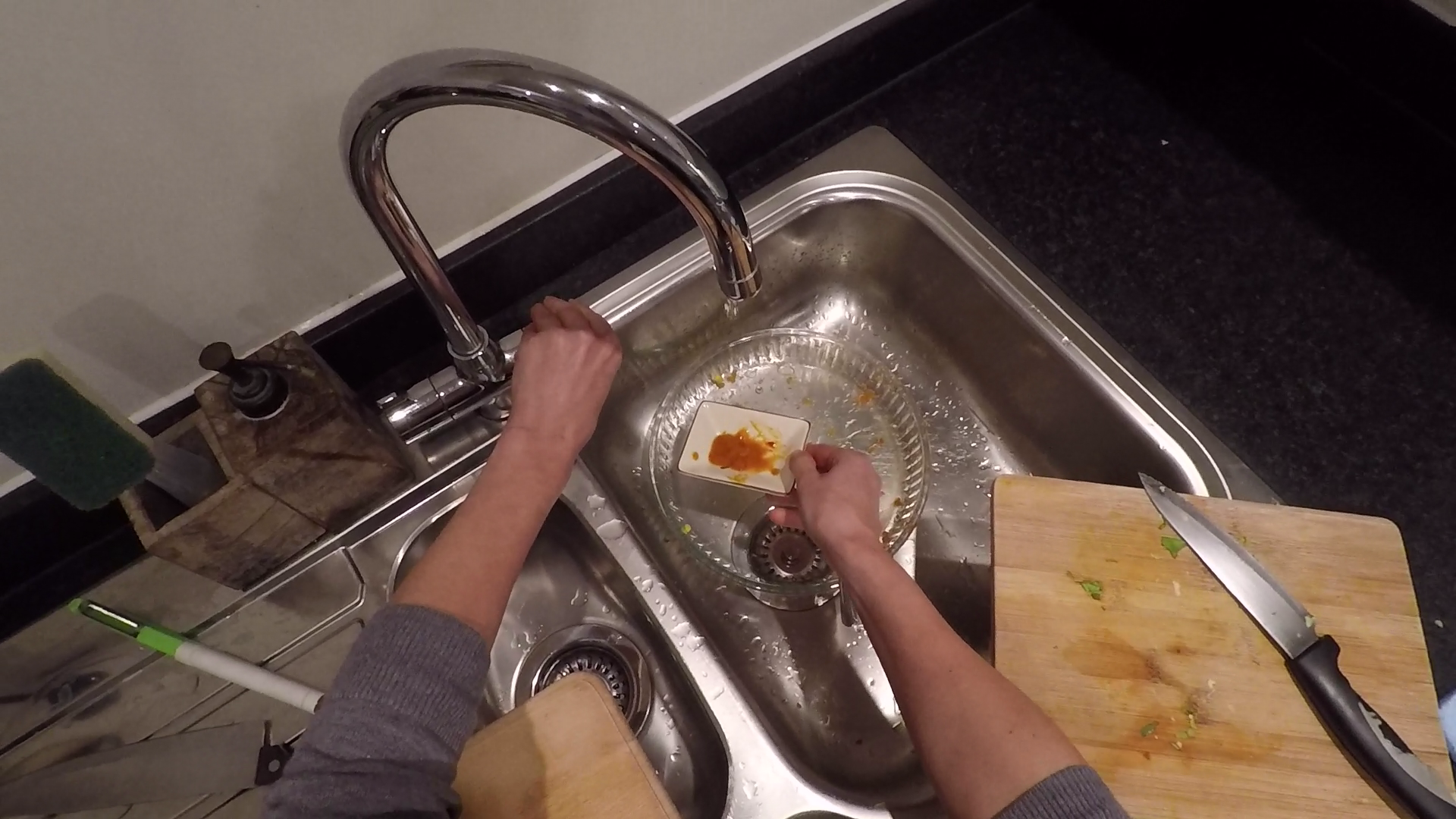} & \adjincludegraphics[valign=M,width=0.27\linewidth]{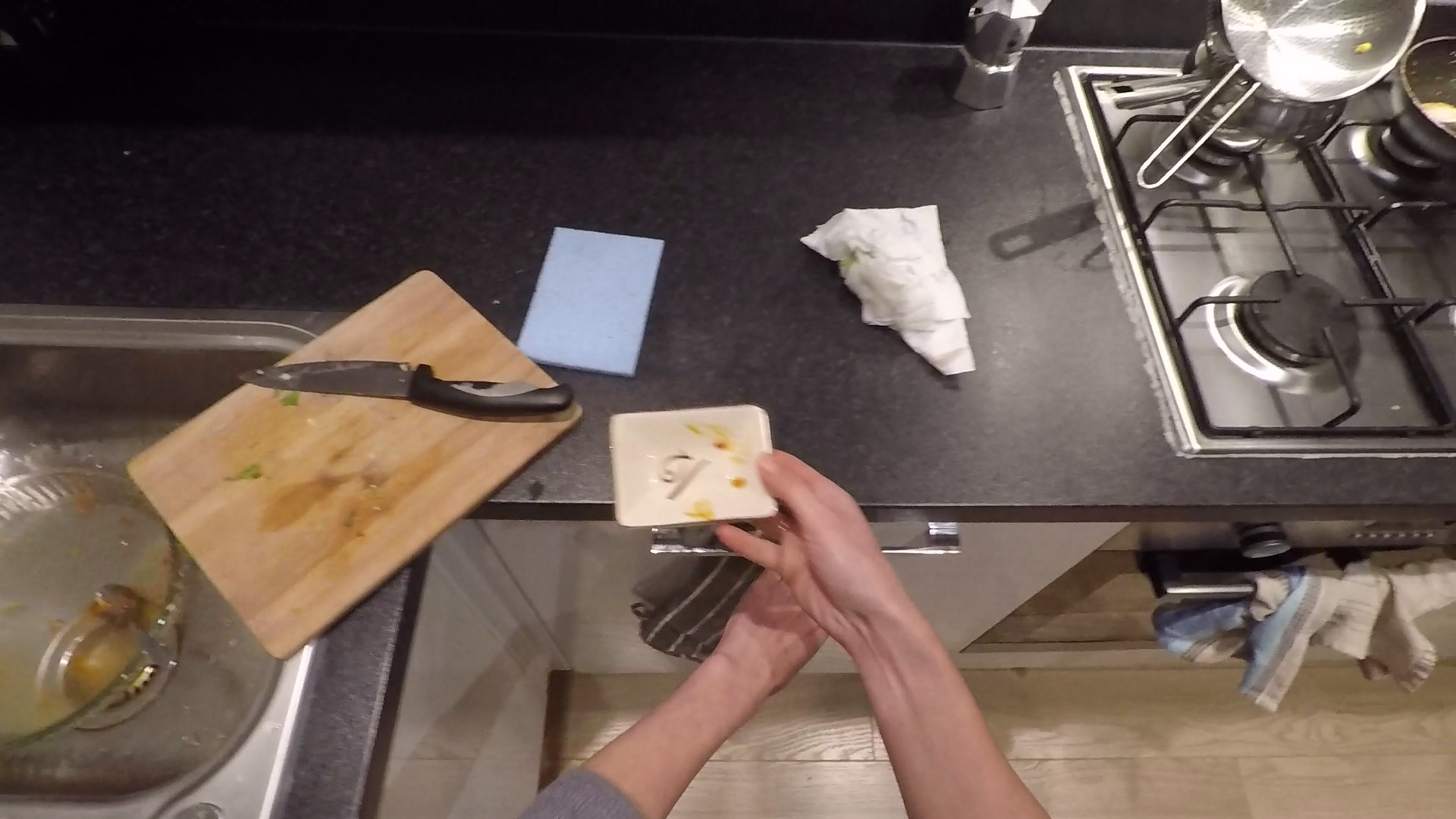} & \adjincludegraphics[valign=M,width=0.27\linewidth]{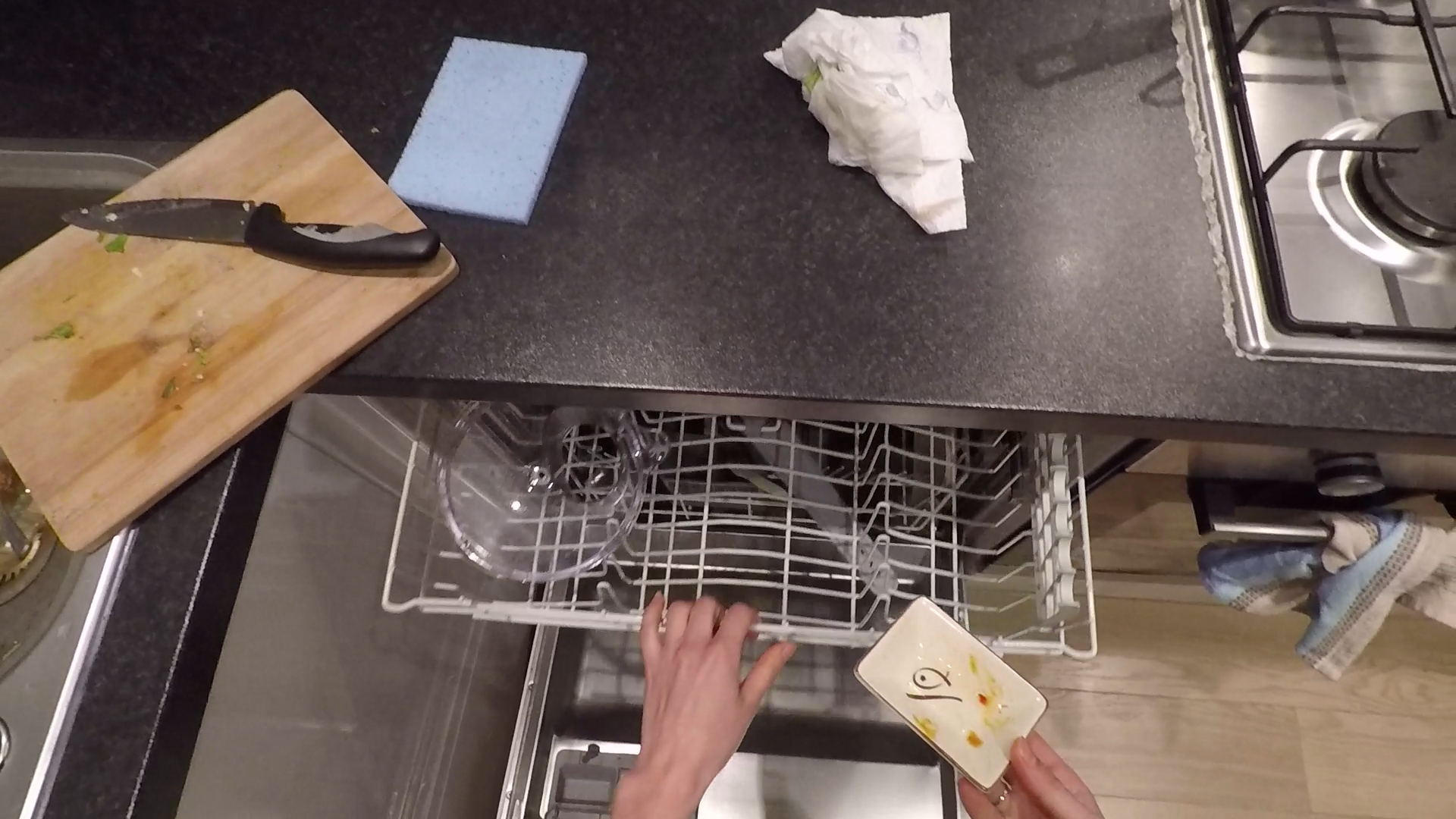}  
            \vspace{0.5em}
            \\
            
            & clean bowl  & open dishwasher  & open drawer \\
              
            \end{tabular}
            & 
            \bgroup
            \def\arraystretch{0.8}
            \resizebox{0.3\linewidth}{!}{
            \begin{tabular}{ll}
            GT &: place bowl \\
            L &: \orange{close dishwasher} \\
            OL &: \blue{place bowl} \\
            VL &: \orange{close dishwasher} \\
            AL &: \orange{close drawer} \\
            AVL &: \orange{dry bowl} \\
            OAL &: \blue{place bowl} \\
            MerlotR &: \orange{close dishwasher} \\
            ImageBind &: \purple{put bowl in dishwasher} \\
            LLaMA2 &: \orange{get clean} \\
            IDEFICS &: \purple{put bowl away.} \\
            \end{tabular}}
            \egroup \\

            \midrule

\begin{tabular}{@{}c@{$\;$}c@{$\;$}c@{$\;$}c@{}}
                \centering   
             & \adjincludegraphics[valign=M,width=0.27\linewidth]{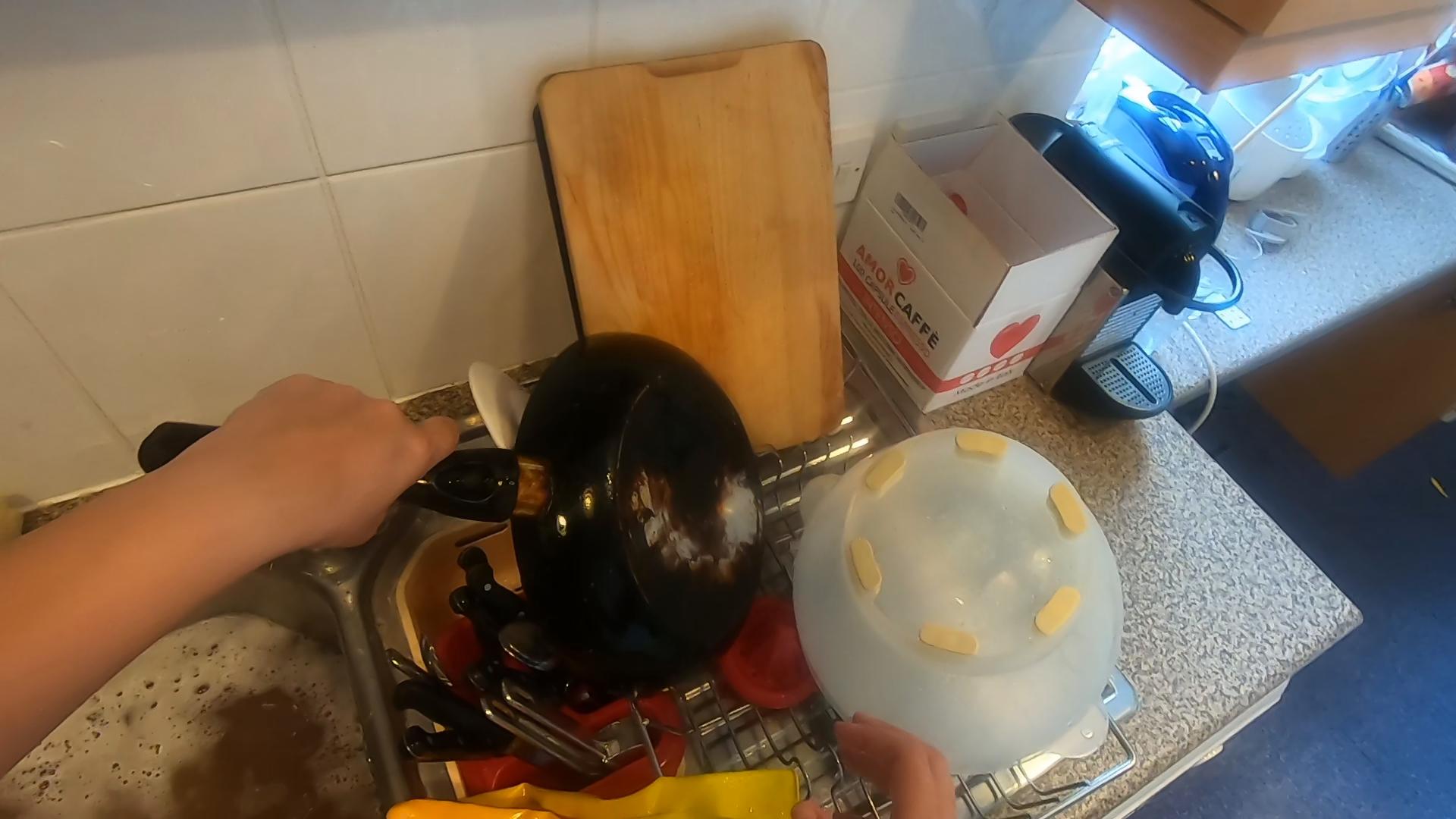} & \adjincludegraphics[valign=M,width=0.27\linewidth]{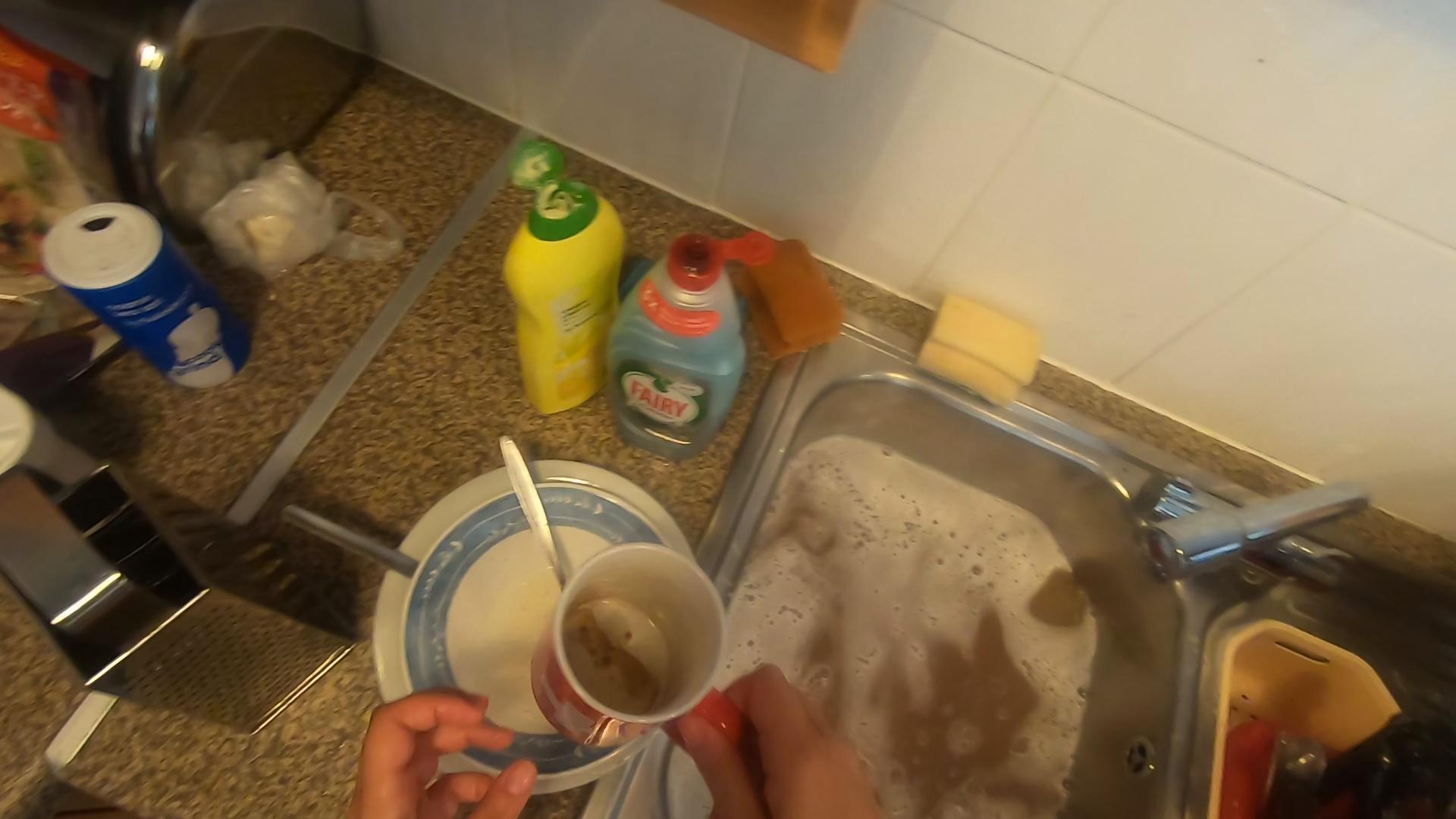} & \adjincludegraphics[valign=M,width=0.27\linewidth]{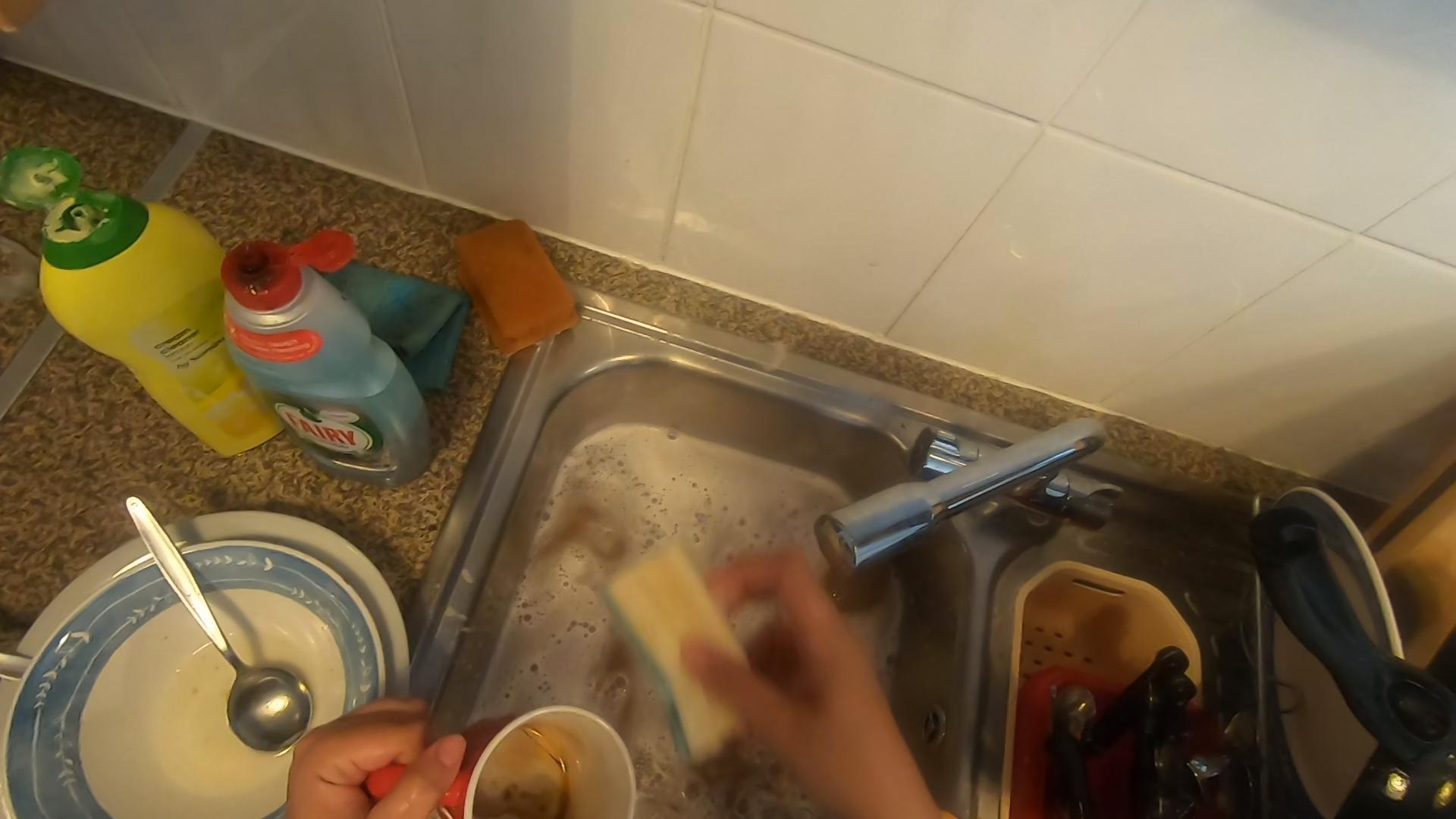}  
             \vspace{0.5em}
             \\
             & put pan in drainer  & pick\_up mug  & pick\_up sponge  \\
              
            \end{tabular}
            & 
            \bgroup
            \def\arraystretch{0.8}
            \resizebox{0.26\linewidth}{!}{
            \begin{tabular}{ll}
            GT &: sponge mug \\
            L &: \orange{put sponge} \\
            OL &: \blue{sponge mug} \\
            VL &: \blue{sponge mug} \\
            AL &: \blue{sponge mug} \\
            AVL &: \blue{sponge mug} \\
            OAL &: \blue{sponge mug} \\
            MerlotR &: \blue{sponge mug} \\
            ImageBind &: \blue{sponge mug} \\
            LLaMA2 &: \orange{put sponge in sink} \\
            IDEFICS &: \blue{sponge mug} \\
            \end{tabular}}
            \egroup \\

                            \midrule
            
\begin{tabular}{@{}c@{$\;$}c@{$\;$}c@{$\;$}c@{}}
                \centering   
             & \adjincludegraphics[valign=M,width=0.27\linewidth]{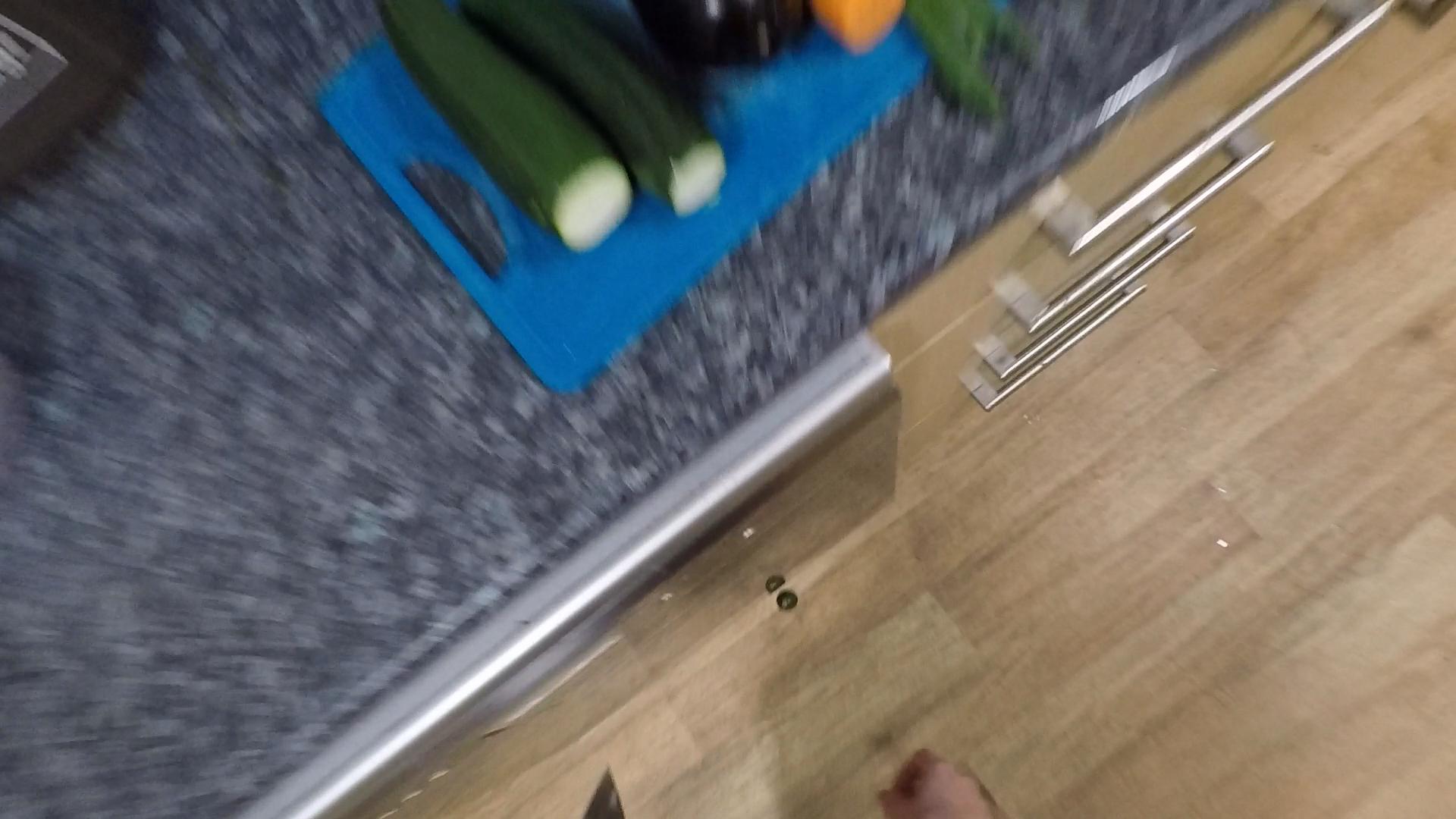} & \adjincludegraphics[valign=M,width=0.27\linewidth]{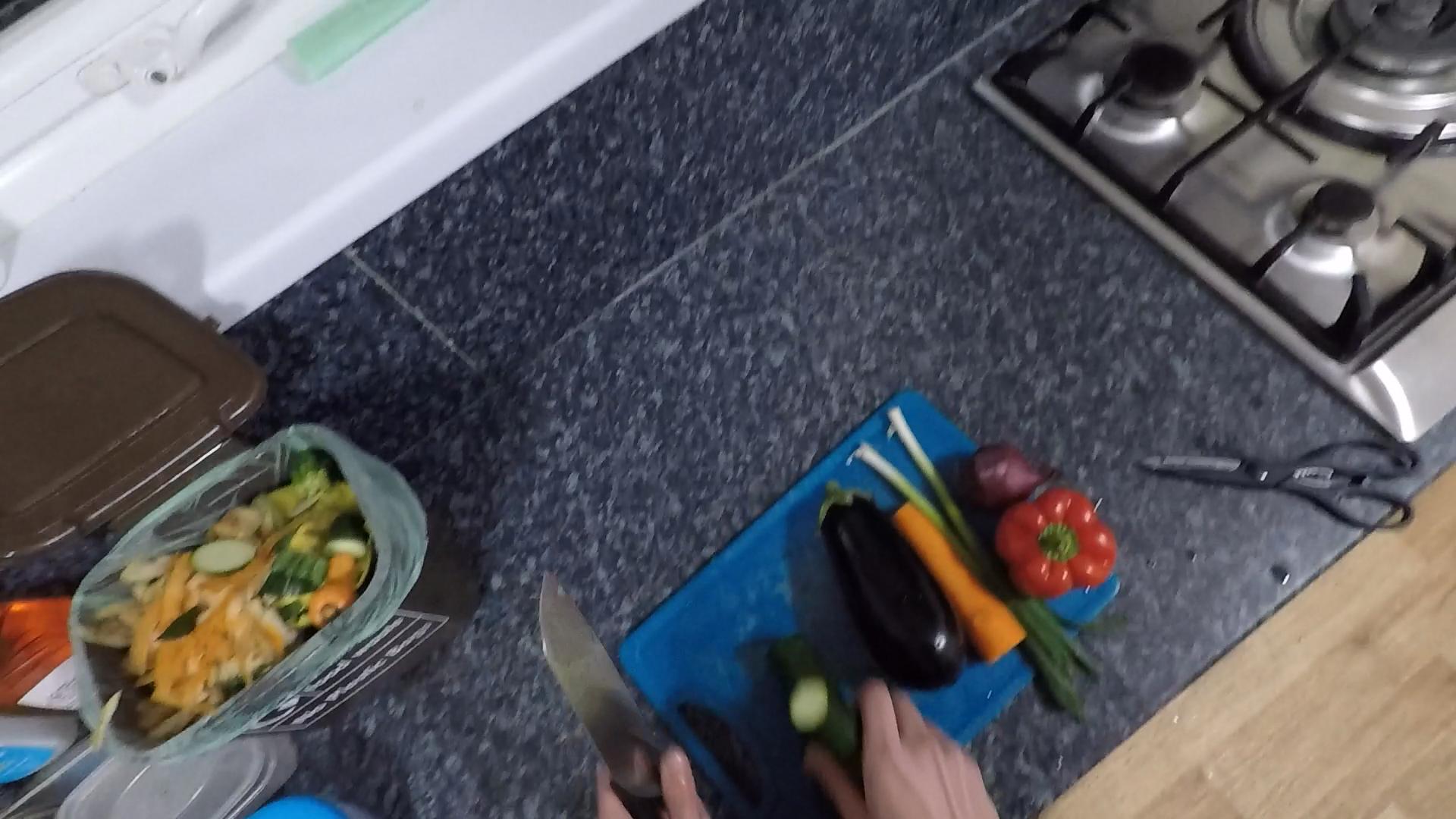} & \adjincludegraphics[valign=M,width=0.27\linewidth]{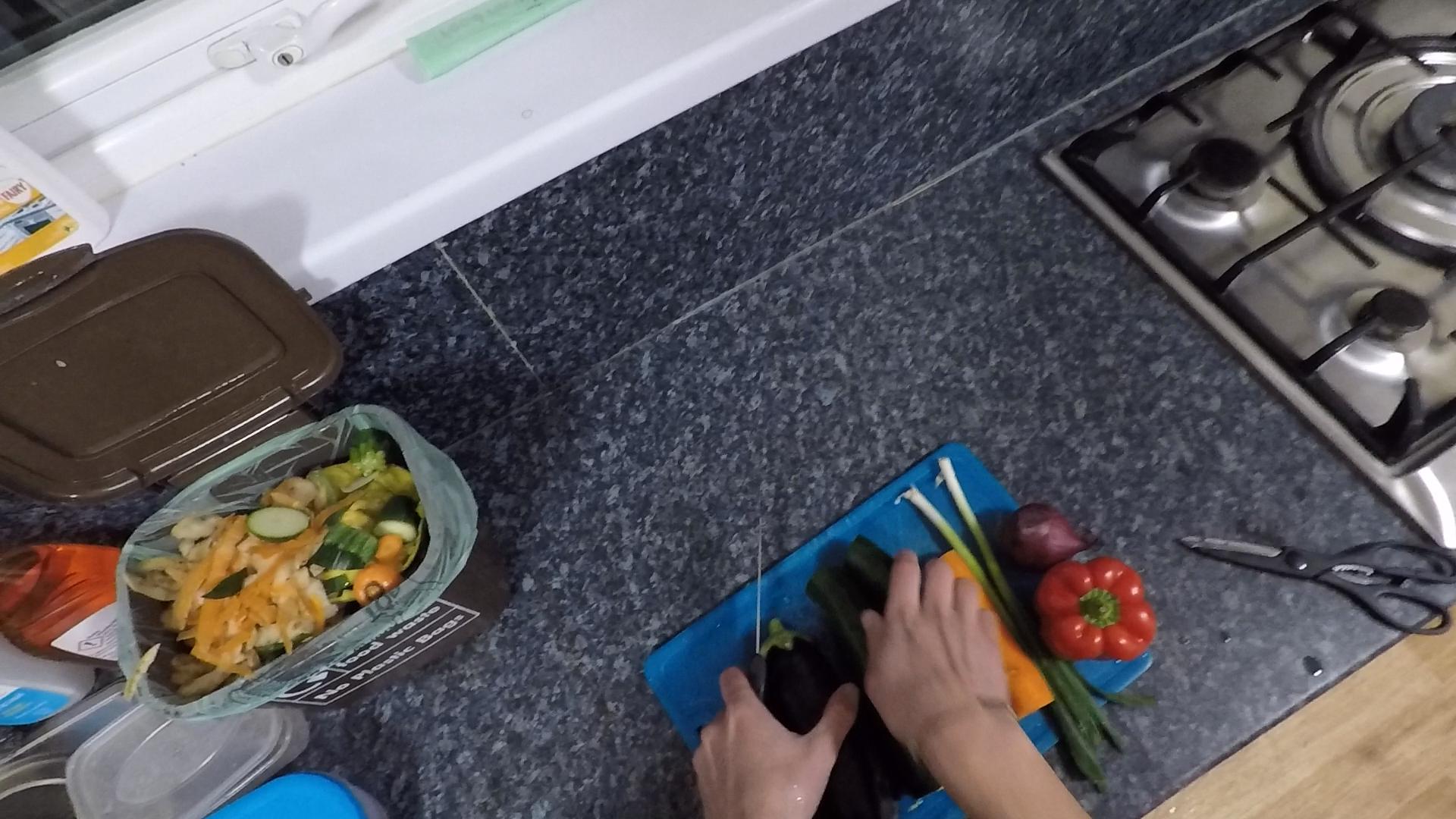}  
             \vspace{0.5em}
             \\
             & put\_down spring\_onions  & take courgettes  & take pepper  \\
              
            \end{tabular}
            & 
            \bgroup
            \def\arraystretch{0.8}
            \resizebox{0.3\linewidth}{!}{
            \begin{tabular}{ll}
            GT &: cut pepper \\
            L &: \orange{cut spring\_onions} \\
            OL &: \orange{put\_down courgette} \\
            VL &: \orange{put\_down spring\_onions} \\
            AL &: \orange{put pepper} \\
            AVL &: \blue{cut pepper} \\
            OAL &: \orange{put\_down courgette} \\
            MerlotR &: \orange{open pepper} \\
            ImageBind &: \blue{cut pepper} \\
            LLaMA2 &: \orange{take spring onions} \\
            IDEFICS &: \orange{put down knife} \\
            \end{tabular}}
            \egroup \\
            
            \bottomrule
            \end{tabular}

    }
    \caption{{Next Utterance Prediction qualitative results.} 
    Models consider different combinations of input modality, as described in Section~\ref{sec:fup_models}. 
    In the predictions, \blue{blue} refers to correct, \orange{orange} incorrect and \purple{purple} semantically close.} 
    \label{fig:qualitative}
\end{figure*}

\section{Experimental Setup}\label{sec:experiments}

\noindent\textbf{Evaluation Metrics:}
We use unigram BLEU~\citep{papineni2002bleu}, Exact Match (EM), Categorical Accuracy (CA) and BERTScore \citep{zhang2019bertscore} metrics. The reported values represent the mean and standard deviation across 3 separate runs. In LLaMA2 and IDEFICS, we use nucleus sampling instead of separate runs. For EM, we calculate the accuracy between the generated text sequence and the ground truth. CA uses the verb and noun categories in EK-100 and calculates the accuracy based on category match between the prediction and ground truth, \eg the verbs \textsl{slice}, \textsl{dice}, and \textsl{chop} fall into the same verb category \textsl{cut}, and the nouns \textsc{cheddar}, \textsc{paneer} and \textsc{parmesan} are grouped into the same noun category \textsc{cheese}. Therefore, the \textsl{slice} \textsc{paneer} prediction is deemed accurate if the ground truth is \textsl{dice} \textsc{parmesan}.

\noindent\textbf{Training Procedure:}
In the next utterance prediction task, models are trained to minimize the negative log-likelihood of generating the next utterance, where the multimodal models are conditioned on additional modalities. Given microsegment $\mathcal{S}$ and model parameters $\theta$, the objective function is to minimize the negative log-likelihood of the \textit{m} tokens in the next utterance: $L(\theta) = -\sum\nolimits_{i=1}^{m} \log p(y_i |\mathcal{S}; \theta) \label{sentence objective}$. 
In the atom classification task, models are trained by attaching an MLP with a multi-class classification layer to the encoding of a microsegment $\mathcal{S}$. The objective function is to minimize the cross-entropy loss of predicting the expected atom (verb or noun): $L(\theta) = -\log p(\mathbf{x}_{K+1}^{C} |\mathcal{S}; \theta)$ (see Sec.~\ref{sec:app_implementation} for details).

\section{Results}\label{sec:results}
\subsection{Next Utterance Prediction}

\begin{table}[h]
\caption{Next Utterance Prediction results on the test split. Using audio, visual, or object features consistently improves performance compared to the language-only baseline. The best results are bolded, while the second-best results are underlined. Here, ImageB. and BERTSc. refer to ImageBind and BERTScore, respectively.}
\centering

\resizebox{\linewidth}{!}{%
\begin{tabular}{lcccc}
\toprule
Inputs & BLEU & EM & CA & BERTSc. \\
\midrule
L & 21.75\sta{}{1.0} & 2.89\sta{}{0.3} & 6.43\sta{}{0.2} & 79.06\sta{}{0.1} \\
VL & 31.25\sta{}{0.3} & 7.27\sta{}{0.1} & 12.95\sta{}{0.4} & 81.27\sta{}{0.1} \\
AL & 30.82\sta{}{0.5} & 6.81\sta{}{0.5} & \underline{13.22\sta{}{0.9}} & 81.20\sta{}{0.0} \\
AVL & 31.73\sta{}{0.4} & 7.04\sta{}{0.4} & 12.93\sta{}{0.8} & 81.50\sta{}{0.1} \\
OL & 30.79\sta{}{0.6} & 6.36\sta{}{0.2} & 12.21\sta{}{0.1} & 81.23\sta{}{0.1} \\
OAL & \underline{32.02\sta{}{0.2}} & \underline{7.32\sta{}{0.6}} & 13.08\sta{}{0.9} & \underline{81.51\sta{}{0.1}} \\ \midrule
MerlotR & 31.50\sta{}{0.3} & 6.75\sta{}{0.2} & 12.85\sta{}{0.1} & 81.37\sta{}{0.2} \\
ImageB. & \textbf{33.52\sta{}{0.3}} & \textbf{9.45\sta{}{0.5}} & \textbf{15.04\sta{}{1.0}} & \textbf{82.31\sta{}{0.2}} \\
IDEFICS & 25.64\sta{}{0.4} & 5.76\sta{}{0.1} & 7.89\sta{}{0.5} & 80.92\sta{}{0.1} \\
LLaMA2 & 27.50\sta{}{0.6} & 5.36\sta{}{0.6} & 7.41\sta{}{0.7} & 78.76\sta{}{0.2} \\
\bottomrule
\end{tabular}}

\label{tbl:results_fup_quantitative_2}
\end{table}

\noindent In Table~\ref{tbl:results_fup_quantitative_2}, we present the results of the next utterance prediction experiments. Notably, all multimodal models surpass the language-only baseline. Our baseline model that incorporates visual features (VL) exhibits consistent increases, showing gains of up to 9 BLEU, 4 EM, 6 CA, and 1 BERTScore points, compared to the language-only variant. Furthermore, harnessing a mix of audio, visual, and language features (AVL) or augmenting audio features with object tags (OAL) leads to additional improvements, emphasizing the contribution of fusing multiple modalities. The most significant boost in performance is observed when visual features are utilized in the ImageBind pretrained model, resulting in approximate increases of 11, 6, 8, and 2 points, for BLEU, EM, CA, and BERTScore metrics, respectively. The fact that LLaMA2 generates utterances with higher BLEU but lower BERTScore than IDEFICS might suggest that LLaMA2 better imitates the required vocabulary than IDEFICS, even though IDEFICS produces more semantically plausible outputs. Conclusively, ImageBind's performance shows the advantages of employing pretrained multimodal features over merely merging separate unimodal encodings.
In Fig.~\ref{fig:qualitative}, we present a qualitative comparison of the baseline models via randomly selected examples from the test set. We believe that these illustrative examples effectively showcase the intricate and challenging nature of the proposed \textsc{CompAct} dataset. During training, the models have never encountered compounds like \textsl{place} \textsc{bowl}, or \textsl{cut} \textsc{pepper}. In all of these illustrative examples, the text-only unimodal model fails to generalize to these novel compositions. However, in the first example, the OL and OAL baselines can predict the target composition correctly. ImageBind and IDEFICS, even though not exact matches, generate semantically plausible predictions. In the third example, all multimodal models correctly predict the next utterance by leveraging the auxiliary modalities. Note that, LLaMa2 also fails in this example whereas IDEFICS can generate correct utterances. In the fourth example, the AVL model and ImageBind models correctly predict the \textsl{cut} \textsc{pepper} utterances. Interestingly, for this example both audio and vision inputs are needed, indicating that for sequential compositional generalization, models might have to leverage the available signal coming from different modalities at the same time.

\subsection{Atom Classification}

\begin{table}[h]
\caption{{Quantitative results for Atom Classification}. The best and the second-best performing results are highlighted in bold and underlined, respectively.} \label{tbl:results_cls_quantitative_combined}
\resizebox{\linewidth}{!}{%
\begin{tabular}{@{}llccc}
    \toprule
    \multirow{12}{*}{\rotatebox{90}{Verb Classification $\;$}} & & EM & CA & BERTScore \\ \midrule
    & L & 13.37\sta{}{0.5} & 28.47\sta{}{3.1} & 75.16\sta{}{0.6} \\
    & VL & 14.02\sta{}{0.2} & 28.68\sta{}{2.3} & 75.29\sta{}{0.5} \\
    & AL & 13.76\sta{}{0.3} & 30.05\sta{}{4.6} & \underline{76.26\sta{}{0.5}} \\
    & AVL & \underline{14.06\sta{}{0.7}} & 30.98\sta{}{2.1} & 76.12\sta{}{0.7} \\
    & OL & 12.79\sta{}{0.1} & 29.97\sta{}{1.5} & 75.66\sta{}{0.2} \\
    & OAL & 13.91\sta{}{0.5} & 29.90\sta{}{1.3} & 75.97\sta{}{0.9} \\  \noalign{\vspace{0.25ex}}\cline{2-5}\noalign{\vspace{0.25ex}}
    & MerlotR & 13.71\sta{}{0.1} & \textbf{33.50\sta{}{1.8}} & 76.07\sta{}{0.2} \\
    & ImageBind & \textbf{15.40\sta{}{0.2}} & \underline{31.54\sta{}{2.5}} & \textbf{76.54\sta{}{0.4}} \\ \noalign{\vspace{0.25ex}}\cline{2-5}\noalign{\vspace{0.25ex}}
    & MRH & 2.39 & 9.61 & 73.60 \\
    \midrule
    \multirow{9}{*}{\rotatebox{90}{Noun Classification $\;$}} & L & 44.91\sta{}{0.3} & 51.83\sta{}{0.3} & 86.27\sta{}{0.2} \\
    & VL & 42.72\sta{}{0.7} & 49.57\sta{}{0.3} & 85.81\sta{}{0.2} \\
    & AL & 43.95\sta{}{0.2} & 51.11\sta{}{0.5} & 86.08\sta{}{0.1} \\
    & AVL & 43.34\sta{}{0.4} & 50.43\sta{}{0.9} & 85.92\sta{}{0.1} \\
    & OL & 44.35\sta{}{0.9} & 51.24\sta{}{0.8} & 86.00\sta{}{0.3} \\
    & OAL & 43.83\sta{}{0.9} & 51.03\sta{}{0.5} & 85.92\sta{}{0.2} \\ \noalign{\vspace{0.25ex}}\cline{2-5}\noalign{\vspace{0.25ex}}
    & MerlotR & \underline{45.42\sta{}{0.6}} & \underline{52.24\sta{}{0.7}} & \underline{86.42\sta{}{0.1}} \\
    & ImageBind & 33.67\sta{}{0.3} & 44.55\sta{}{0.1} & 83.96\sta{}{0.1} \\ \noalign{\vspace{0.25ex}}\cline{2-5}\noalign{\vspace{0.25ex}}
    & MRH & \textbf{57.24} & \textbf{61.15} & \textbf{89.75} \\ \bottomrule
\end{tabular}}

\end{table}

In Table~\ref{tbl:results_cls_quantitative_combined}, we present the outcomes of our atom classification task, which seeks to understand models' abilities to predict verb and noun atoms in isolation. 

For predicting verbs, we observe a similar trend in performance with the next utterance prediction results. However, all models perform poorly in predicting nouns compared to the MRH baseline (Most Recent Heuristic). This baseline employs the most recently referenced object in the input microsegment as a prediction for the target noun, and the most recently referenced verb as a prediction for the target verb. 
While the language-only baseline outperforms the multimodal baselines in noun prediction, we observe an improvement over the language-only model in predicting verbs within the multimodal models. This subtle, yet noteworthy improvement underlines the value of leveraging multiple modalities for verb-related predictions.

\subsection{Random Split Experiments}

The \textsc{CompAct} dataset, designed to highlight out-of-distribution characteristics, features distinct compositional distributions between training and testing splits, as delineated in Figure \ref{fig:atomcompound}. This intentional design underscores the out-of-distribution characteristics essential for evaluating the models' generalization abilities.
We expand our analysis to include both in-domain and out-of-domain data, enabling a more comprehensive evaluation of model capabilities and their generalization potential. This analysis aims to compare baseline models on two distinct datasets: our original dataset, which emphasizes compositionality (out-domain), and a new, non-compositional dataset (in-domain) created through random splits of the EK-100 dataset.

\begin{table}[ht!]\caption{In-domain Next Utterance Prediction Results}
\centering
\resizebox{\linewidth}{!}{%
\begin{tabular}{@{}lcccc@{}}
\toprule
          & BLEU  & EM    & CA    & BERTScore \\ \midrule
L         & 23.37 & 8.95  & 12.01 & 80.09     \\
VL        & 36.49 & 18.74 & 22.92 & 82.83     \\
AL        & 37.04 & 19.81 & 23.86 & 83.07     \\
AVL       & 36.08 & 17.88 & 21.56 & 82.84     \\
OL        & 36.62 & 19.87 & 24.22 & 83.11     \\
OAL       & 38.59 & 21.86 & 25.54 & 83.64     \\
MerlotR   & 38.53 & 21.40 & 25.82 & 83.64     \\
ImageBind & 40.86 & 22.62 & 26.86 & 84.28     \\
IDEFICS   & 35.09 & 17.43 & 19.11 & 83.61     \\
LLaMA     & 30.20 & 12.10 & 14.09 & 79.90      \\ \bottomrule
\end{tabular}
}
\label{tbl:rnd_vs_sys_nup}
\end{table}

\begin{table}[ht!]\caption{In-domain Verb Classification Results}
\centering
\begin{tabular}{@{}lccc@{}}
\toprule
          & EM    & CA    & BERTScore \\ \midrule
L         & 20.26 & 33.90 & 77.65     \\
VL        & 21.90 & 33.19 & 77.65     \\
AL        & 21.44 & 33.73 & 77.88     \\
AVL       & 22.19 & 34.57 & 78.17     \\
OL        & 20.04 & 33.77 & 77.51     \\
OAL       & 21.39 & 33.21 & 77.55     \\
MerlotR   & 21.50 & 34.39 & 77.86     \\
ImageBind & 21.48 & 34.00 & 77.77     \\ \bottomrule
\end{tabular}
\label{tbl:rnd_vs_sys_verb}
\end{table}

\begin{table}[ht!]\caption{In-domain Noun Classification Results}
\centering
\begin{tabular}{@{}lccc@{}}
\toprule
          & EM    & CA    & BERTScore \\ \midrule
L         & 48.42 & 53.45 & 86.83     \\
VL        & 49.08 & 54.51 & 87.11     \\
AL        & 47.75 & 53.67 & 86.76     \\
AVL       & 47.47 & 53.05 & 86.57     \\
OL        & 48.55 & 54.02 & 86.86     \\
OAL       & 48.99 & 54.20 & 87.02     \\
MerlotR   & 48.07 & 53.30 & 86.72     \\
ImageBind & 44.45 & 52.19 & 86.26     \\ \bottomrule
\end{tabular}
\label{tbl:rnd_vs_sys_noun}
\end{table}

Our in-depth analysis, reflected in Table \ref{tbl:rnd_vs_sys_nup}, indicates a marked improvement in model performance when dealing with in-domain data. %

We extend our analysis for the next utterance prediction task in Table \ref{tbl:rnd_vs_sys_nup}, to the atom classification task (see Table \ref{tbl:rnd_vs_sys_verb} and Table \ref{tbl:rnd_vs_sys_noun}). The results from these analyses consistently show a significant performance improvement for models in the in-domain (non-compositional) setup compared to the compositionally challenging split. This highlights the added complexity and challenge introduced by compositionality, which is not present in standard atomic classification tasks. The results specifically illustrate that, despite models being trained with examples that include every individual primitive (nouns and verbs), their performance decline when tasked with generalizing to novel compositions. This picture does not change even when most similar examples are used as demonstrations within an in-context learning setup. This decline in performance is not merely due to unfamiliarity with certain primitives but stems from the inherent challenge of understanding and predicting new combinations of these primitives.
This divergence in performance between in-domain (random split) and out-of-domain data demonstrates the unique contribution of our work. It underscores the challenge we introduce: pushing the boundaries of current models to not only recognize but also effectively generalize and adapt to novel compositionsal structures. %

\section{Related Work}\label{sec:related}

\noindent\textbf{Compositionality.} \citet{baroni2019linguistic} studied the linguistic generalization capabilities of artificial neural networks. \citet{lake2019human} explored compositionality in a human-like few-shot setting, while others have studied compositionality at the representation level such as \citep{dasgupta2018evaluating, ettinger2018assessing}. Unimodal compositional generalization datasets such as SCAN~\citep{scandataset}, CFQ~\citep{keysers2019measuring}, and COGS~\citep{cogsdataset} have been widely used in the literature to assess generalization abilities of neural networks. In parallel, researchers have been exploring different directions towards compositional generalization \eg meta-learning, \citep{lake2019compositional}, altering existing architectures \citep{akyurek-andreas-2021-lexicon}, and data augmentation \citep{qiu-etal-2022-improving}. %

\noindent\textbf{Grounded Learning.} \citet{zhou2023non} studied grounded learning in a multimodal procedural setup. \citet{wu2022understanding} benchmarked reasoning and sequencing capabilities of models in a grounded multimodal instructional setting. \citet{johnson2017clevr} studied systematic generalization in visual reasoning tasks. \citet{bahdanau2018systematic} investigated systematic generalization in a VQA-like context while \citet{nikolaus2019compositional} focused on compositionality to construct unseen combinations of concepts while describing images.  
\citet{seolookbefore2020} transcribed speech to rank correct utterances in instructional videos. \citet{suris2019learning} studied compositionality in word acquisition from narrated videos. 
\citet{DBLP:journals/corr/abs-2005-00785} investigated continual learning in unseen compound acquisition from paired image-caption streams.

Other existing studies revolve around crafting conceptual benchmark datasets specifically designed to evaluate compositionality, \eg \citep{bahdanau2019closure,vani2021iterated}. Grounded compositional generalization is explored in \citep{ruis2020benchmark, wu2021reascan} within a 2D grid environment. \citet{xu-etal-2021-zero, DBLP:journals/tmlr/YunBPS23} investigated grounded compositional generalization for the concept learning problem. \citet{li2022maqa} studied compositionality in a grounded setup with audio-language, while \citet{chen2021distilling} leveraged audio-vision modality pairs.

\noindent\textbf{Foundation Models.} Recently, researchers have been studying foundation models to explore the possibilities of utilizing different modalities such as audio, vision, and text to solve grounded real-world problems~\citep{guzhov2022audioclip, girdhar2023imagebind, driess2023palm}. More recently, to assess visually-grounded compositional generalization capabilities of models, \citet{DBLP:journals/corr/abs-2109-10613} proposed COVR, \citet{zhuo-etal-2023-vilpact} proposed ViLPAct, 
\citet{Ma_2023_CVPR} proposed CREPE. Unlike previous studies, in this work, we focus on a real-world audio, vision, and language setting for compositional generalization (see Fig. \ref{fig:overview} for an overview). We believe this study contributes toward a better understanding of the open challenges in multimodal sequential compositional generalization for foundation models and spur interest in this direction.

\section{Conclusion}

In this paper, we investigate linguistic compositionality and systematic generalization in a grounded setting for multimodal sequential compositional generalization. We show how a multimodal dataset can be utilized as a challenging test bed for this purpose. We design next utterance prediction and atom classification tasks adopting a methodical approach in generating the training, validation and test sets for our compositional splits. We experiment with several baseline models and investigate models' ability to generalize to novel compositions and show how multimodal data can contribute towards solving systematic generalization problem and highlight major challenges. We hope our work will stimulate further research in these directions.

\section{Limitations}

Despite the promising results, there are a few limitations of our work. In our work, we introduce a novel dataset called \textsc{CompAct} carefully curated from the EK-100 dataset~\citep{damen2020ek100}, which involves videos of daily kitchen activities, to dissect the impact of visual and auditory signals on linguistic compositionality. Hence, our conclusions may hinge upon certain domain-specific variables. It could be interesting to conduct future studies in an open-domain setting which might unravel additional insights \eg \citep{grauman2022ego4d}.
We investigate several different multimodal models for both the next utterance prediction and atom classification tasks. However, it is important to note that for multimodal learning how to integrate different modalities is considered as an open research problem. In the literature, different strategies for multimodal data fusion have been proposed. Our experimental analysis could be further extended by considering some models that fuse the modalities in a way different than ours. More interestingly, from a systematic generalization point of view, an analysis could be carried out to explore the most effective fusion scheme. Finally, we acknowledge the textual utterances that we use in our work are inherently simplistic and do not capture all of the complexities in natural languages. Consequently, extending this work to a more natural source of language data that mirrors those complexities could be quite interesting direction for future research.

\section*{Ethics Statement} \label{sec:app_ethics}
We curated our \textsc{CompAct} dataset using the video clips from the published EPIC-Kitchens-100 dataset ~\citep{damen2020ek100}, which is publicly available under CC BY-NC 4.0 DEED license. The videos that exist in this dataset were recorded voluntarily by the participants who were not financially rewarded. To the best of our knowledge, the proposed systems or models do not pose any risk in terms of fairness, environmental impact and unintended harm or bias.

\section*{Acknowledgment}
This work was partly supported by the KUIS AI Center Fellowship to Osman Batur İnce. This work was supported by a research grant (VIL53122) from VILLUM FONDEN.

\bibliography{main}

\begin{thebibliography}{70}
\expandafter\ifx\csname natexlab\endcsname\relax\def\natexlab#1{#1}\fi

\bibitem[{Akyurek and Andreas(2021)}]{akyurek-andreas-2021-lexicon}
Ekin Akyurek and Jacob Andreas. 2021.
\newblock \href {https://doi.org/10.18653/v1/2021.acl-long.382} {Lexicon learning for few shot sequence modeling}.
\newblock In \emph{Proceedings of the 59th Annual Meeting of the Association for Computational Linguistics and the 11th International Joint Conference on Natural Language Processing (Volume 1: Long Papers)}, pages 4934--4946, Online. Association for Computational Linguistics.

\bibitem[{Alayrac et~al.(2022)Alayrac, Donahue, Luc, Miech, Barr, Hasson, Lenc, Mensch, Millican, Reynolds et~al.}]{alayrac2022flamingo}
Jean-Baptiste Alayrac, Jeff Donahue, Pauline Luc, Antoine Miech, Iain Barr, Yana Hasson, Karel Lenc, Arthur Mensch, Katherine Millican, Malcolm Reynolds, et~al. 2022.
\newblock Flamingo: a visual language model for few-shot learning.
\newblock \emph{Advances in Neural Information Processing Systems}, 35:23716--23736.

\bibitem[{Awadalla et~al.(2023)Awadalla, Gao, Gardner, Hessel, Hanafy, Zhu, Marathe, Bitton, Gadre, Sagawa, Jitsev, Kornblith, Koh, Ilharco, Wortsman, and Schmidt}]{Awadalla2023OpenFlamingoAO}
Anas Awadalla, Irena Gao, Josh Gardner, Jack Hessel, Yusuf Hanafy, Wanrong Zhu, Kalyani Marathe, Yonatan Bitton, Samir~Yitzhak Gadre, Shiori Sagawa, Jenia Jitsev, Simon Kornblith, Pang~Wei Koh, Gabriel Ilharco, Mitchell Wortsman, and Ludwig Schmidt. 2023.
\newblock \href {https://api.semanticscholar.org/CorpusID:261043320} {Openflamingo: An open-source framework for training large autoregressive vision-language models}.
\newblock \emph{ArXiv}, abs/2308.01390.

\bibitem[{Bahdanau et~al.(2019)Bahdanau, Murty, Noukhovitch, Nguyen, de~Vries, and Courville}]{bahdanau2018systematic}
Dzmitry Bahdanau, Shikhar Murty, Michael Noukhovitch, Thien~Huu Nguyen, Harm de~Vries, and Aaron~C. Courville. 2019.
\newblock Systematic generalization: What is required and can it be learned?
\newblock In \emph{7th International Conference on Learning Representations, {ICLR} 2019, New Orleans, LA, USA, May 6-9, 2019}. OpenReview.net.

\bibitem[{Bar(2007)}]{bar2007proactive}
Moshe Bar. 2007.
\newblock The proactive brain: using analogies and associations to generate predictions.
\newblock \emph{Trends in cognitive sciences}, 11(7):280--289.

\bibitem[{Baroni(2020)}]{baroni2019linguistic}
Marco Baroni. 2020.
\newblock Linguistic generalization and compositionality in modern artificial neural networks.
\newblock \emph{Phil. Trans. R. Soc. B}, 375(1791):20190307.

\bibitem[{Bogin et~al.(2021)Bogin, Gupta, Gardner, and Berant}]{DBLP:journals/corr/abs-2109-10613}
Ben Bogin, Shivanshu Gupta, Matt Gardner, and Jonathan Berant. 2021.
\newblock \href {http://arxiv.org/abs/2109.10613} {{COVR:} {A} test-bed for visually grounded compositional generalization with real images}.
\newblock \emph{CoRR}, abs/2109.10613.

\bibitem[{Chen et~al.(2020)Chen, Xie, Vedaldi, and Zisserman}]{chen2020vggsound}
Honglie Chen, Weidi Xie, Andrea Vedaldi, and Andrew Zisserman. 2020.
\newblock Vggsound: A large-scale audio-visual dataset.
\newblock In \emph{ICASSP 2020-2020 IEEE International Conference on Acoustics, Speech and Signal Processing (ICASSP)}, pages 721--725. IEEE.

\bibitem[{Chen et~al.(2021)Chen, Xian, Koepke, Shan, and Akata}]{chen2021distilling}
Yanbei Chen, Yongqin Xian, A~Koepke, Ying Shan, and Zeynep Akata. 2021.
\newblock Distilling audio-visual knowledge by compositional contrastive learning.
\newblock In \emph{Proceedings of the IEEE/CVF Conference on Computer Vision and Pattern Recognition}, pages 7016--7025.

\bibitem[{Chung et~al.(1989)Chung, Kannappan, Ng, and Sahoo}]{chung1989measures}
JK~Chung, PL~Kannappan, CT~Ng, and PK~Sahoo. 1989.
\newblock Measures of distance between probability distributions.
\newblock \emph{Journal of mathematical analysis and applications}, 138(1):280--292.

\bibitem[{Clark(2015)}]{clark2015surfing}
Andy Clark. 2015.
\newblock \emph{Surfing uncertainty: Prediction, action, and the embodied mind}.
\newblock Oxford University Press.

\bibitem[{Csord{\'{a}}s et~al.(2021)Csord{\'{a}}s, Irie, and Schmidhuber}]{DBLP:conf/emnlp/CsordasIS21}
R{\'{o}}bert Csord{\'{a}}s, Kazuki Irie, and J{\"{u}}rgen Schmidhuber. 2021.
\newblock \href {https://doi.org/10.18653/v1/2021.emnlp-main.49} {The devil is in the detail: Simple tricks improve systematic generalization of transformers}.
\newblock In \emph{Proceedings of the 2021 Conference on Empirical Methods in Natural Language Processing, {EMNLP} 2021, Virtual Event / Punta Cana, Dominican Republic, 7-11 November, 2021}, pages 619--634. Association for Computational Linguistics.

\bibitem[{Damen et~al.(2022)Damen, Doughty, Farinella, , Furnari, Ma, Kazakos, Moltisanti, Munro, Perrett, Price, and Wray}]{damen2020ek100}
Dima Damen, Hazel Doughty, Giovanni~Maria Farinella, , Antonino Furnari, Jian Ma, Evangelos Kazakos, Davide Moltisanti, Jonathan Munro, Toby Perrett, Will Price, and Michael Wray. 2022.
\newblock \href {https://doi.org/10.1007/s11263-021-01531-2} {Rescaling egocentric vision: Collection, pipeline and challenges for epic-kitchens-100}.
\newblock \emph{International Journal of Computer Vision (IJCV)}, 130:33–55.

\bibitem[{Dasgupta et~al.(2018)Dasgupta, Guo, Stuhlm{\"{u}}ller, Gershman, and Goodman}]{dasgupta2018evaluating}
Ishita Dasgupta, Demi Guo, Andreas Stuhlm{\"{u}}ller, Samuel~J. Gershman, and Noah~D. Goodman. 2018.
\newblock \href {http://arxiv.org/abs/1802.04302} {Evaluating compositionality in sentence embeddings}.
\newblock \emph{CoRR}, abs/1802.04302.

\bibitem[{de~Vries et~al.(2019)de~Vries, Bahdanau, Murty, Courville, and Beaudoin}]{bahdanau2019closure}
Harm de~Vries, Dzmitry Bahdanau, Shikhar Murty, Aaron~C. Courville, and Philippe Beaudoin. 2019.
\newblock \href {https://vigilworkshop.github.io/static/papers/28.pdf} {{CLOSURE:} assessing systematic generalization of {CLEVR} models}.
\newblock In \emph{Visually Grounded Interaction and Language (ViGIL), NeurIPS 2019 Workshop, Vancouver, Canada, December 13, 2019}.

\bibitem[{Driess et~al.(2023)Driess, Xia, Sajjadi, Lynch, Chowdhery, Ichter, Wahid, Tompson, Vuong, Yu et~al.}]{driess2023palm}
Danny Driess, Fei Xia, Mehdi~SM Sajjadi, Corey Lynch, Aakanksha Chowdhery, Brian Ichter, Ayzaan Wahid, Jonathan Tompson, Quan Vuong, Tianhe Yu, et~al. 2023.
\newblock Palm-e: An embodied multimodal language model.
\newblock \emph{arXiv preprint arXiv:2303.03378}.

\bibitem[{Ettinger et~al.(2018)Ettinger, Elgohary, Phillips, and Resnik}]{ettinger2018assessing}
Allyson Ettinger, Ahmed Elgohary, Colin Phillips, and Philip Resnik. 2018.
\newblock \href {https://www.aclweb.org/anthology/C18-1152/} {Assessing composition in sentence vector representations}.
\newblock In \emph{Proceedings of the 27th International Conference on Computational Linguistics, {COLING} 2018, Santa Fe, New Mexico, USA, August 20-26, 2018}, pages 1790--1801. Association for Computational Linguistics.

\bibitem[{Gammulle et~al.(2019)Gammulle, Denman, Sridharan, and Fookes}]{gammulle2019predicting}
Harshala Gammulle, Simon Denman, Sridha Sridharan, and Clinton Fookes. 2019.
\newblock Predicting the future: A jointly learnt model for action anticipation.
\newblock In \emph{Proceedings of the IEEE/CVF International Conference on Computer Vision}, pages 5562--5571.

\bibitem[{Girdhar et~al.(2023)Girdhar, El-Nouby, Liu, Singh, Alwala, Joulin, and Misra}]{girdhar2023imagebind}
Rohit Girdhar, Alaaeldin El-Nouby, Zhuang Liu, Mannat Singh, Kalyan~Vasudev Alwala, Armand Joulin, and Ishan Misra. 2023.
\newblock Imagebind: One embedding space to bind them all.
\newblock In \emph{Proceedings of the IEEE/CVF Conference on Computer Vision and Pattern Recognition}, pages 15180--15190.

\bibitem[{Grauman et~al.(2022)Grauman, Westbury, Byrne, Chavis, Furnari, Girdhar, Hamburger, Jiang, Liu, Liu et~al.}]{grauman2022ego4d}
Kristen Grauman, Andrew Westbury, Eugene Byrne, Zachary Chavis, Antonino Furnari, Rohit Girdhar, Jackson Hamburger, Hao Jiang, Miao Liu, Xingyu Liu, et~al. 2022.
\newblock Ego4d: Around the world in 3,000 hours of egocentric video.
\newblock In \emph{Proceedings of the IEEE/CVF Conference on Computer Vision and Pattern Recognition}, pages 18995--19012.

\bibitem[{Guzhov et~al.(2022)Guzhov, Raue, Hees, and Dengel}]{guzhov2022audioclip}
Andrey Guzhov, Federico Raue, J{\"o}rn Hees, and Andreas Dengel. 2022.
\newblock Audioclip: Extending clip to image, text and audio.
\newblock In \emph{ICASSP 2022-2022 IEEE International Conference on Acoustics, Speech and Signal Processing (ICASSP)}, pages 976--980. IEEE.

\bibitem[{He et~al.(2016)He, Zhang, Ren, and Sun}]{he2016deep}
Kaiming He, Xiangyu Zhang, Shaoqing Ren, and Jian Sun. 2016.
\newblock Deep residual learning for image recognition.
\newblock In \emph{Proceedings of the IEEE conference on CVPR}, pages 770--778.

\bibitem[{Hill et~al.(2019)Hill, Lampinen, Schneider, Clark, Botvinick, McClelland, and Santoro}]{hill2019emergent}
Felix Hill, Andrew~K. Lampinen, Rosalia Schneider, Stephen Clark, Matthew Botvinick, James~L. McClelland, and Adam Santoro. 2019.
\newblock \href {http://arxiv.org/abs/1910.00571} {Emergent systematic generalization in a situated agent}.
\newblock \emph{CoRR}, abs/1910.00571.

\bibitem[{Jin et~al.(2020)Jin, Du, and Ren}]{DBLP:journals/corr/abs-2005-00785}
Xisen Jin, Junyi Du, and Xiang Ren. 2020.
\newblock \href {http://arxiv.org/abs/2005.00785} {Visually grounded continual learning of compositional semantics}.
\newblock \emph{CoRR}, abs/2005.00785.

\bibitem[{Johnson et~al.(2017)Johnson, Hariharan, van~der Maaten, Fei{-}Fei, Zitnick, and Girshick}]{johnson2017clevr}
Justin Johnson, Bharath Hariharan, Laurens van~der Maaten, Li~Fei{-}Fei, C.~Lawrence Zitnick, and Ross~B. Girshick. 2017.
\newblock \href {https://doi.org/10.1109/CVPR.2017.215} {{CLEVR:} {A} diagnostic dataset for compositional language and elementary visual reasoning}.
\newblock In \emph{2017 {IEEE} Conference on Computer Vision and Pattern Recognition, {CVPR} 2017, Honolulu, HI, USA, July 21-26, 2017}, pages 1988--1997. {IEEE} Computer Society.

\bibitem[{Ke et~al.(2019)Ke, Fritz, and Schiele}]{ke2019time}
Qiuhong Ke, Mario Fritz, and Bernt Schiele. 2019.
\newblock Time-conditioned action anticipation in one shot.
\newblock In \emph{Proceedings of the IEEE/CVF Conference on Computer Vision and Pattern Recognition}, pages 9925--9934.

\bibitem[{Keysers et~al.(2020)Keysers, Sch{\"{a}}rli, Scales, Buisman, Furrer, Kashubin, Momchev, Sinopalnikov, Stafiniak, Tihon, Tsarkov, Wang, van Zee, and Bousquet}]{keysers2019measuring}
Daniel Keysers, Nathanael Sch{\"{a}}rli, Nathan Scales, Hylke Buisman, Daniel Furrer, Sergii Kashubin, Nikola Momchev, Danila Sinopalnikov, Lukasz Stafiniak, Tibor Tihon, Dmitry Tsarkov, Xiao Wang, Marc van Zee, and Olivier Bousquet. 2020.
\newblock \href {https://openreview.net/forum?id=SygcCnNKwr} {Measuring compositional generalization: {A} comprehensive method on realistic data}.
\newblock In \emph{8th International Conference on Learning Representations, {ICLR} 2020, Addis Ababa, Ethiopia, April 26-30, 2020}. OpenReview.net.

\bibitem[{Kim and Linzen(2020)}]{cogsdataset}
Najoung Kim and Tal Linzen. 2020.
\newblock \href {https://doi.org/10.18653/v1/2020.emnlp-main.731} {{COGS}: A compositional generalization challenge based on semantic interpretation}.
\newblock In \emph{Proceedings of the 2020 Conference on Empirical Methods in Natural Language Processing (EMNLP)}, pages 9087--9105, Online. Association for Computational Linguistics.

\bibitem[{Kim et~al.(2022)Kim, Linzen, and Smolensky}]{kim2022uncontrolled}
Najoung Kim, Tal Linzen, and Paul Smolensky. 2022.
\newblock Uncontrolled lexical exposure leads to overestimation of compositional generalization in pretrained models.
\newblock \emph{arXiv preprint arXiv:2212.10769}.

\bibitem[{Lake(2019)}]{lake2019compositional}
Brenden~M. Lake. 2019.
\newblock \href {https://proceedings.neurips.cc/paper/2019/hash/f4d0e2e7fc057a58f7ca4a391f01940a-Abstract.html} {Compositional generalization through meta sequence-to-sequence learning}.
\newblock In \emph{Advances in Neural Information Processing Systems 32: Annual Conference on Neural Information Processing Systems 2019, NeurIPS 2019, December 8-14, 2019, Vancouver, BC, Canada}, pages 9788--9798.

\bibitem[{Lake and Baroni(2017)}]{scandataset}
Brenden~M. Lake and Marco Baroni. 2017.
\newblock \href {http://arxiv.org/abs/1711.00350} {Still not systematic after all these years: On the compositional skills of sequence-to-sequence recurrent networks}.
\newblock \emph{CoRR}, abs/1711.00350.

\bibitem[{Lake and Baroni(2018)}]{lake2017generalization}
Brenden~M. Lake and Marco Baroni. 2018.
\newblock \href {http://proceedings.mlr.press/v80/lake18a.html} {Generalization without systematicity: On the compositional skills of sequence-to-sequence recurrent networks}.
\newblock In \emph{Proceedings of the 35th International Conference on Machine Learning, {ICML} 2018, Stockholmsm{\"{a}}ssan, Stockholm, Sweden, July 10-15, 2018}, volume~80 of \emph{Proceedings of Machine Learning Research}, pages 2879--2888. {PMLR}.

\bibitem[{Lake et~al.(2019)Lake, Linzen, and Baroni}]{lake2019human}
Brenden~M. Lake, Tal Linzen, and Marco Baroni. 2019.
\newblock \href {https://mindmodeling.org/cogsci2019/papers/0123/index.html} {Human few-shot learning of compositional instructions}.
\newblock In \emph{Proceedings of the 41th Annual Meeting of the Cognitive Science Society, CogSci 2019: Creativity + Cognition + Computation, Montreal, Canada, July 24-27, 2019}, pages 611--617. cognitivesciencesociety.org.

\bibitem[{Lauren{\c{c}}on et~al.(2023)Lauren{\c{c}}on, Saulnier, Tronchon, Bekman, Singh, Lozhkov, Wang, Karamcheti, Rush, Kiela et~al.}]{Laurenccon2023OBELISCAO}
Hugo Lauren{\c{c}}on, Lucile Saulnier, L{\'e}o Tronchon, Stas Bekman, Amanpreet Singh, Anton Lozhkov, Thomas Wang, Siddharth Karamcheti, Alexander~M Rush, Douwe Kiela, et~al. 2023.
\newblock Obelisc: An open web-scale filtered dataset of interleaved image-text documents.
\newblock \emph{arXiv preprint arXiv:2306.16527}.

\bibitem[{Lazaridou et~al.(2021)Lazaridou, Kuncoro, Gribovskaya, Agrawal, Liska, Terzi, Gimenez, de~Masson~d'Autume, Kocisky, Ruder et~al.}]{lazaridou2021mind}
Angeliki Lazaridou, Adhi Kuncoro, Elena Gribovskaya, Devang Agrawal, Adam Liska, Tayfun Terzi, Mai Gimenez, Cyprien de~Masson~d'Autume, Tomas Kocisky, Sebastian Ruder, et~al. 2021.
\newblock Mind the gap: Assessing temporal generalization in neural language models.
\newblock \emph{Advances in Neural Information Processing Systems}, 34:29348--29363.

\bibitem[{Li et~al.(2023)Li, Zhang, Chen, Wang, Yang, and Liu}]{li2023otter}
Bo~Li, Yuanhan Zhang, Liangyu Chen, Jinghao Wang, Jingkang Yang, and Ziwei Liu. 2023.
\newblock Otter: A multi-modal model with in-context instruction tuning.
\newblock \emph{arXiv preprint arXiv:2305.03726}.

\bibitem[{Li et~al.(2022{\natexlab{a}})Li, Jansen, Huang, Ganti, Lee, and Kuzmin}]{li2022maqa}
Judith~Yue Li, Aren Jansen, Qingqing Huang, Ravi Ganti, Joonseok Lee, and Dima Kuzmin. 2022{\natexlab{a}}.
\newblock Maqa: A multimodal qa benchmark for negation.
\newblock In \emph{NeurIPS 2022 Workshop on Synthetic Data for Empowering ML Research}.

\bibitem[{Li et~al.(2022{\natexlab{b}})Li, Chen, Tang, Bao, Zhang, Zhou, and Lu}]{li2022gain}
Junlong Li, Guangyi Chen, Yansong Tang, Jinan Bao, Kun Zhang, Jie Zhou, and Jiwen Lu. 2022{\natexlab{b}}.
\newblock Gain: On the generalization of instructional action understanding.
\newblock In \emph{The Eleventh International Conference on Learning Representations}.

\bibitem[{Lin et~al.(2014)Lin, Maire, Belongie, Hays, Perona, Ramanan, Doll{\'{a}}r, and Zitnick}]{lin2014microsoft}
Tsung{-}Yi Lin, Michael Maire, Serge~J. Belongie, James Hays, Pietro Perona, Deva Ramanan, Piotr Doll{\'{a}}r, and C.~Lawrence Zitnick. 2014.
\newblock \href {https://doi.org/10.1007/978-3-319-10602-1\_48} {Microsoft {COCO:} common objects in context}.
\newblock In \emph{Computer Vision - {ECCV} 2014 - 13th European Conference, Zurich, Switzerland, September 6-12, 2014, Proceedings, Part {V}}, volume 8693 of \emph{Lecture Notes in Computer Science}, pages 740--755. Springer.

\bibitem[{Liska et~al.(2022)Liska, Kocisky, Gribovskaya, Terzi, Sezener, Agrawal, Cyprien De~Masson, Scholtes, Zaheer, Young et~al.}]{liska2022streamingqa}
Adam Liska, Tomas Kocisky, Elena Gribovskaya, Tayfun Terzi, Eren Sezener, Devang Agrawal, D’Autume Cyprien De~Masson, Tim Scholtes, Manzil Zaheer, Susannah Young, et~al. 2022.
\newblock Streamingqa: A benchmark for adaptation to new knowledge over time in question answering models.
\newblock In \emph{International Conference on Machine Learning}, pages 13604--13622. PMLR.

\bibitem[{Liu et~al.(2022)Liu, Shen, Zhang, Dolan, Carin, and Chen}]{liu-etal-2022-makes}
Jiachang Liu, Dinghan Shen, Yizhe Zhang, Bill Dolan, Lawrence Carin, and Weizhu Chen. 2022.
\newblock \href {https://doi.org/10.18653/v1/2022.deelio-1.10} {What makes good in-context examples for {GPT}-3?}
\newblock In \emph{Proceedings of Deep Learning Inside Out (DeeLIO 2022): The 3rd Workshop on Knowledge Extraction and Integration for Deep Learning Architectures}, pages 100--114, Dublin, Ireland and Online. Association for Computational Linguistics.

\bibitem[{Loshchilov and Hutter(2017)}]{loshchilov2017decoupled}
Ilya Loshchilov and Frank Hutter. 2017.
\newblock Decoupled weight decay regularization.
\newblock \emph{arXiv preprint arXiv:1711.05101}.

\bibitem[{Ma et~al.(2023)Ma, Hong, Gul, Gandhi, Gao, and Krishna}]{Ma_2023_CVPR}
Zixian Ma, Jerry Hong, Mustafa~Omer Gul, Mona Gandhi, Irena Gao, and Ranjay Krishna. 2023.
\newblock Crepe: Can vision-language foundation models reason compositionally?
\newblock In \emph{Proceedings of the IEEE/CVF Conference on Computer Vision and Pattern Recognition (CVPR)}, pages 10910--10921.

\bibitem[{Nikolaus et~al.(2019)Nikolaus, Abdou, Lamm, Aralikatte, and Elliott}]{nikolaus2019compositional}
Mitja Nikolaus, Mostafa Abdou, Matthew Lamm, Rahul Aralikatte, and Desmond Elliott. 2019.
\newblock \href {https://doi.org/10.18653/v1/K19-1009} {Compositional generalization in image captioning}.
\newblock In \emph{Proceedings of the 23rd Conference on Computational Natural Language Learning, CoNLL 2019, Hong Kong, China, November 3-4, 2019}, pages 87--98. Association for Computational Linguistics.

\bibitem[{Nityasya et~al.(2023)Nityasya, Wibowo, Aji, Winata, Prasojo, Blunsom, and Kuncoro}]{nityasya2023scientific}
Made~Nindyatama Nityasya, Haryo~Akbarianto Wibowo, Alham~Fikri Aji, Genta~Indra Winata, Radityo~Eko Prasojo, Phil Blunsom, and Adhiguna Kuncoro. 2023.
\newblock On "scientific debt" in nlp: A case for more rigour in language model pre-training research.
\newblock \emph{arXiv preprint arXiv:2306.02870}.

\bibitem[{OpenAI(2023)}]{openai2023gpt4}
OpenAI. 2023.
\newblock \href {http://arxiv.org/abs/2303.08774} {Gpt-4 technical report}.

\bibitem[{Papineni et~al.(2002)Papineni, Roukos, Ward, and Zhu}]{papineni2002bleu}
Kishore Papineni, Salim Roukos, Todd Ward, and Wei-Jing Zhu. 2002.
\newblock Bleu: a method for automatic evaluation of machine translation.
\newblock In \emph{Proceedings of the 40th annual meeting of the Association for Computational Linguistics}, pages 311--318.

\bibitem[{Qiu et~al.(2022{\natexlab{a}})Qiu, Shaw, Pasupat, Nowak, Linzen, Sha, and Toutanova}]{qiu-etal-2022-improving}
Linlu Qiu, Peter Shaw, Panupong Pasupat, Pawel Nowak, Tal Linzen, Fei Sha, and Kristina Toutanova. 2022{\natexlab{a}}.
\newblock \href {https://doi.org/10.18653/v1/2022.naacl-main.323} {Improving compositional generalization with latent structure and data augmentation}.
\newblock In \emph{Proceedings of the 2022 Conference of the North American Chapter of the Association for Computational Linguistics: Human Language Technologies}, pages 4341--4362, Seattle, United States. Association for Computational Linguistics.

\bibitem[{Qiu et~al.(2022{\natexlab{b}})Qiu, Shaw, Pasupat, Shi, Herzig, Pitler, Sha, and Toutanova}]{qiu2022evaluating}
Linlu Qiu, Peter Shaw, Panupong Pasupat, Tianze Shi, Jonathan Herzig, Emily Pitler, Fei Sha, and Kristina Toutanova. 2022{\natexlab{b}}.
\newblock Evaluating the impact of model scale for compositional generalization in semantic parsing.
\newblock In \emph{Proceedings of the 2022 Conference on Empirical Methods in Natural Language Processing}, pages 9157--9179.

\bibitem[{Ren et~al.(2017)Ren, He, Girshick, and Sun}]{ren2015faster}
Shaoqing Ren, Kaiming He, Ross~B. Girshick, and Jian Sun. 2017.
\newblock \href {https://doi.org/10.1109/TPAMI.2016.2577031} {Faster {R-CNN:} {T}owards real-time object detection with region proposal networks}.
\newblock \emph{{IEEE} Trans. Pattern Anal. Mach. Intell.}, 39(6):1137--1149.

\bibitem[{Ruis et~al.(2020)Ruis, Andreas, Baroni, Bouchacourt, and Lake}]{ruis2020benchmark}
Laura Ruis, Jacob Andreas, Marco Baroni, Diane Bouchacourt, and Brenden~M. Lake. 2020.
\newblock \href {https://proceedings.neurips.cc/paper/2020/hash/e5a90182cc81e12ab5e72d66e0b46fe3-Abstract.html} {A benchmark for systematic generalization in grounded language understanding}.
\newblock In \emph{Advances in Neural Information Processing Systems 33: Annual Conference on Neural Information Processing Systems 2020, NeurIPS 2020, December 6-12, 2020, virtual}.

\bibitem[{Russakovsky et~al.(2015)Russakovsky, Deng, Su, Krause, Satheesh, Ma, Huang, Karpathy, Khosla, Bernstein et~al.}]{russakovsky2015imagenet}
Olga Russakovsky, Jia Deng, Hao Su, Jonathan Krause, Sanjeev Satheesh, Sean Ma, Zhiheng Huang, Andrej Karpathy, Aditya Khosla, Michael Bernstein, et~al. 2015.
\newblock Imagenet large scale visual recognition challenge.
\newblock \emph{International journal of computer vision}, 115(3):211--252.

\bibitem[{Seo et~al.(2020)Seo, Nagrani, and Schmid}]{seolookbefore2020}
Paul~Hongsuck Seo, Arsha Nagrani, and Cordelia Schmid. 2020.
\newblock \href {http://arxiv.org/abs/2012.05710} {Look before you speak: Visually contextualized utterances}.
\newblock \emph{CoRR}, abs/2012.05710.

\bibitem[{Sur{\'{\i}}s et~al.(2020)Sur{\'{\i}}s, Epstein, Ji, Chang, and Vondrick}]{suris2019learning}
D{\'{\i}}dac Sur{\'{\i}}s, Dave Epstein, Heng Ji, Shih{-}Fu Chang, and Carl Vondrick. 2020.
\newblock \href {https://doi.org/10.1007/978-3-030-58526-6\_26} {Learning to learn words from visual scenes}.
\newblock In \emph{Computer Vision - {ECCV} 2020 - 16th European Conference, Glasgow, UK, August 23-28, 2020, Proceedings, Part {XXIX}}, volume 12374 of \emph{Lecture Notes in Computer Science}, pages 434--452. Springer.

\bibitem[{Thomason et~al.(2019)Thomason, Gordon, and Bisk}]{thomason-etal-2019-shifting}
Jesse Thomason, Daniel Gordon, and Yonatan Bisk. 2019.
\newblock \href {https://doi.org/10.18653/v1/N19-1197} {Shifting the baseline: Single modality performance on visual navigation {\&} {QA}}.
\newblock In \emph{Proceedings of the 2019 Conference of the North {A}merican Chapter of the Association for Computational Linguistics: Human Language Technologies, Volume 1 (Long and Short Papers)}, pages 1977--1983, Minneapolis, Minnesota. Association for Computational Linguistics.

\bibitem[{Thrush et~al.(2022)Thrush, Jiang, Bartolo, Singh, Williams, Kiela, and Ross}]{thrush2022winoground}
Tristan Thrush, Ryan Jiang, Max Bartolo, Amanpreet Singh, Adina Williams, Douwe Kiela, and Candace Ross. 2022.
\newblock Winoground: Probing vision and language models for visio-linguistic compositionality.
\newblock In \emph{Proceedings of the IEEE/CVF Conference on Computer Vision and Pattern Recognition}, pages 5238--5248.

\bibitem[{Touvron et~al.(2023)Touvron, Martin, and et~al.}]{Touvron2023Llama2O}
Hugo Touvron, Louis Martin, and Kevin~Stone et~al. 2023.
\newblock \href {http://arxiv.org/abs/2307.09288} {Llama 2: Open foundation and fine-tuned chat models}.

\bibitem[{Tsai et~al.(2019)Tsai, Bai, Liang, Kolter, Morency, and Salakhutdinov}]{tsai2019MULT}
Yao-Hung~Hubert Tsai, Shaojie Bai, Paul~Pu Liang, J.~Zico Kolter, Louis-Philippe Morency, and Ruslan Salakhutdinov. 2019.
\newblock Multimodal transformer for unaligned multimodal language sequences.
\newblock In \emph{Proceedings of the 57th Annual Meeting of the Association for Computational Linguistics (Volume 1: Long Papers)}, Florence, Italy. ACL.

\bibitem[{Vani et~al.(2021)Vani, Schwarzer, Lu, Dhekane, and Courville}]{vani2021iterated}
Ankit Vani, Max Schwarzer, Yuchen Lu, Eeshan Dhekane, and Aaron Courville. 2021.
\newblock \href {https://openreview.net/forum?id=Pd_oMxH8IlF} {Iterated learning for emergent systematicity in {VQA}}.
\newblock In \emph{International Conference on Learning Representations}.

\bibitem[{Vaswani et~al.(2017)Vaswani, Shazeer, Parmar, Uszkoreit, Jones, Gomez, Kaiser, and Polosukhin}]{vaswani2017attention}
Ashish Vaswani, Noam Shazeer, Niki Parmar, Jakob Uszkoreit, Llion Jones, Aidan~N. Gomez, Lukasz Kaiser, and Illia Polosukhin. 2017.
\newblock \href {https://proceedings.neurips.cc/paper/2017/hash/3f5ee243547dee91fbd053c1c4a845aa-Abstract.html} {Attention is all you need}.
\newblock In \emph{Advances in Neural Information Processing Systems 30: Annual Conference on Neural Information Processing Systems 2017, December 4-9, 2017, Long Beach, CA, {USA}}, pages 5998--6008.

\bibitem[{Wu et~al.(2022)Wu, Spangher, Alipoormolabashi, Freedman, Weischedel, and Peng}]{wu2022understanding}
Te-Lin Wu, Alex Spangher, Pegah Alipoormolabashi, Marjorie Freedman, Ralph Weischedel, and Nanyun Peng. 2022.
\newblock Understanding multimodal procedural knowledge by sequencing multimodal instructional manuals.
\newblock In \emph{Proceedings of the 60th Annual Meeting of the Association for Computational Linguistics (Volume 1: Long Papers)}, pages 4525--4542.

\bibitem[{Wu et~al.(2021)Wu, Kreiss, Ong, and Potts}]{wu2021reascan}
Zhengxuan Wu, Elisa Kreiss, Desmond~C Ong, and Christopher Potts. 2021.
\newblock Reascan: Compositional reasoning in language grounding.
\newblock \emph{arXiv preprint arXiv:2109.08994}.

\bibitem[{Xu et~al.(2021)Xu, Kordjamshidi, and Chai}]{xu-etal-2021-zero}
Guangyue Xu, Parisa Kordjamshidi, and Joyce Chai. 2021.
\newblock \href {https://doi.org/10.18653/v1/2021.metanlp-1.3} {Zero-shot compositional concept learning}.
\newblock In \emph{Proceedings of the 1st Workshop on Meta Learning and Its Applications to Natural Language Processing}, pages 19--27, Online. Association for Computational Linguistics.

\bibitem[{Xu et~al.(2023)Xu, Zhu, and Clifton}]{xu2023multimodal}
Peng Xu, Xiatian Zhu, and David~A Clifton. 2023.
\newblock Multimodal learning with transformers: A survey.
\newblock \emph{IEEE Transactions on Pattern Analysis and Machine Intelligence}.

\bibitem[{Yun et~al.(2023)Yun, Bhalla, Pavlick, and Sun}]{DBLP:journals/tmlr/YunBPS23}
Tian Yun, Usha Bhalla, Ellie Pavlick, and Chen Sun. 2023.
\newblock \href {https://openreview.net/forum?id=YwNrPLjHSL} {Do vision-language pretrained models learn composable primitive concepts?}
\newblock \emph{Trans. Mach. Learn. Res.}, 2023.

\bibitem[{Zellers et~al.(2022)Zellers, Lu, Lu, Yu, Zhao, Salehi, Kusupati, Hessel, Farhadi, and Choi}]{zellers2022merlot}
Rowan Zellers, Jiasen Lu, Ximing Lu, Youngjae Yu, Yanpeng Zhao, Mohammadreza Salehi, Aditya Kusupati, Jack Hessel, Ali Farhadi, and Yejin Choi. 2022.
\newblock Merlot reserve: Neural script knowledge through vision and language and sound.
\newblock In \emph{Proceedings of the IEEE/CVF Conference on Computer Vision and Pattern Recognition}, pages 16375--16387.

\bibitem[{Zhang et~al.(2019)Zhang, Kishore, Wu, Weinberger, and Artzi}]{zhang2019bertscore}
Tianyi Zhang, Varsha Kishore, Felix Wu, Kilian~Q Weinberger, and Yoav Artzi. 2019.
\newblock Bertscore: Evaluating text generation with bert.
\newblock \emph{arXiv preprint arXiv:1904.09675}.

\bibitem[{Zhou et~al.(2023)Zhou, Li, Li, Lin, Chang, Bansal, and Ji}]{zhou2023non}
Yu~Zhou, Sha Li, Manling Li, Xudong Lin, Shih-Fu Chang, Mohit Bansal, and Heng Ji. 2023.
\newblock Non-sequential graph script induction via multimedia grounding.
\newblock \emph{arXiv preprint arXiv:2305.17542}.

\bibitem[{Zhukov et~al.(2019)Zhukov, Alayrac, Cinbis, Fouhey, Laptev, and Sivic}]{zhukov2019cross}
Dimitri Zhukov, Jean-Baptiste Alayrac, Ramazan~Gokberk Cinbis, David Fouhey, Ivan Laptev, and Josef Sivic. 2019.
\newblock Cross-task weakly supervised learning from instructional videos.
\newblock In \emph{Proceedings of the IEEE/CVF Conference on Computer Vision and Pattern Recognition}, pages 3537--3545.

\bibitem[{Zhuo et~al.(2023)Zhuo, Liao, Lei, Qu, de~Melo, Chang, Ren, and Xu}]{zhuo-etal-2023-vilpact}
Terry~Yue Zhuo, Yaqing Liao, Yuecheng Lei, Lizhen Qu, Gerard de~Melo, Xiaojun Chang, Yazhou Ren, and Zenglin Xu. 2023.
\newblock \href {https://doi.org/10.18653/v1/2023.findings-eacl.164} {{V}i{LPA}ct: A benchmark for compositional generalization on multimodal human activities}.
\newblock In \emph{Findings of the Association for Computational Linguistics: EACL 2023}, pages 2192--2207, Dubrovnik, Croatia. Association for Computational Linguistics.

\end{thebibliography}

\clearpage

\appendix

\section*{Appendix}
In the following, we provide a comprehensive set of supplementary notes that delve deeper into various aspects of our research:

\begin{itemize}
    \item \textbf{Data Curation, Algorithms, and Preprocessing (Section \ref{sec:app_data}):} This section outlines the steps taken in data curation, algorithmic processes, and preprocessing techniques applied.
    \item \textbf{Exploratory Analysis of \textsc{CompAct} (Section \ref{sec:app_exploratory}):} Here, we present a detailed analysis of the \textsc{CompAct} dataset, highlighting its unique characteristics.
    \item \textbf{Implementation Details and Reproducibility (Section  \ref{sec:app_implementation}):} This section offers a detailed account of our implementation methodology, providing valuable information for those interested in replicating or extending our work.
    \item \textbf{Further Analysis (Section \ref{sec:app_analysis}):} We conduct additional analyses, expanding on key findings and offering deeper insights into the compositional generalization phenomenon.
    \item \textbf{Ethics Statement (Section \ref{sec:app_ethics}):} In this section, we present a comprehensive ethics statement detailing our commitment to ethical research practices throughout the study.
\end{itemize}

\section{Data Curation, Algorithms and Preprocessing}\label{sec:app_data}

\subsection{Curating \textsc{CompAct}: an Overview} 
\label{appendix:curating}
In our data curation and preprocessing for \textsc{CompAct}, we leverage the EPIC-Kitchens-100  (EK-100) dataset, a collection of egocentric kitchen activity videos, which are split into shorter clips with accompanying narrations and audio tracks—referred to as ``microsegments". 

To curate our sequences of microsegments, we employ a window of 4 clips, with the initial 3 clips serving as context and the last one designated for prediction, yielding a total of 22,136 instances. We filter out repeated utterances that represent a continuation of the same action, treating them as duplicates. 

\begin{figure}[h]
\centering
\includegraphics[width=\linewidth]{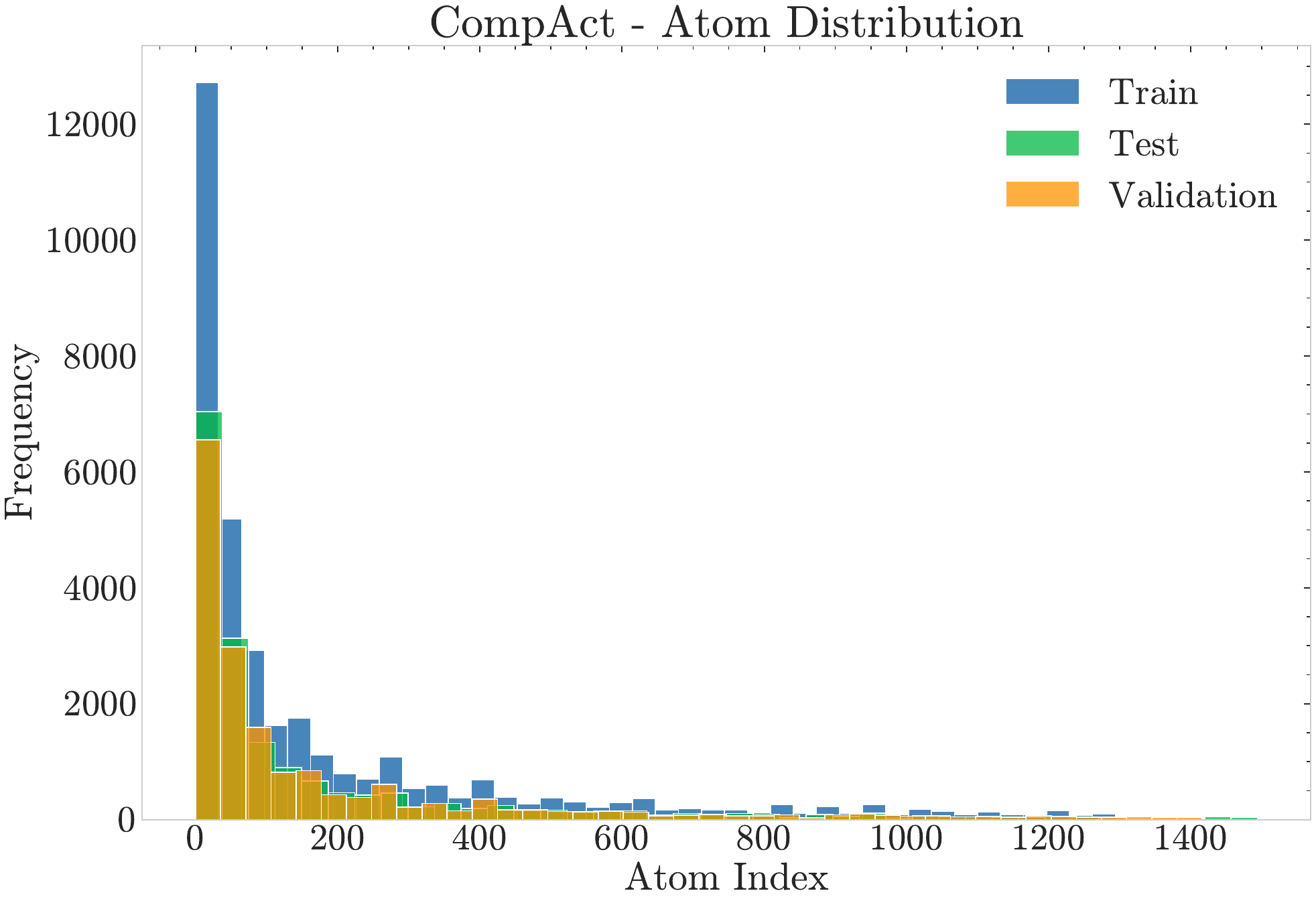}
\includegraphics[width=\linewidth]{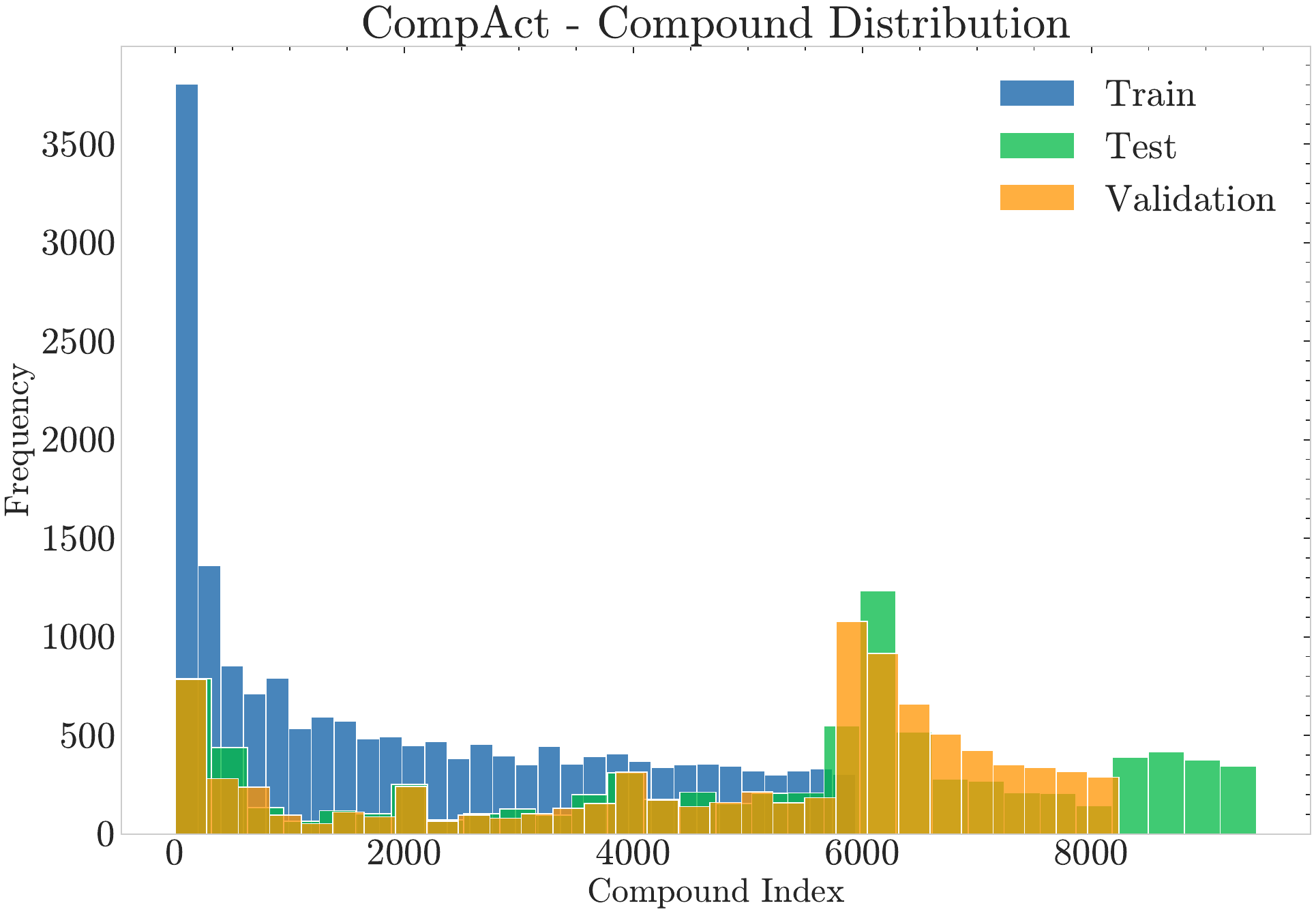}
\caption{Plot on the top demonstrates the distribution of atoms while the plot on the bottom shows the distribution of compounds for the train/validation/test splits in compositional split setup.}
\label{fig:atomcompound}
\end{figure}
Additionally, we exclusively consider text descriptions that share common nouns, ensuring that the noun mentioned in the target description also appears in the source text. This heuristic guarantees the presence of the target noun in both input sequences during both inference and training, allowing our setup to solely evaluate compositionality and systematic generalization.

\begin{figure*}[h]
    \centering
    \scalebox{0.80}{
        \begin{tabular}{cl}
            \toprule
            \textbf{Inputs (utterances and auxiliary modalities)} & \textbf{Target (next utterance)} \\ \midrule

            \begin{tabular}{@{}lccc@{}}
                \centering  
             Image    & \adjincludegraphics[valign=M,width=0.225\linewidth]{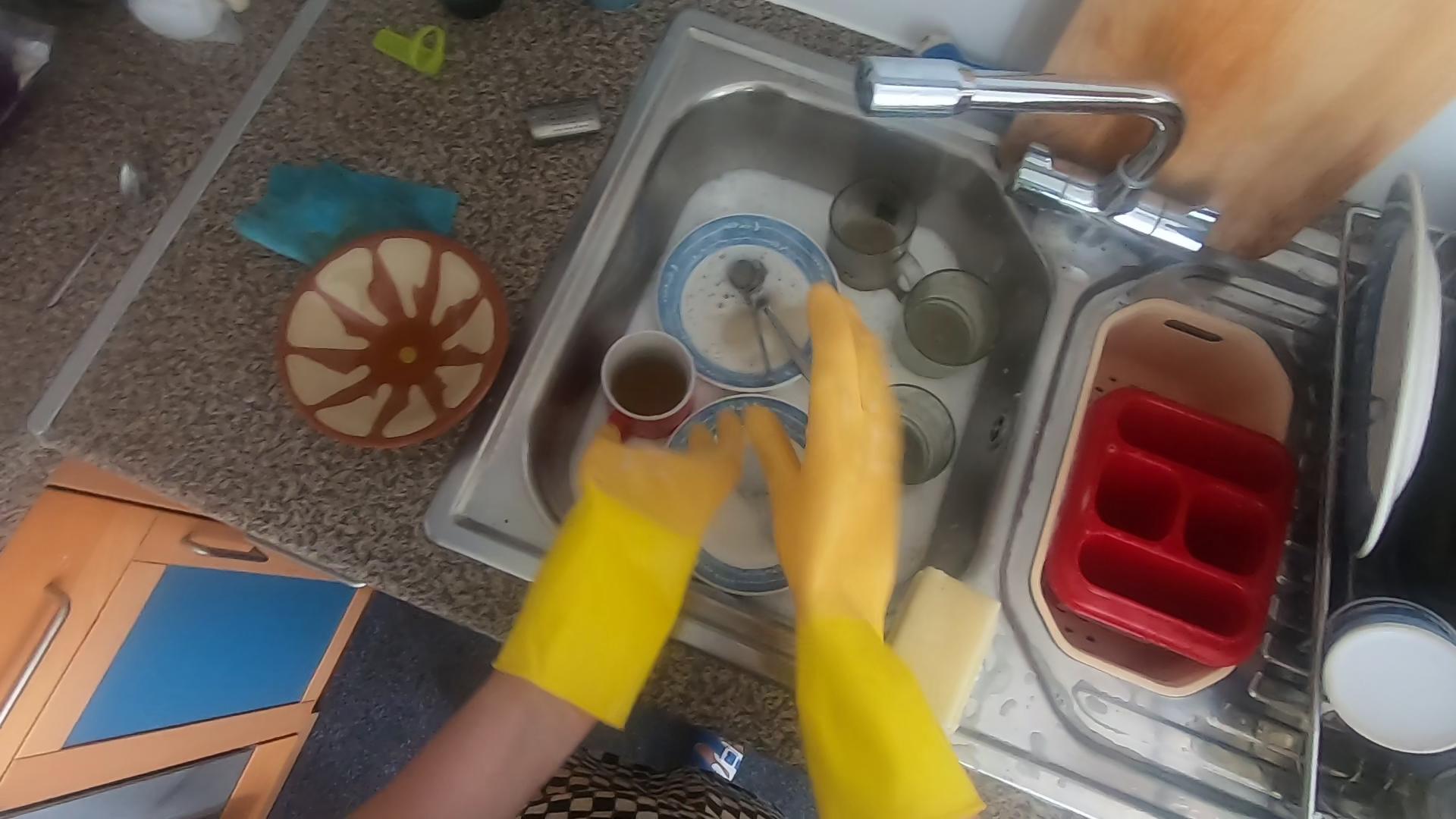} & \adjincludegraphics[valign=M,width=0.225\linewidth]{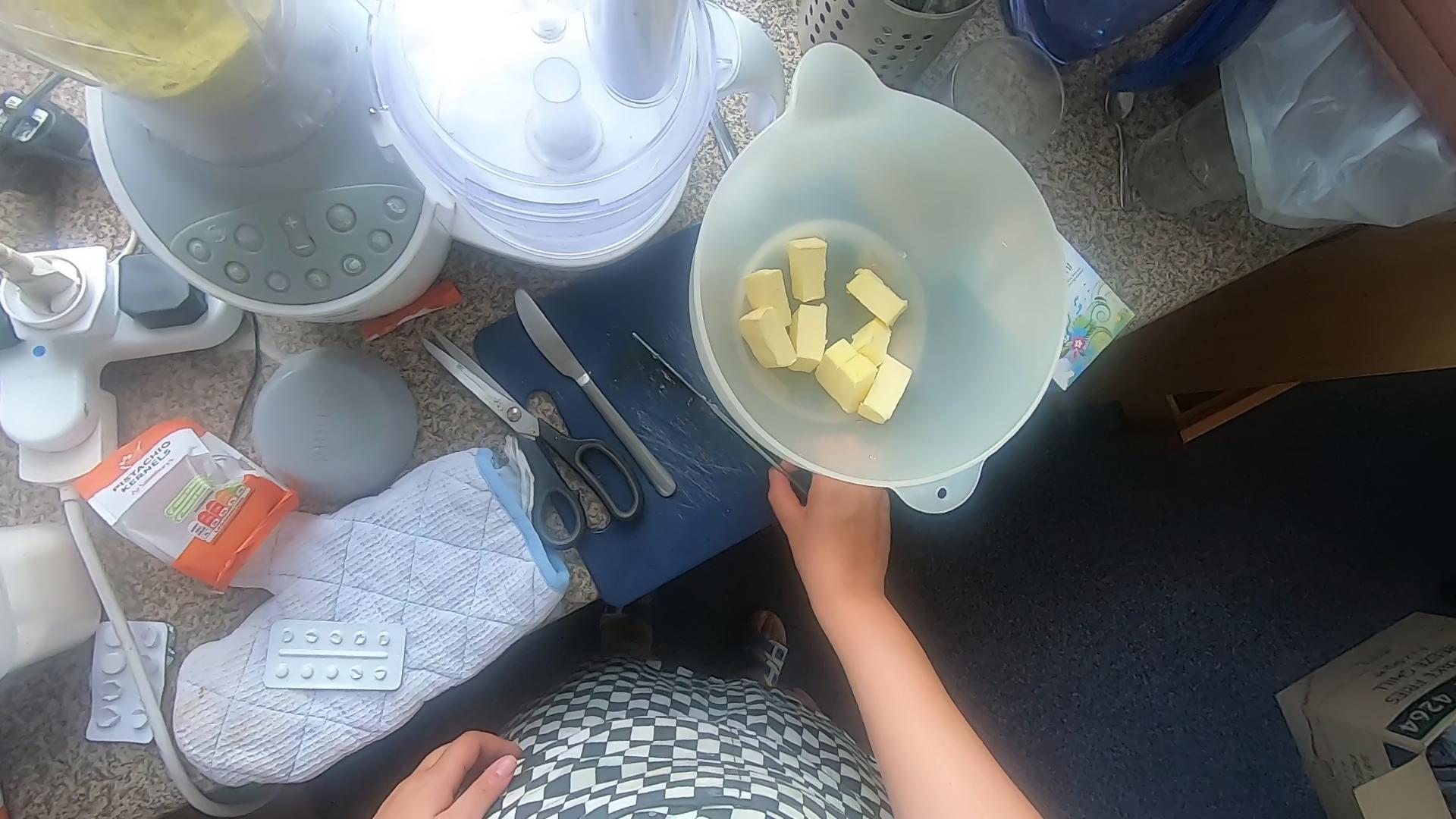} & \adjincludegraphics[valign=M,width=0.225\linewidth]{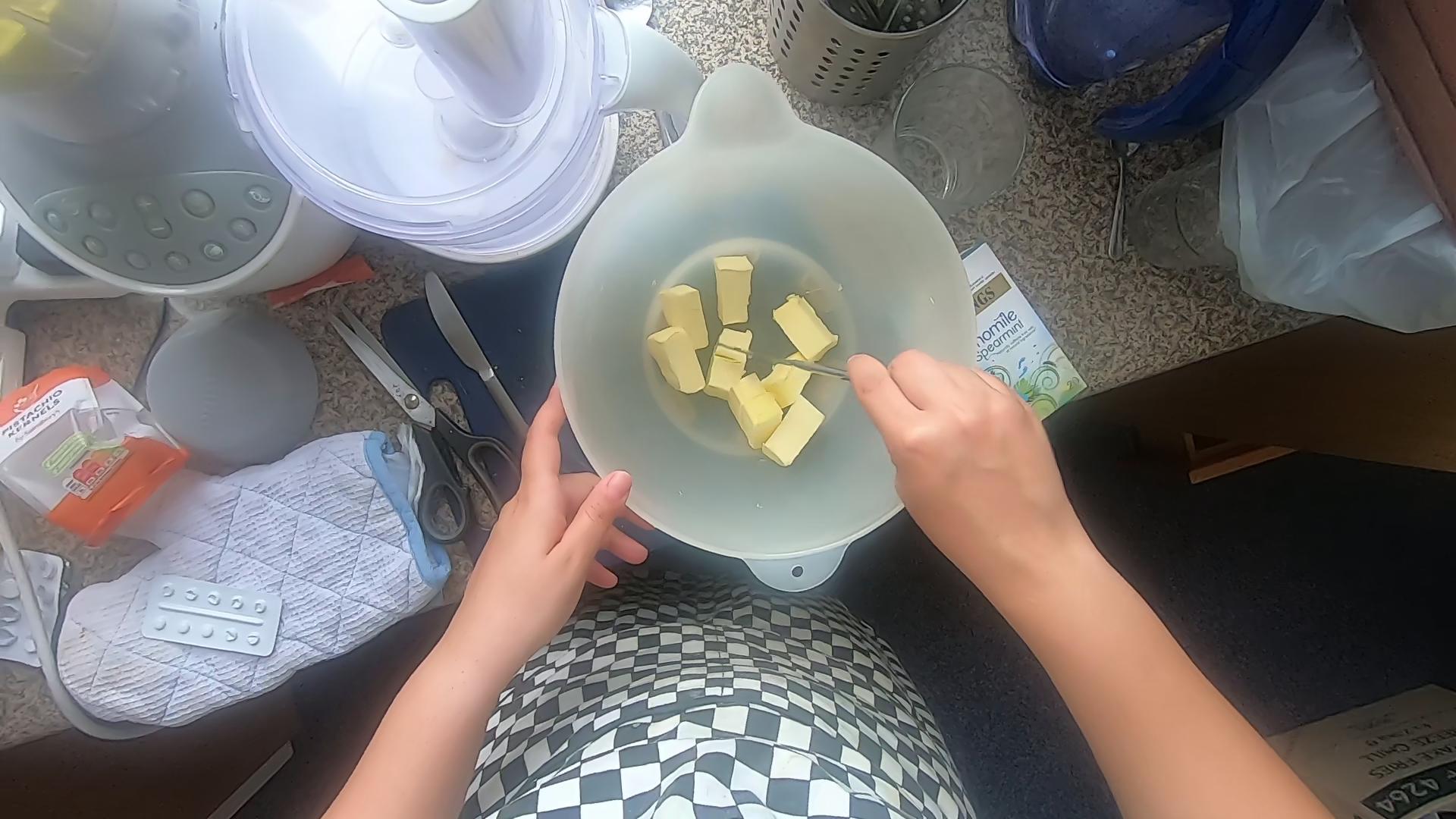}  
             \vspace{0.5em}
             \\
              Text   & take\_off gloves & pick\_up knife & check butter in bowl \\
              
            \end{tabular}
            \vspace{0.5em}
            & 
            \bgroup
            \def\arraystretch{0.8}
            \begin{tabular}{ll}
            GT &: put\_down knife \\
            
            \end{tabular}
            
            \egroup \\
            
            \midrule

            \begin{tabular}{@{}lccc@{}}
            
                \centering
                
             Image    & \adjincludegraphics[valign=M,width=0.225\linewidth]{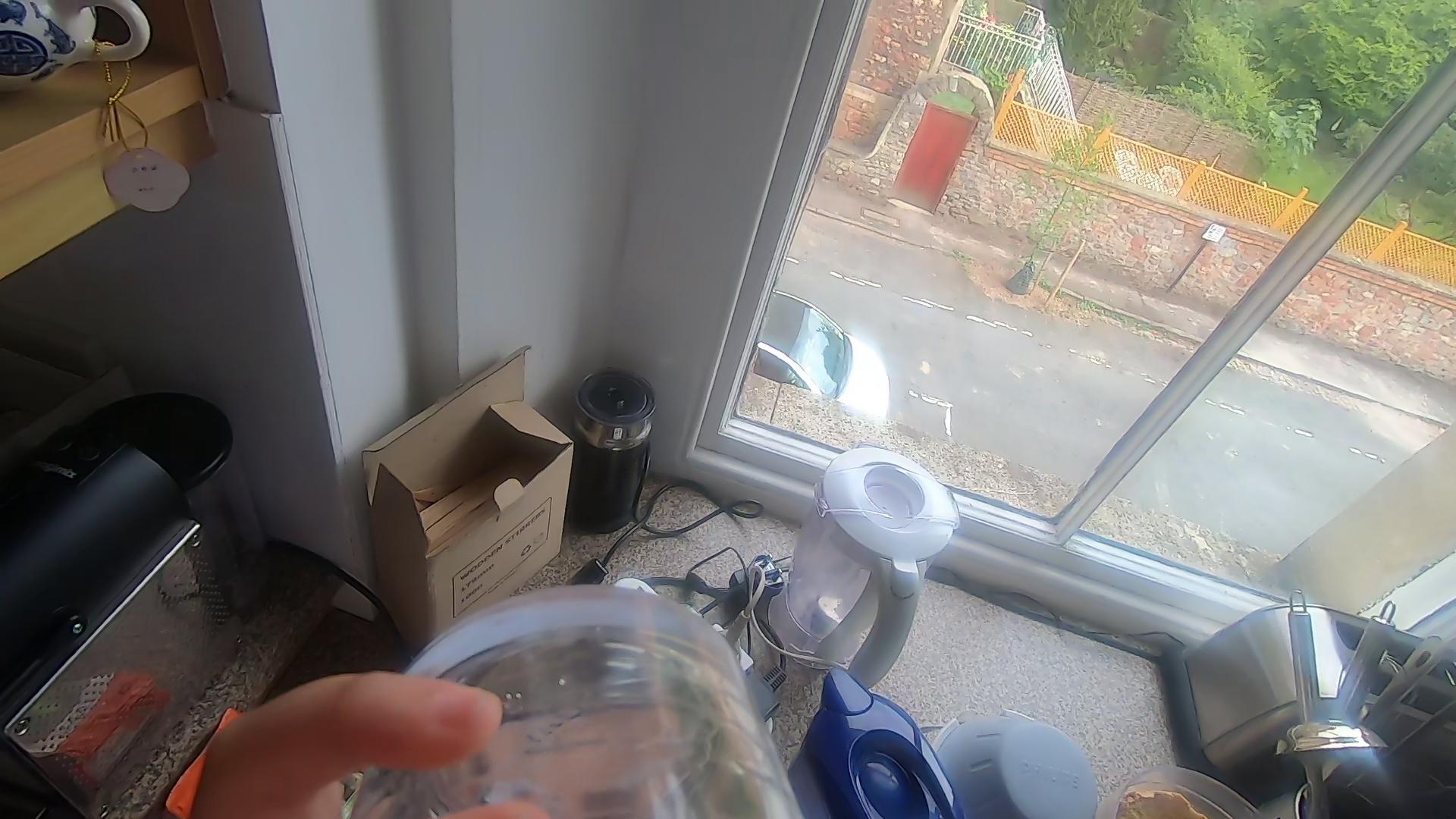} & \adjincludegraphics[valign=M,width=0.225\linewidth]{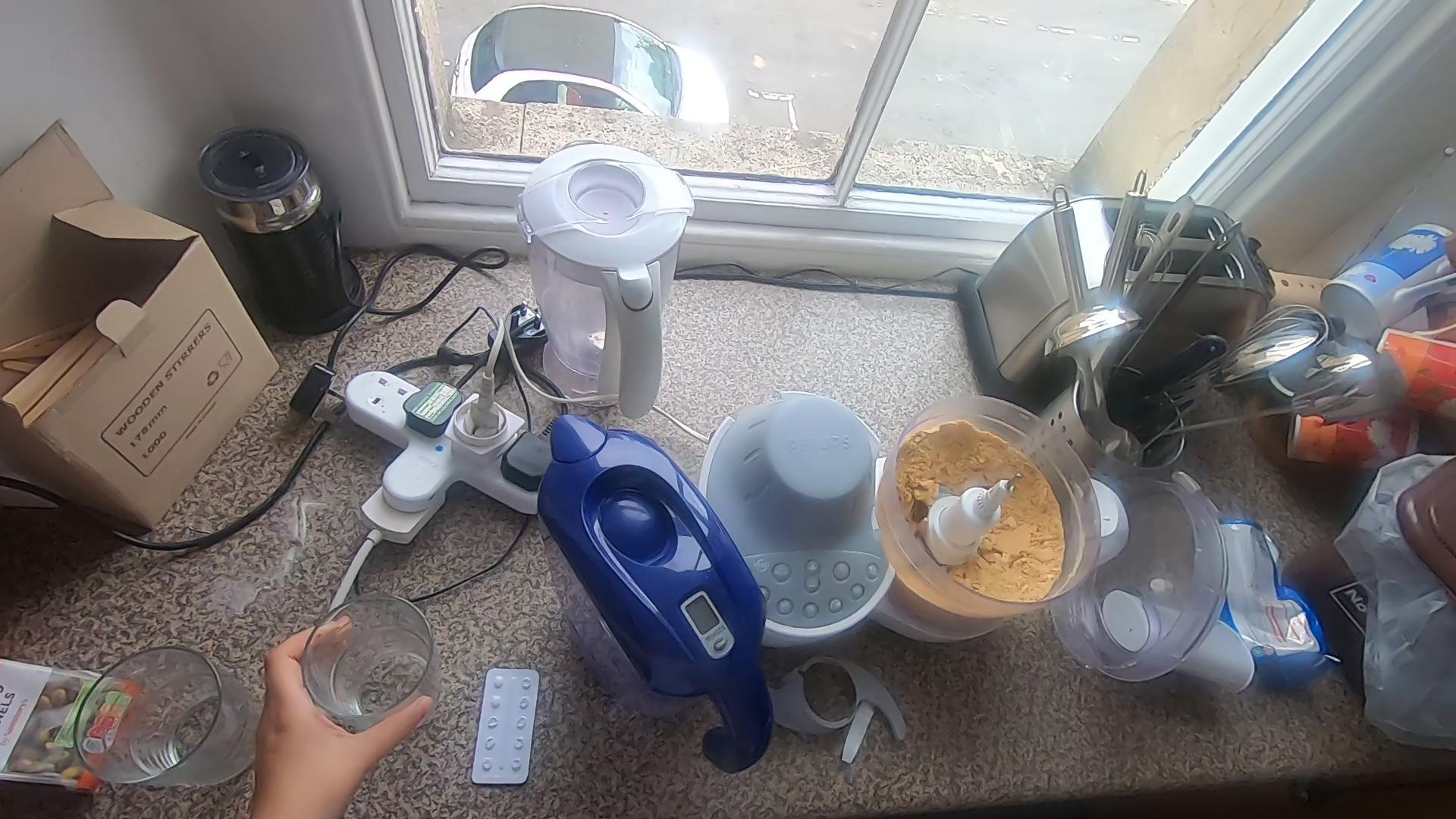} & \adjincludegraphics[valign=M,width=0.225\linewidth]{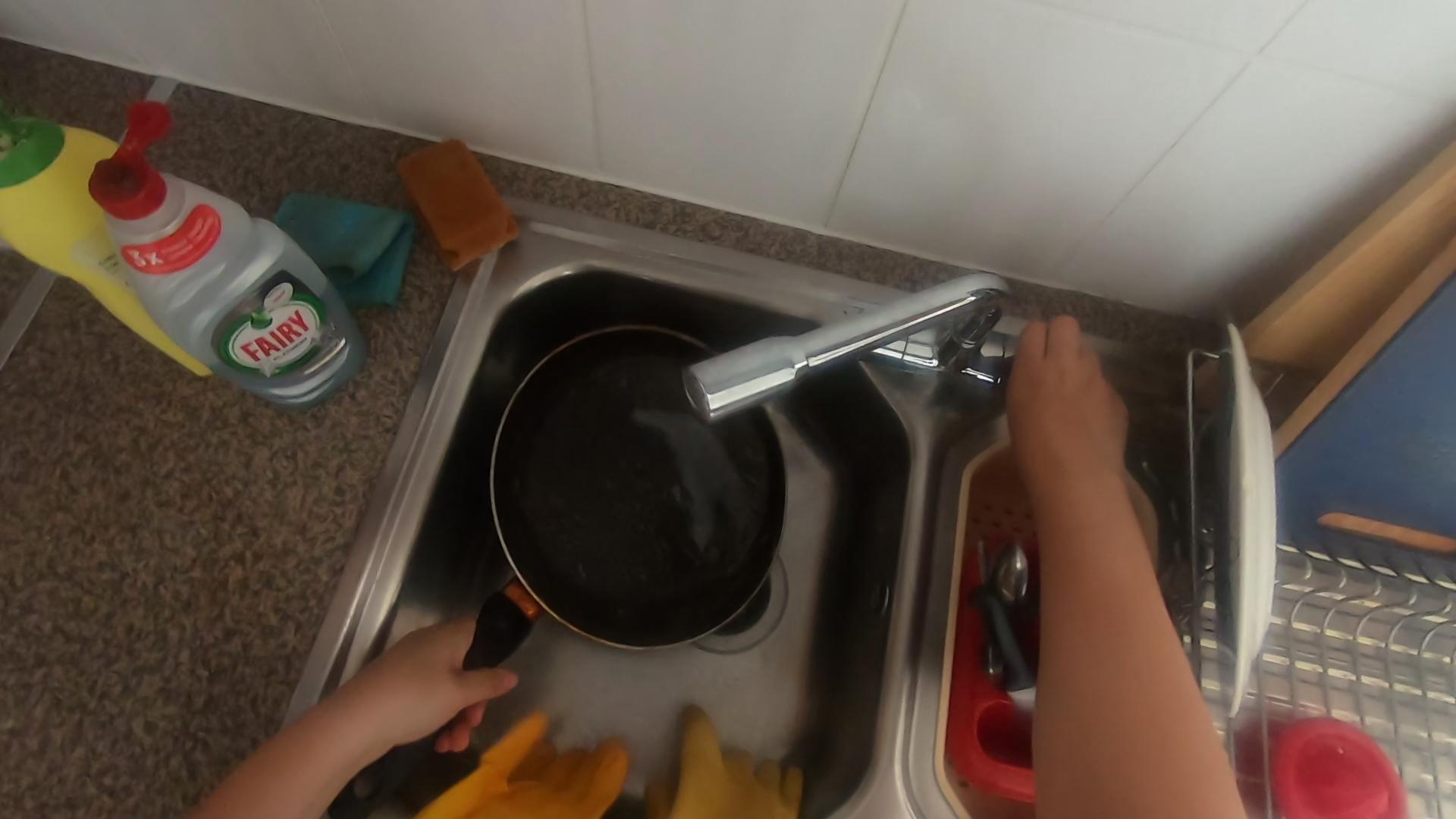}  
             \vspace{0.5em}
             \\
              Text   & put\_down water\_filter & put\_down glass & pick\_up pan from sink \\
            \end{tabular}
            \vspace{0.5em}
            & 
            \bgroup
            \def\arraystretch{0.8}
            \begin{tabular}{ll}
                 GT &: rinse pan 
                
            \end{tabular}
            \egroup \\
            \bottomrule
            
        \end{tabular}
    }
    \caption{{Curating dataset instances for compositional generalization.} 
    Targets such as \textsl{put\_down} \textsc{knife} and \textsl{rinse} \textsc{pan} have never been observed by the learner during the training phase. 
    }
    \label{fig:sample}
\end{figure*}

In our experimental setup, we introduce a scenario where a model must have prior exposure to all constituent atoms within a test instance, such as \textsc{grab the plate. wash cucumber. take knife.}, and is then tasked with predicting the subsequent utterance, such as \textsc{slice cucumber}, during inference. It is important to emphasize that this target composition has never been encountered during the model's training phase (refer to Fig. \ref{fig:sample}). This setup allows testing models' ability to generalize to entirely unobserved compositions, even those with zero probability of occurrence in the training data.
To create such dataset splits, we employ the Maximum Compound Divergence (MCD) heuristic, crafting distributions that maintain similarity in the distribution of individual concepts (atoms), while deliberately introducing disparities in the distributions of concept combinations. In our case, we utilize 97 verb classes and 300 noun classes from the EK-100 dataset as the atoms. In particular, each sample is assigned to a specific split based on the atomic and compound divergence (similarity) based on weighted distributions using Chernoff coefficient ~\citep{chung1989measures}. This process yields 8,766 instances, which are further partitioned into 4,407 for training, 2,184 for validation, and 2,175 for testing.

In Fig.~\ref{fig:atomcompound}, we visualize the atomic and compound distributions over the constructed training, validation, and test splits of our proposed compositional setup. Notably, these splits exhibit similar distributions concerning atoms while training and val/test splits do differ in terms of compounds.

\subsection{Atom and Compound Selection}

In Algorithm \ref{alg:mcd}, we describe the heuristic we use to create the compositional splits in \textsc{CompAct} following the Maximum Compound Divergence \citep{keysers2019measuring}

\begin{algorithm}
\caption{Split Generation Algorithm}
\label{alg:mcd}
\KwData{Dataset $M$}
\KwResult{Train split $U$, Test split $W$}

Init $U$, $W$\;
Init Atom Divergence $D_A$, Compound Divergence $D_C$\;
Init $M_{\text{T}}$ with items in $M$\;
Init $i$ to 0\;

\While{$M_{\text{T}}$ is not empty}{
    Randomly choose $T \in \{U, W\}$ to add an item\;
    \If{$i = 0$}{
        Randomly select and remove an item $m$ from $M_{\text{T}}$\;
        Add $m$ to split $T$ \;
    }
    \Else{
        Calculate $D_A$ for remaining items if added to $T$\;
        Filter items with $D_A$ below a threshold\;
        \If{no items meet the criteria}{
            Select item with highest $D_C$ as the best candidate\;
        }
        \Else{
            Calculate $D_C$ for items if added to $T$\;
            Select the item with highest $D_C$ as the best candidate\;
        }
        Add the best candidate item to split $T$ \;

    }
    Increment $i$ by 1\;
}
\end{algorithm}

\begin{figure*}[!ht]
    \centering
\noindent\fbox{%
    \parbox{.98\textwidth}{%
{\footnotesize \texttt{Predict the next narration given 3 sequential previous narrations from a cooking video}}

{\footnotesize  \texttt{put down bowl . move frying pan . pick up spatula $=>$ put down spatula}}

{\footnotesize \texttt{put down bowl . move jar . pick up egg $=>$ crack egg}}

{\footnotesize \texttt{move yoghurt . put down bowl . pick up yogurt $=>$ put yoghurt}}

{\footnotesize \texttt{put down bowl . grab wok . move tap $=>$ lather wok}}

{\footnotesize \texttt{put down bowl . pick up spatula . stir meat pieces with spatula $=>$ put down spatula}}

{\footnotesize \texttt{pick up tins . put down tins . move bowl $=>$}}

    }%
}
\caption{Prompt template utilized for LLaMA2 evaluation.}
\label{fig:app_prompt_format_llama2}
\end{figure*}

\begin{figure*}
    \centering
    \noindent\fbox{%
    \parbox{.98\textwidth}{%
{\footnotesize \texttt{Predict the next action narration given 3 sequential previous actions (image-narration pairs) in a cooking video.}}

 {\footnotesize \texttt{put down bowl $<$Image 1$>$ . move frying pan $<$Image 2$>$ . pick up spatula $<$Image 3$>$ $=>$ put down spatula}}
 
 {\footnotesize \texttt{pick up tins $<$Image 1$>$ . put down tins $<$Image 2$>$ . move bowl $<$Image 3$>$ $=>$}}
    }%
}
    \caption{Prompt template utilized for IDEFICS evaluation.}
    \label{fig:app_prompt_format_idefics}
\end{figure*}

\subsection{Preprocessing}

\subsubsection{Choosing Keyframes from Videos}
We adopt a straightforward yet effective approach to select representative images from each microsegment. We employ a simple heuristic to identify which keyframes to be selected for the span of the video clip.  In particular, we run an object detector on the video frames and select the frames containing the highest count of object proposals detected by the object detector. This selection ensures that we capture the most visually informative frame from among the available candidates. In the case of ImageBind, we opt for the middle frame from each narration video.

\subsubsection{Tokenization}
As a preprocessing step, we replace multiword tokens with a single word. For instance, each occurrence of \textsl{put-down} is replaced with \textsl{put\_down}, and each occurrence of \textsc{olive oil} is replaced with \textsc{olive\_oil}. This preprocessing step is not applied to LLaMA2 and IDEFICS, since these models have their vocabulary. Similarly, LLaMA2 and IDEFICS use their tokenizers while other models simply use a whitespace tokenizer.

\subsection{Prompt Format}\label{sec:app_prompting_prep}

In this section, we describe the heuristic we employ to formulate the inputs for our evaluation prompts targeting generative models. It is worth noting that the prompting templates for IDEFICS and LLaMA2, though similar, are not interchangeable as IDEFICS can harness both visual and language data.

First, for both LLMs, we include an instruction at the start of the prompt as our language models are instruction-tuned. Then, we enumerate a set of few-shot examples. Finally, we provide the source section at the end of the prompt,  leaving the target to be predicted.

\subsubsection{LLaMA2 Prompt Example}
An example LLaMA2 5-shot prompt can be seen in Fig. \ref{fig:app_prompt_format_llama2}.

\subsubsection{IDEFICS Prompt Example}
An example IDEFICS 1-shot prompt can be seen in the Fig. \ref{fig:app_prompt_format_idefics}. $<$\texttt{Image n}$>$ denotes the image for the $n^{th}$ narration scene.

\section{Exploratory Analysis of \textsc{CompAct}}\label{sec:app_exploratory}

\begin{figure*}[h]
\centering
\includegraphics[width=\textwidth]{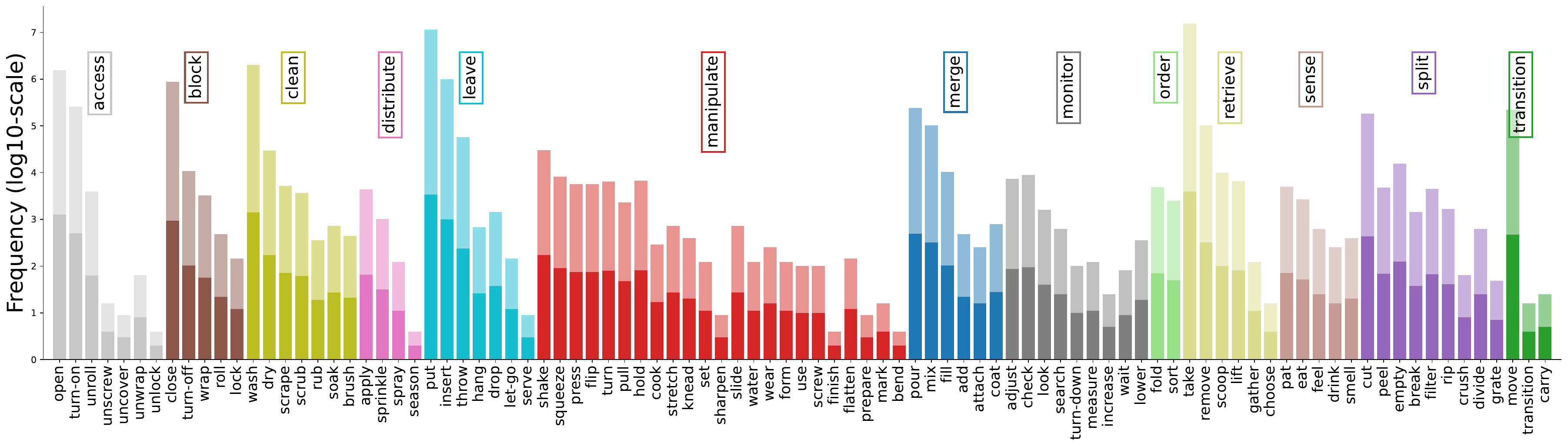}
\\
\includegraphics[width=\textwidth]{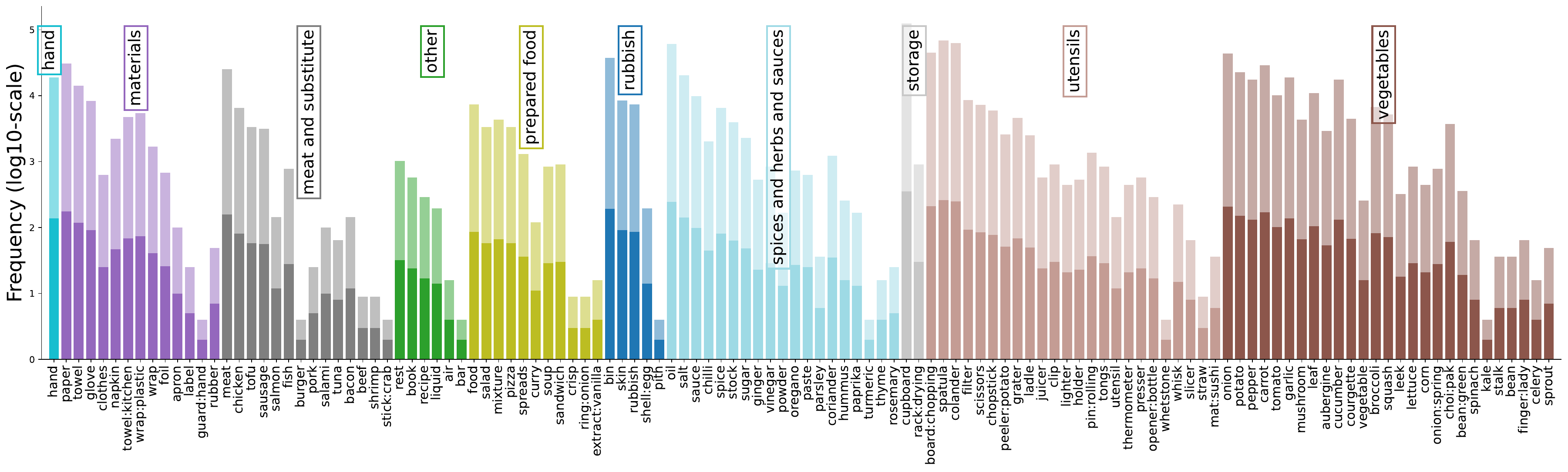}
\\
\includegraphics[width=\textwidth]{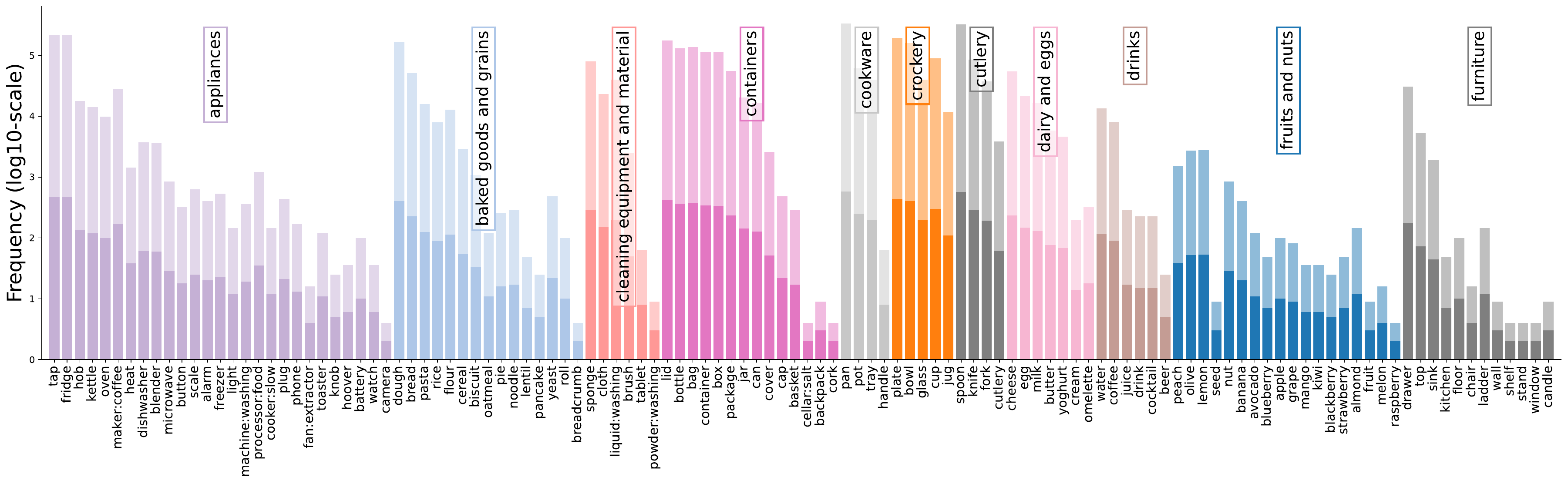}
\caption{Distribution of verbs (top) and nouns (middle and bottom) from \textsc{CompAct}}
\label{fig:verbnoundistributions}
\end{figure*}

In this section, we share an exploratory analysis of the \textsc{CompAct}. Fig. \ref{fig:verbnoundistributions} illustrates the verb and noun distributions in the \textsc{CompAct} dataset where the validation and test splits are jointly stacked on top of the train split occurrences and displayed in a lighter color. % 

\section{Choice of EK-100 for \textsc{CompAct}}

The EPIC-Kitchens-100 (EK-100) dataset was chosen due to its established reputation in the research community and its densely annotated instructions, offering a rich and diverse dataset. It also has a clear segmentation of instructions, including verb and noun annotations, making it an ideal candidate for curating the \textsc{CompAct} dataset, allowing us to leverage audio, vision, and text modalities effectively.
Previously proposed datasets in the literature such as CrossTask \citep{zhukov2019cross} and GAIN \citep{li2022gain} also consists of text instructions and multimodal components. Unlike CrossTask, which focuses on cross-task generalization, our study centers on compositional generalization. Similarly, while sharing similarities with GAIN in dataset formulation and the use of instructional videos, \textsc{CompAct} differs in the description of atomic concepts and the mathematical definition of out-of-distribution (OOD) scenarios. We also conduct further analysis to evaluate whether the proposed benchmarks such as CrossTask or GAIN could be considered for a compositional generalization benchmark. Nevertheless, the lack of proper annotations for atoms and compounds and the number of instances seem to be a challenge to generate compositional splits for these benchmarks.

\section{Implementation Details and Reproducibility}\label{sec:app_implementation}
For the reproducibility of our results, we plan to make the code, models, \textsc{CompAct} splits, and extracted features publicly available under CC BY-NC 4.0 DEED license. All models are implemented with PyTorch. We use torchtext library for the BLEU metric and evaluate library for the BERTScore metric.

\begin{table*}[h]
\small
\caption{Model sizes and their training times for our experiments. Training times are averaged over 3 runs.}
        \centering
        
        \begin{tabular}{@{}lrr@{\extracolsep{6pt}}rrr}
            \toprule
            & \multicolumn{2}{c}{Next Utterance Prediction}  &  \multicolumn{3}{c}{Atom Classification}\\ \cmidrule{2-3}\cmidrule{4-6}
Model & \#params & Train Time & \#params & Noun Train Time  & Verb Train Time \\ \midrule
L     & 4.8M  & 20:45 & 2.1M  & 2:45 & 2:00 \\
OL    & 12.0M  &  38:15 & 9.3M & 12:15 & 8:30 \\
VL    & 12.5M  & 49:15 & 9.7M & 26:45 & 22:00 \\ 
AL    & 12.0M  & 40:00 & 9.3M & 10:45 & 8:15  \\
AVL   & 12.6M  & 52:00 & 9.9M & 15:30 & 13:00  \\
OAL   & 12.1M  & 39:00 & 9.4M & 14:15 & 10:30 \\ \midrule
MerlotR   & 12.1M  & 18:30 & 9.4M & 6:30 & 4:45  \\
ImageBind & 8.4M  & 28:15 & 5.7M  & 9:45  & 8:30 \\
\bottomrule
\end{tabular}\label{tbl:model_sizes_fup}
\label{tbl:model_sizes_all}
\end{table*}

\subsection{Training Regime and Hyperparameters}

We use the AdamW optimizer \citep{loshchilov2017decoupled} with ReduceLROnPlateau learning rate scheduler to reduce the learning rate during training when validation BLEU plateaus. To train the models for the next utterance prediction, we employ cross-entropy loss, and initialize network weights via uniform distribution for both the encoder and the decoder. We use an early stopping strategy and stop the training if validation BLEU does not improve after a certain threshold. We clip gradients set the gradient threshold to 0.1, and use a 3-layer multihead attention with 4-heads in the crossmodal self-attention block in all our multimodal models.  We use the same strategy for atom classification, with one distinction where we use accuracy for early stopping and learning rate scheduler. 

For both tasks, we use 50 epochs as the early stopping threshold. We use a dropout rate of 0.3 and the AdamW optimizer with a 3e-4 learning rate and 5e-5 weight decay. We use the ReduceLROnPlateau learning rate scheduler with patience of 40. Following the insights from \citet{DBLP:conf/emnlp/CsordasIS21}, we use the performance score as a monitoring metric for the scheduler (also early stopping) rather than using loss. For the next utterance prediction and action classification tasks, we use BLEU and accuracy scores, respectively. %

\subsection{Model Sizes and Training Time}

\noindent In Table~\ref{tbl:model_sizes_all}, we present the number of trainable parameters and training time (MM:SS) for all of our trainable baseline models for both the next utterance prediction and atom classification tasks. %

LLaMA2 and IDEFICS experiments are run on NVIDIA Tesla T4 and NVIDIA Tesla V100 GPUs respectively. Other experiments are run on NVIDIA 1080Ti GPUs.

\section{Further Analysis}\label{sec:app_analysis}

\subsection{Generalization on Validation Split}

\begin{table}[h]

\caption{Next utterance prediction results on validation split. Using audio, visual, or object features always improves performance compared to the language-only unimodal baseline. We report the mean and the standard deviation across three runs.}
\centering

\resizebox{\linewidth}{!}{
\begin{tabular}{lcccc}
\toprule
 & BLEU & EM & CA  & BERTScore  \\
\midrule
L & 21.43\sta{}{0.5} & 2.88\sta{}{0.1} & 6.22\sta{}{0.2} & 79.20\sta{}{0.1} \\
VL & 30.59\sta{}{0.4} & 7.35\sta{}{0.6} & 12.39\sta{}{1} & 81.24\sta{}{0.4} \\
AL & 30.47\sta{}{0.1} & 7.06\sta{}{0.3} & 12.16\sta{}{0.2} & 81.19\sta{}{0.1} \\
AVL & 31.22\sta{}{0.1} & 7.44\sta{}{0.4} & 12.54\sta{}{0.3} & 81.44\sta{}{0.1} \\
OL & 30.50\sta{}{0.4} & 7.03\sta{}{0.2} & 12.42\sta{}{0.8} & 81.10\sta{}{0.1} \\
OAL & \underline{31.42\sta{}{0.1}} & \underline{7.99\sta{}{0.5}} & \underline{13.36\sta{}{0.2}} & \underline{81.50\sta{}{0.1}} \\ \midrule
MerlotR & 31.36\sta{}{0.4} & 7.17\sta{}{0.6} & 12.68\sta{}{0.5} & 81.34\sta{}{0.1} \\
ImageBind & \textbf{34.13\sta{}{0.5}} & \textbf{10.45\sta{}{0.8}} & \textbf{16.08\sta{}{0.8}} & \textbf{82.45\sta{}{0.2}} \\
IDEFICS & 25.15\sta{}{0.8} & 5.66\sta{}{0.5} & 7.17\sta{}{0.5} & 80.75\sta{}{0.2} \\
LLaMA2 & 26.52\sta{}{0.5} & 5.37\sta{}{0.3} & 6.99\sta{}{0.4} & 78.59\sta{}{0.1} \\

\bottomrule
\end{tabular}
}

\label{tbl:app_results_fup_quantitative_val}
\end{table}

\begin{table}[!ht]
\caption{Atom classification results on validation split. We report the mean across three runs. The best and second best-performing results are highlighted in bold and underlined, respectively. }
\centering
\resizebox{\linewidth}{!}{%{
\begin{tabular}{@{}llccc}
    \toprule
    \multirow{12}{*}{\rotatebox{90}{Verb Classification $\;$}} & & EM & CA & BERTScore \\ \midrule
    & L & 12.92\sta{}{0.8} & 28.96\sta{}{3.3} & 74.96\sta{}{0.4} \\
    & VL & 14.02\sta{}{0.2} & 30.23\sta{}{1.7} & 75.19\sta{}{0.5} \\
    & AL & \underline{14.48\sta{}{0.7}} & 31.07\sta{}{3.7} & \underline{76.21\sta{}{0.2}} \\
    & AVL & 14.01\sta{}{0.3} & 31.15\sta{}{2.5} & 75.84\sta{}{0.9} \\
    & OL & 12.77\sta{}{0.3} & 30.87\sta{}{1.1} & 75.53\sta{}{0.3} \\
    & OAL & 14.30\sta{}{0.2} & 30.79\sta{}{1.5} & 76.02\sta{}{0.6} \\ \noalign{\vspace{0.25ex}}\cline{2-5}\noalign{\vspace{0.25ex}}
    & MerlotR & 13.15\sta{}{0.5} & \textbf{32.53\sta{}{0.6}} & 75.74\sta{}{0.2} \\
    & ImageBind & \textbf{14.91\sta{}{0.2}} & \underline{31.31\sta{}{3.1}} & \textbf{76.28\sta{}{0.5}} \\ \noalign{\vspace{0.25ex}}\cline{2-5}\noalign{\vspace{0.25ex}}
    &         MRH & -- & -- & -- \\
    \midrule
    \multirow{9}{*}{\rotatebox{90}{Noun Classification $\;$}} & L          & \underline{44.78\sta{}{0.8}} & \underline{52.28\sta{}{0.8}} & \underline{86.35\sta{}{0.1}} \\
     & VL & 42.35\sta{}{0.3} & 49.38\sta{}{0.4} & 85.88\sta{}{0.1} \\
     & AL & 43.71\sta{}{0.2} & 50.79\sta{}{0.2} & 86.05\sta{}{0.1} \\
     & AVL & 43.48\sta{}{0.5} & 50.59\sta{}{0.8} & 86.10\sta{}{0.2} \\
     & OL & 44.03\sta{}{0.3} & 51.40\sta{}{0.1} & 86.03\sta{}{0.1} \\
     & OAL & 44.13\sta{}{0.5} & 50.86\sta{}{0.8} & 86.09\sta{}{0.2} \\ \noalign{\vspace{0.25ex}}\cline{2-5}\noalign{\vspace{0.25ex}}
     & MerlotR & 44.52\sta{}{0.8} & 51.31\sta{}{0.6} & 86.22\sta{}{0.2} \\
     & ImageBind & 34.15\sta{}{0.5} & 44.59\sta{}{0.2} & 84.11\sta{}{0.1} \\ \noalign{\vspace{0.25ex}}\cline{2-5}\noalign{\vspace{0.25ex}}
      & MRH      & \textbf{57.51} & \textbf{60.90} & \textbf{89.89} \\\bottomrule
\end{tabular}}
    
\label{tbl:results_cls_quantitative_combined_val}
\end{table}

In Table \ref{tbl:app_results_fup_quantitative_val} we present generalization performance on the validation split for the next utterance prediction task and in Table \ref{tbl:results_cls_quantitative_combined_val} we demonstrate the generalization performance on validation split for the atom classification task.

\subsection{Generalization Performance over Epochs}

\begin{figure*}[h]
\centering
\includegraphics[width=0.32\textwidth]{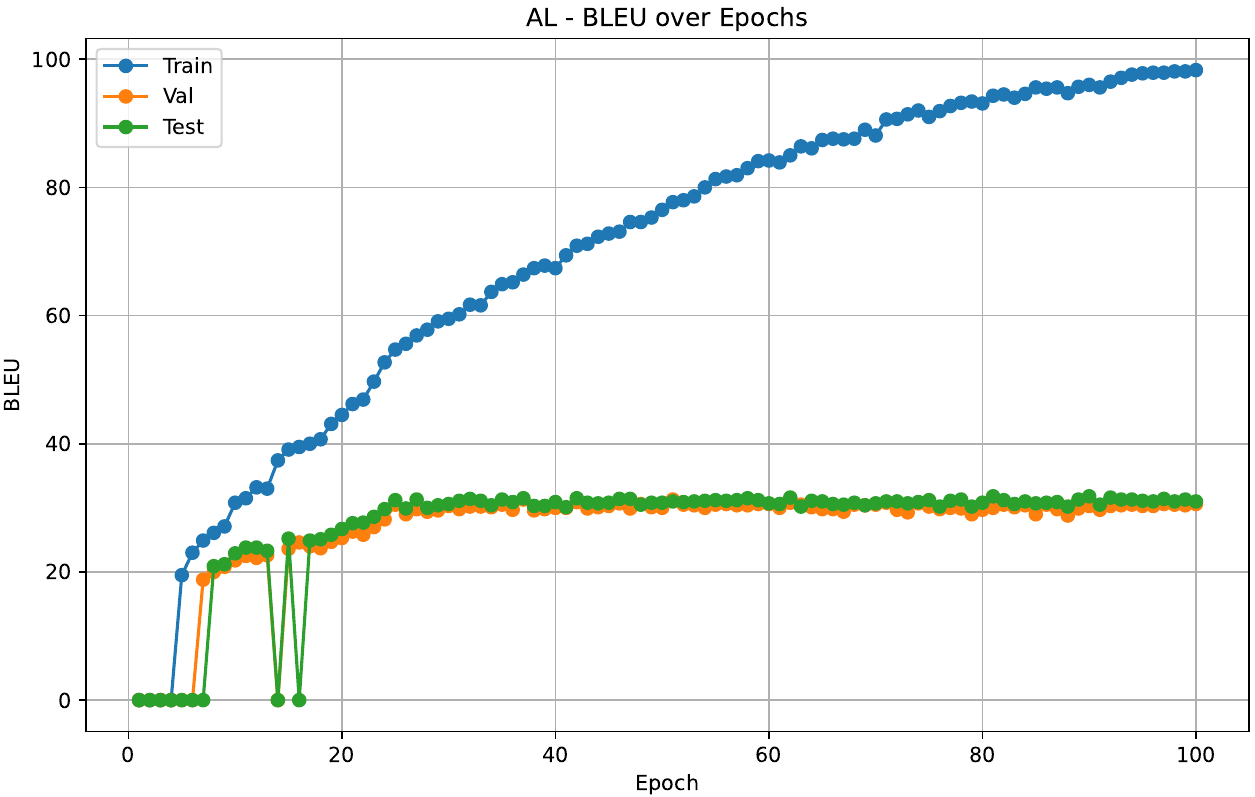}
\includegraphics[width=0.32\textwidth]{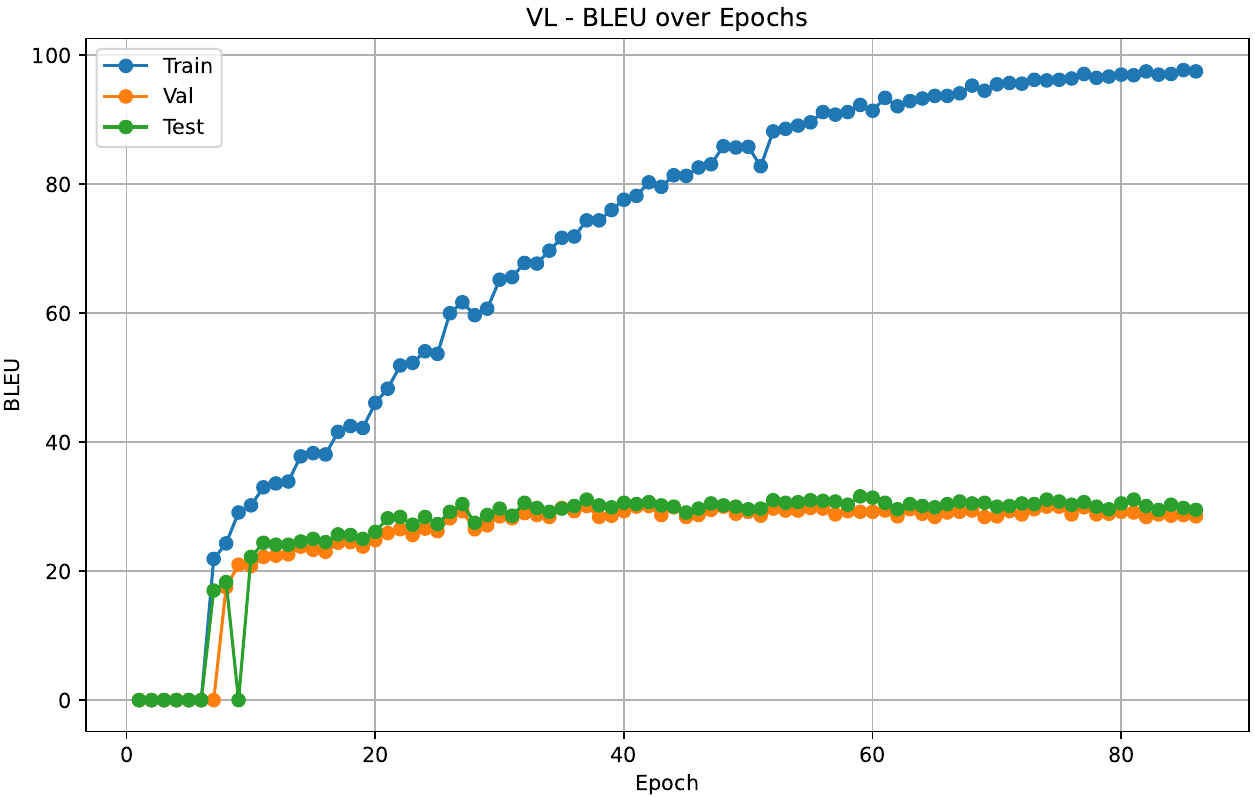}
\includegraphics[width=0.32\textwidth]{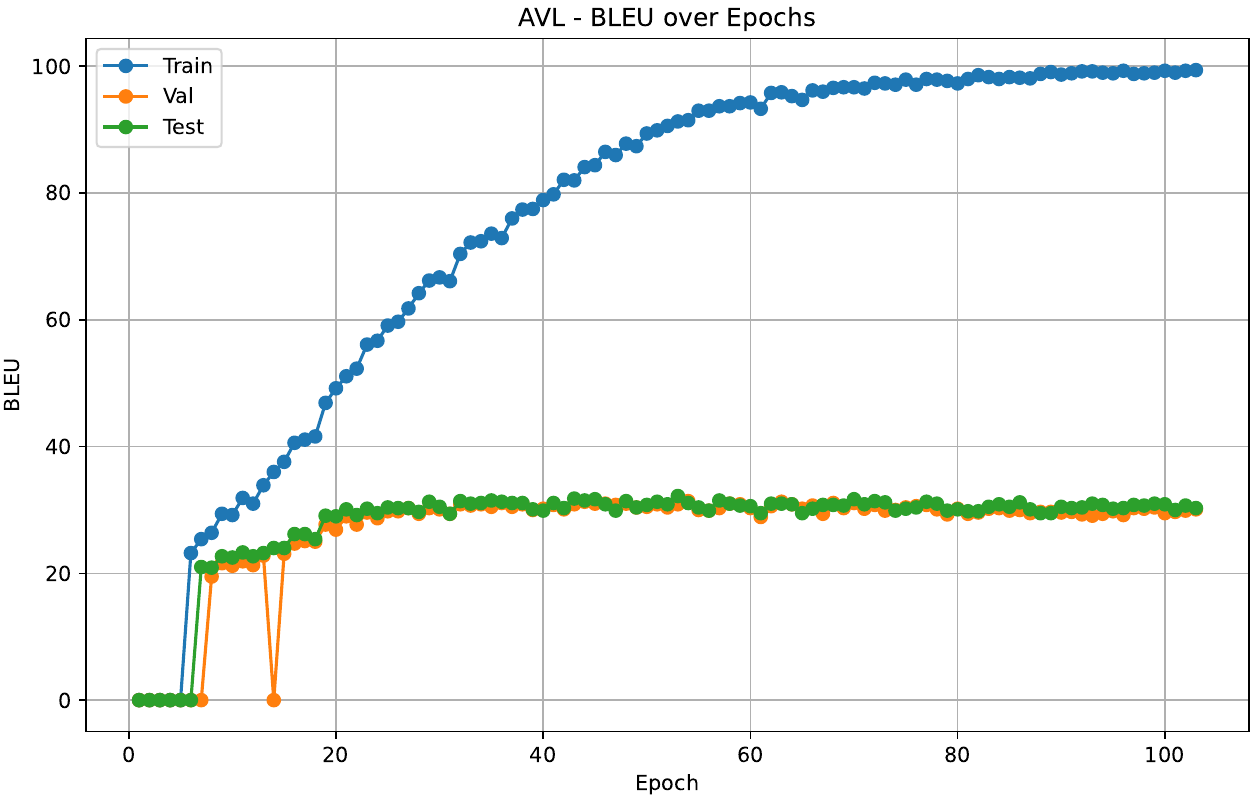}
\caption{Generalization performance of the AL, VL, and AVL models over the epochs. Even though the training performance of a model improves on \textsc{CompAct}, this does not necessarily mean that its validation and test performance will also become better due to the compositional nature of the \textsc{CompAct} dataset.}
\label{fig:epochs}
\end{figure*}

In Fig. \ref{fig:epochs}, we report the BLEU scores of the models over the training, validation, and test splits at different epochs. These plots demonstrate that in a compositional setup, models can perform well in the training set but this does not mean they can generalize to unseen distributions.

\subsection{Prompting Ablations}\label{sec:app_prompting_abl}

\subsubsection{Additional Few-Shot Results}
Table \ref{tbl:app_results_fup_quant_kshot_idefics} and \ref{tbl:app_results_fup_quant_kshot_llama2} offer interesting insights regarding few-shot compositional capabilities of IDEFICS and LLaMA2 models. First, we see a significant performance discrepancy between IDEFICS and LLaMA2 on zero-shot prediction results. As IDEFICS additionally utilizes visual information over LLaMA2, it displays better zero-shot generalization capabilities. While LLaMA2 outperforms IDEFICS in one-shot and few-shot BLEU scores, contrastingly, IDEFICS outperforms LLaMA2 on BERTScores. As LLaMA2 outperforms the LLM of IDEFICS (instruct-tuned LLaMA1) on many benchmarks~\citep{Touvron2023Llama2O}, we infer that LLaMA2 can imitate the vocabulary of few-shot examples better than IDEFICS, resulting in higher BLEU scores. However, higher BERTScores imply that IDEFICS can reflect the semantics of the ground truth prediction better.

\begin{table}[h]
\caption{Next utterance prediction results on test split for IDEFICS. As the few-shot example count increases, performance improves on every metric consistently.}
\centering
\resizebox{\linewidth}{!}{%
\begin{tabular}{lcccc}
\toprule
$k$-shot & BLEU & EM & CA  & BERTScore  \\
\midrule
0-shot & 8.98\sta{}{0.2} & 0.06\sta{}{0.1} & 0.12\sta{}{0.0} & 75.58\sta{}{0.1} \\
1-shot & 20.25\sta{}{0.2} & 4.12\sta{}{0.1} & 5.30\sta{}{0.1} & 79.75\sta{}{0.0} \\
3-shot & 24.85\sta{}{0.7} & 5.37\sta{}{0.4} & 7.20\sta{}{0.3} & 80.78\sta{}{0.1} \\
5-shot & 25.64\sta{}{0.4} & 5.76\sta{}{0.1} & 7.89\sta{}{0.5} & 80.92\sta{}{0.1} \\
8-shot & 26.18\sta{}{0.3} & 6.06\sta{}{0.2} & 7.92\sta{}{0.3} & 81.19\sta{}{0.1} \\
\bottomrule
\end{tabular}}
\label{tbl:app_results_fup_quant_kshot_idefics}
\end{table}

\begin{table}[h]
\caption{Next utterance prediction results on test split for LLaMA2. As the few-shot example count increases, performance improves on every metric consistently.}
\centering
\resizebox{\linewidth}{!}{%
\begin{tabular}{lcccc}
\toprule
$k$-shot & BLEU & EM & CA  & BERTScore  \\
\midrule
0-shot & 2.02\sta{}{3.5} & 0.13\sta{}{0.1} & 0.15\sta{}{0.1} & 71.68\sta{}{0.1} \\
1-shot & 23.89\sta{}{0.7} & 3.98\sta{}{0.2} & 5.77\sta{}{0.1} & 77.90\sta{}{0.2} \\
3-shot & 26.17\sta{}{0.6} & 5.07\sta{}{0.1} & 7.00\sta{}{0.1} & 78.35\sta{}{0.1} \\
5-shot & 27.50\sta{}{0.6} & 5.36\sta{}{0.6} & 7.41\sta{}{0.7} & 78.76\sta{}{0.2} \\
8-shot & 27.58\sta{}{0.3} & 5.60\sta{}{0.2} & 8.01\sta{}{0.4} & 78.95\sta{}{0.1} \\
\bottomrule
\end{tabular}}
\label{tbl:app_results_fup_quant_kshot_llama2}
\end{table}

\subsubsection{Few-Shot Example Selection}
For few-shot example selection, rather than randomly picking $k$-shot examples, we employ a simple heuristic. As \citet{liu-etal-2022-makes} highlights that selecting similar examples improves in-context learning performance, we select the most similar $k$ examples as few-shot examples. The similarity measure between the two examples is based on the noun and verb overlap. First, the intersection between the set of nouns and the set of verbs between the main example and all training examples is computed. If the sum of the cardinality of these sets is the largest between the main example and a few-shot example, the few-shot example is the most similar example of the main example. We provide a validation comparison between the random example selection and our heuristic in Table \ref{tbl:app_results_fup_quant_fshot_selection}.

\begin{table}[h]
\caption{Next utterance prediction BLEU scores on validation split for IDEFICS for a single run. Greedy decoding is used and the best score is bolded.}
\centering
\resizebox{\linewidth}{!}{%
\begin{tabular}{lcccc}
\toprule
Strategy & 0-shot & 1-shot & 3-shot & 5-shot  \\
\midrule
Random selection & \textbf{28.5} & 31.0 & 18.7 & 19.3 \\
Our heuristic & \textbf{28.5} & \textbf{34.8} & \textbf{23.4} & \textbf{23.5} \\
\bottomrule
\end{tabular}}
\label{tbl:app_results_fup_quant_fshot_selection}
\end{table}

\subsubsection{Prompt Template Selection}
For IDEFICS, as images should be included in the prompt, the selection of a prompt template is important. We compared two prompt templates (see Fig. \ref{fig:app_prompt_format_idefics} and Fig. \ref{fig:app_prompt_format_idefics_alt}) and after preliminary analysis, used the best-performing template in our paper (see Table \ref{tbl:app_results_fup_quant_template_select}).

\begin{table}[!ht]
\caption{Next utterance prediction results on validation split for IDEFICS. Overall, the used template outperforms unused template.}
\centering
\resizebox{\linewidth}{!}{%
\begin{tabular}{lcccc}
\toprule
Template & BLEU & EM & CA  & BERTScore  \\
\midrule
Used & \textbf{25.64\sta{}{0.4}} & 5.76\sta{}{0.1} & \textbf{7.89\sta{}{0.5}} & \textbf{80.92\sta{}{0.1}} \\
Unused & 22.19\sta{}{0.3} & \textbf{5.91\sta{}{0.4}} & 7.14\sta{}{0.5} & 80.53\sta{}{0.1} \\
\bottomrule
\end{tabular}}
\label{tbl:app_results_fup_quant_template_select}
\end{table}

\begin{figure}[ht]
    \centering
    \noindent\fbox{%
    \parbox{\linewidth}{%
{\footnotesize \texttt{Predict the next action narration given 3 sequential previous actions (image-narration pairs) in a cooking video.}}

 {\footnotesize \texttt{Narration 1: put down bowl Image 1: $<$Image 1$>$}}
 
{\footnotesize \texttt{Narration 2: move frying pan Image 2: $<$Image 2$>$}}
 
 {\footnotesize \texttt{Narration 3: pick up spatula Image 3: $<$Image 3$>$}}
 
 {\footnotesize \texttt{Narration 4: put down spatula}}

 \vspace{1em}

 {\footnotesize \texttt{Narration 1: pick up tins Image 1: $<$Image 1$>$}}
 
 {\footnotesize \texttt{Narration 2: put down tins Image 2: $<$Image 2$>$}}
 
 {\footnotesize \texttt{Narration 3: move bowl Image 3: $<$Image 3$>$}}
 
 {\footnotesize \texttt{Narration 4: }}

    }%
}
    \caption{The unused alternative prompt template for IDEFICS evaluation.}
    \label{fig:app_prompt_format_idefics_alt}
\end{figure}

\subsection{Unseen Compositions for Pre-trained LLMs}

We recognize the challenge of ensuring that pre-trained LLMs have not been exposed to certain compositions during training. Our motivation to explore pre-trained LLMs is inspired by these models' recent successes in various tasks, and to understand how these models succeed in compositional generalization, we conduct experiments with in-distribution data (randomly generated training/validation/test splits), to implicitly determine the extent of prior exposure to unseen compositions in these models.

Our comprehensive analysis, depicted in Table \ref{tbl:rnd_vs_sys_nup}, showcases significant performance improvements in in-domain setting for various baseline models and multimodal LLMs. This contrasts with their performance in out-of-domain setting, highlighting the rigorous nature of our compositional task. The difficulty these models face in generalizing to novel compositions, despite the possibility of exposure to similar examples, indicates a crucial challenge in current multimodal learning.

\subsection{Video-Based Baseline Experiments}

We expand our evaluation to understand the video-based baseline performance through an analysis with the Otter model \citet{li2023otter}. This instruction-tuned Video Language Model (VLM) processes videos as sequential images and was assessed using few-shot prompting. Our findings, detailed in Table \ref{tbl:otter_results}, indicate that while Otter performs similarly over OpenFlamingo-9B, the gains with increased context examples are not as substantial as anticipated. 

\begin{table}[]\caption{Otter Model Results}
\centering
\begin{tabular}{@{}lcccc@{}}
\toprule
       & BLEU  & EM   & CA   & BERTScore \\ \midrule
0-shot & 9.31  & 0.01 & 0.03 & 74.54     \\
1-shot & 9.26  & 0.16 & 0.26 & 73.30     \\
3-shot & 9.95  & 0.45 & 0.54 & 74.21     \\
5-shot & 10.5  & 0.41 & 0.49 & 74.83     \\
8-shot & 10.85 & 0.34 & 0.43 & 74.38     \\ \bottomrule
\end{tabular}

\label{tbl:otter_results}
\end{table}

\subsection{Comparison of Model Sizes}

We conduct additional experiments to compare model performance across different scales using different sizes of the OpenFlamingo model, specifically OpenFlamingo-3B-vitl-mpt1b and OpenFlamingo-9B-vitl-mpt7b. Using a 5-shot prompting approach, we present our evaluation results in Table \ref{tbl:model_sizes}.

\begin{table}[]\caption{Model size comparison results for OpenFlamingo}
\centering
\resizebox{\linewidth}{!}{%
\begin{tabular}{@{}lcccc@{}}
\toprule
                & BLEU  & EM   & CA   & BERTScore \\ \midrule
OpenFlamingo-3B & 8.96  & 0.09 & 0.15 & 73.34     \\
OpenFlamingo-9B & 11.15 & 0.75 & 0.87 & 72.38     \\ \bottomrule
\end{tabular}
}
\label{tbl:model_sizes}
\end{table}

Interestingly, the OpenFlamingo-9B's performance is substantially lower than the IDEFICS results (11 BLEU for OpenFlamingo vs. 24 BLEU for IDEFICS). We attribute this discrepancy primarily to the models' differing abilities to integrate interleaved images and to the instruction-tuned nature of IDEFICS's LLM, enhancing its prompt adherence. Although scaling up model size shows some performance improvement, the disparities with other baselines are noteworthy. This observation aligns with ongoing discussions in the field about the impact of model scale on performance, as thoroughly investigated by \citet{qiu2022evaluating}.

\subsection{Analysis of the Failure Cases}

We delve deeper into failure cases, particularly examining instances where a correctly predicted verb is paired with a misclassified noun. Our analysis focus on whether these nouns are more likely to match with training compositions. Additionally, we want to reiterate that our train/val/test split curation in CompAct was meticulously designed to ensure similar primitive coverage across splits while varying compound compositions (see Fig. \ref{fig:atomcompound}).

In particular, in Table \ref{tbl:failure_cases}, we share the percentage of the misclassified nouns for correctly predicted verbs for the training and test compositions. These ratios are averaged across 3 runs for each model.

\begin{table}[]\caption{Percentage of misclassified nouns in train and test split V-N compositions}
\centering

\begin{tabular}{@{}lcc@{}}
\toprule
          & Train Ratio & Test Ratio \\ \midrule
L         & 9.97\%      & 7.63\%     \\
VL        & 10.64\%     & 8.04\%     \\
AL        & 10.98\%     & 8.34\%     \\
AVL       & 10.65\%     & 8.21\%     \\
OL        & 10.47\%     & 8.06\%     \\
OAL       & 10.44\%     & 8.12\%     \\
MerlotR   & 10.58\%     & 8.13\%     \\
ImageBind & 10.39\%     & 8.01\%     \\ \bottomrule
\end{tabular}
\label{tbl:failure_cases}
\end{table}

\subsection{t-SNE Visualization for Audio and Visual Embeddings}

\begin{figure}[ht]
\centering
\includegraphics[width=0.5\textwidth]{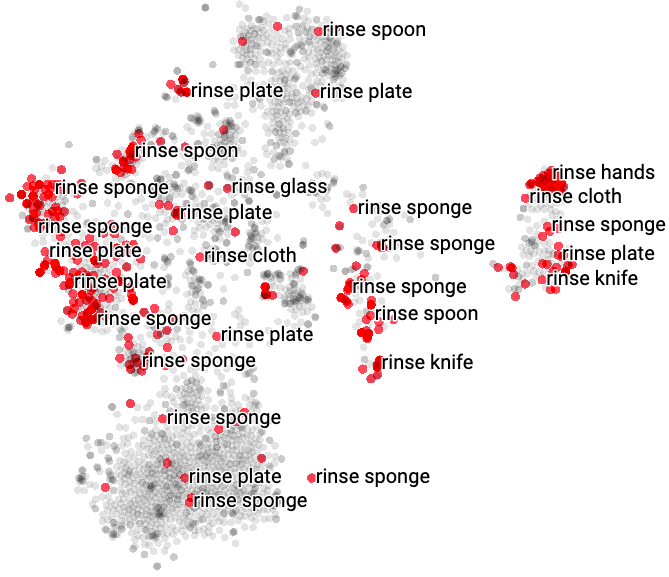}
\hspace{5mm}
\includegraphics[width=0.45\textwidth]{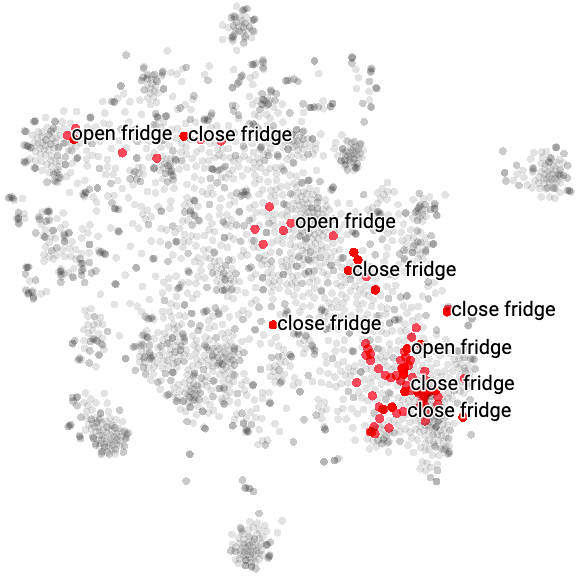}\\
\caption{Feature projection to 2D space with t-SNE using raw audio and global visual features. At the top, audio space is shown with the verb \textsl{rinse} being specifically highlighted. At the bottom, visual space is given with noun \textsc{fridge} being particularly highlighted. Sampled by most common compounds appearing at least 25 times in \textsc{CompAct}, equally distributed for each \mbox{compound~($N=\;$25).}}
\label{fig:tsne}
\end{figure}

To understand the features we extracted via VGGSound and ResNet50 backbones and how well they encode the audio and visual spaces, we visualized the raw feature embeddings by projecting them to 2D space via t-SNE. In Fig.~\ref{fig:tsne}, we highlight the compounds with the verb \textsl{rinse} and observe that audio features can meaningfully encode activities. Similarly, we analyzed the raw global visual embeddings and highlighted the compounds with the noun \textsc{fridge}, the visualization shows the extracted global visual embeddings can effectively encode visual surroundings.

\end{document}